\documentclass{article}
\PassOptionsToPackage{table}{xcolor}

\usepackage[preprint,nonatbib]{neurips_2025}

\usepackage[utf8]{inputenc} % allow utf-8 input
\usepackage[T1]{fontenc}    % use 8-bit T1 fonts
\usepackage{hyperref}       % hyperlinks
\usepackage{url}            % simple URL typesetting
\usepackage{booktabs}       % professional-quality tables
\usepackage{amsfonts}       % blackboard math symbols
\usepackage{nicefrac}       % compact symbols for 1/2, etc.
\usepackage{microtype}      % microtypography

\usepackage{xcolor}         % colors
\usepackage{graphicx}    % 提供resizebox
\usepackage{tabularx}    % 支持自动换行的X列
\usepackage{makecell}    % 处理单元格内换行
\usepackage{booktabs}    % 提供\toprule等命令
\usepackage{multirow}
\usepackage{tikz}
\usepackage{booktabs}
\usepackage{enumitem}
\usepackage{amssymb}    % 提供 ✓ 和 ✗ 符号
\usepackage{pifont}     % 提供更多符号\
\usepackage{fontawesome5} % 提供图标库
\usepackage{amsmath}
\usepackage{longtable}
\usepackage{listings}
\usepackage{tcolorbox}
\tcbuselibrary{listings}
\tcbuselibrary{skins, breakable}

\newtcolorbox{promptbox}{
    colback=white,       % 背景色
    colframe=green!40,    % 淡蓝色边框
    fontupper=\ttfamily\small, % 等宽字体，并使用 \small 命令减小字号
    breakable,           % 允许跨页
    enhanced,            % 启用增强样式
    boxrule=0.5pt,       % 边框粗细
    arc=3pt,             % 边框圆角
    left=6pt,            % 左内边距
    right=6pt,           % 右内边距
    top=6pt,             % 顶部内边距
    bottom=6pt,          % 底部内边距
    boxsep=0pt,          % 内容与边框间距
    before skip=10pt,    % 上方间距
    after skip=10pt      % 下方间距
}

\definecolor{DeepOrange}{RGB}{255,102,0}  % 深橘红 (Pantone 158C)
\definecolor{LightOrange}{RGB}{255,178,102} % 浅橘红 (Pantone 155C)

\title{ArtiMuse: Fine-Grained Image Aesthetics Assessment with Joint Scoring and Expert-Level Understanding}

\author{
 \makebox[.23\textwidth][c]{ Shuo Cao\thanks{This work was done during his internship at Shanghai AI Laboratory. \\ \hspace*{1em} $^{\dagger}$ Corresponding authors.}}\\
\makebox[.23\textwidth][c]{USTC, Shanghai AI Lab}\\
 \makebox[.23\textwidth][c]{\texttt{caoshuo@pjlab.org.cn}} 
 \vspace{-0.5em}
\And
\hspace{2em} \makebox[.23\textwidth][c]{Nan Ma}\\
\hspace{2em}\makebox[.23\textwidth][c]{China Academy of Art}\\
\hspace{2em}\makebox[.23\textwidth][c]{\texttt{0118019@caa.edu.cn}} \vspace{-0.5em}
\And
\hspace{2em}\makebox[.23\textwidth][c]{Jiayang Li}\\
\hspace{2em}\makebox[.23\textwidth][c]{Peking University}\\
\hspace{2em}\makebox[.23\textwidth][c]{\texttt{lijiayang.cs@gmail.com}} \vspace{-0.5em}
\AND
 \makebox[.23\textwidth][c]{Xiaohui Li}\\
\makebox[.23\textwidth][c]{SJTU, Shanghai AI Lab}\\
\makebox[.23\textwidth][c]{\texttt{lixiaohui@pjlab.org.cn}} \vspace{-0.5em}
\And
\hspace{2em}\makebox[.23\textwidth][c]{Lihao Shao}\\
\hspace{2em}\makebox[.23\textwidth][c]{China Academy of Art}\\
\hspace{2em}\makebox[.23\textwidth][c]{\texttt{shaolihao90@caa.edu.cn}} \vspace{-0.5em}
\And
\hspace{2em} \makebox[.23\textwidth][c]{Kaiwen Zhu}\\
\hspace{2em} \makebox[.23\textwidth][c]{SJTU, Shanghai AI Lab}\\
\hspace{2em} \makebox[.23\textwidth][c]{\texttt{zhukaiwen@pjlab.org.cn}} \vspace{-0.5em}
\AND
 \makebox[.23\textwidth][c]{Yu Zhou}\\
 \makebox[.23\textwidth][c]{Sun Yat-sen University}\\
 \makebox[.23\textwidth][c]{\texttt{zhouy635@mail2.sysu.edu.cn}} \vspace{-0.5em}
\And
\hspace{2em}\makebox[.23\textwidth][c]{Yuandong Pu}\\
\hspace{2em}\makebox[.23\textwidth][c]{SJTU, Shanghai AI Lab}\\
\hspace{2em}\makebox[.23\textwidth][c]{\texttt{puyuandong@pjlab.org.cn}} \vspace{-0.5em}
\And
\hspace{2em} \makebox[.23\textwidth][c]{Jiarui Wu}\\
 \hspace{2em} \makebox[.23\textwidth][c]{CUHK}\\
 \hspace{2em} \makebox[.23\textwidth][c]{\texttt{wujiarui@buaa.edu.cn}} \vspace{-0.5em}
\AND
\makebox[.23\textwidth][c]{Jiaquan Wang}\\
\makebox[.23\textwidth][c]{Hong Kong PolyU}\\
\makebox[.23\textwidth][c]{\texttt{23114819g@connect.polyu.hk}} \vspace{-0.5em}
\And
\hspace{2em}\makebox[.23\textwidth][c]{Bo Qu}\\
\hspace{2em}\makebox[.23\textwidth][c]{Shanghai AI Lab}\\
\hspace{2em}\makebox[.23\textwidth][c]{\texttt{qubo@pjlab.org.cn}} \vspace{-0.5em}
\And
\hspace{2em} \makebox[.23\textwidth][c]{Wenhai Wang}\\
\hspace{2em} \makebox[.23\textwidth][c]{Shanghai AI Lab, CUHK}\\
\hspace{2em} \makebox[.23\textwidth][c]{\texttt{wangwenhai362@gmail.com}} \vspace{-0.5em}
\And
\makebox[.23\textwidth][c]{Yu Qiao}\\
\makebox[.23\textwidth][c]{Shanghai AI Lab}\\
\makebox[.23\textwidth][c]{\texttt{yu.qiao@siat.ac.cn}} \vspace{-0.5em}
\And
\hspace{2em} \makebox[.23\textwidth][c]{Dajuin Yao$^\dagger$}\\
\hspace{2em} \makebox[.23\textwidth][c]{China Academy of Art}\\
\hspace{2em} \makebox[.23\textwidth][c]{\texttt{0616009@caa.edu.cn}} \vspace{-0.5em}
\And
\hspace{2em} \makebox[.23\textwidth][c]{Yihao Liu$^\dagger$}\\
\hspace{2em} \makebox[.23\textwidth][c]{Shanghai AI Lab}\\
\hspace{2em} \makebox[.23\textwidth][c]{\texttt{liuyihao@pjlab.org.cn}} \vspace{-0.5em}
}
\begin{document}
\maketitle

\begin{figure}[htbp]
  \centering
  \vspace{-20pt}
   \includegraphics[width=0.99\linewidth]{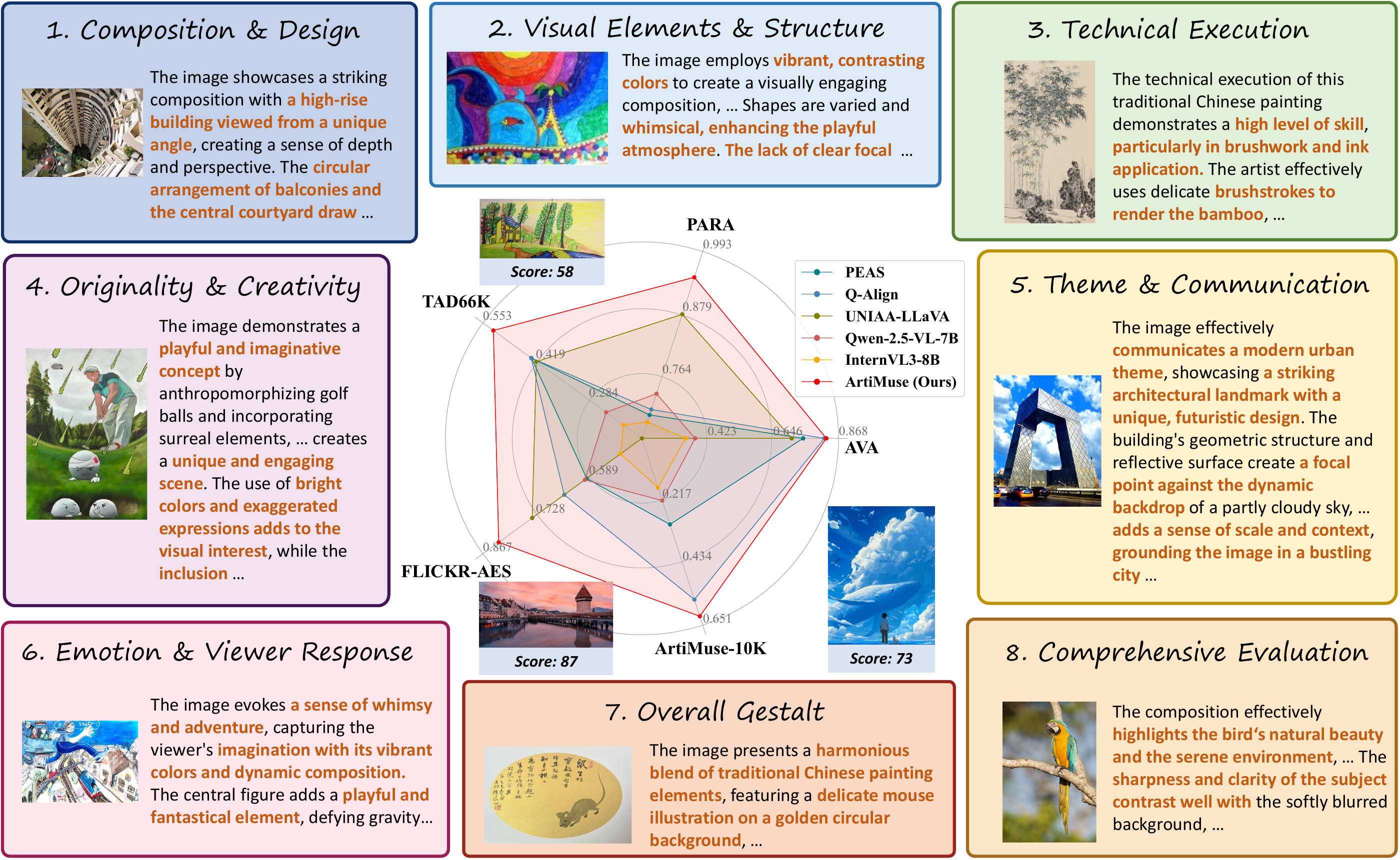}
   \vspace{-10pt}
   \caption{ArtiMuse provides granular, expert-level textual understanding results for images across eight fine-grained aesthetic attributes. Additionally, it achieves precise image aesthetics scoring, significantly outperforming state-of-the-art models across multiple widely-used benchmarks.}
   \label{fig:teaser}
   \vspace{-5pt}
\end{figure}

% significantly outperforming state-of-the-art models—including Q-Align, Qwen-2.5-VL-7B, and InternVL-3-8B—on multiple widely-used benchmarks

% \footnotetext[1]{Since UNIAA-LLaVA has not been open-sourced, its test results on ArtiMuse-10K are recorded as zero.}
% ArtiMuse从8个fine-grained aesthetic analysis维度对图像进行美学评估，能够给出细粒度的、专家级别的文字分析。此外，ArtiMuse还可以同时实现精确的image aesthetics scoring，在多个widely-used benchmark上大幅超越了包括Q-Align、Qwen-2.5-VL-7B和InternVL-3-8B在内的最新模型。
% 需要指出,由于UNIAA-LLaVA无法进行测试,因此其在ArtiMuse-10K上的测试结果被标记为0

\begin{abstract}
  The rapid advancement of educational applications, artistic creation, and AI-generated content (AIGC) technologies has substantially increased practical requirements for comprehensive Image Aesthetics Assessment (IAA), particularly demanding methods capable of delivering both quantitative scoring and professional understanding. Multimodal Large Language Model (MLLM)-based IAA methods demonstrate stronger perceptual and generalization capabilities compared to traditional approaches, yet they suffer from modality bias (score-only or text-only) and lack fine-grained attribute decomposition, thereby failing to support further aesthetic assessment. In this paper, we present:
  (1) \textbf{ArtiMuse}, an innovative MLLM-based IAA model with Joint Scoring and Expert-Level Understanding capabilities; (2) \textbf{ArtiMuse-10K}, the first expert-curated image aesthetic dataset comprising 10,000 images spanning 5 main categories and 15 subcategories, each annotated by professional experts with 8-dimensional attributes analysis and a holistic score. Both the model and dataset will be made public to advance the field. The project page is available at \url{https://thunderbolt215.github.io/ArtiMuse-project/}.
\end{abstract}

\section{Introduction}
\label{sec:intro}

% significantly lowering the barriers to image generation and enabling a surge of creative possibilities.

In the era of digitalization and visual information explosion, images have become an essential medium for human beings to perceive the world, document daily life, and express emotions. From professional photography and painting to casual snapshots and sharing, images play a crucial role in conveying aesthetic values, emotional narratives, and storytelling. The advent of artificial intelligence generated content (AIGC) technologies~\cite{dreamlike_photoreal_2023, flux, sd} has further democratized visual content creation. However, this abundance of visual content also poses new challenges for quality assessment, filtering, and recommendation. While existing image quality assessment (IQA) techniques~\cite{depictqa_v1, depictqa_v2, deqa_score} have matured in detecting low-level degradations such as blurriness, noise, and compression artifacts, they largely focus on the technical fidelity of images and fail to capture their higher-level aesthetic attributes. Image aesthetics assessment (IAA)~\cite{aesexpert, AesMamba, apddv2}, which evaluates aspects such as artistic appeal, color harmony, and emotional expression, is increasingly recognized as a fundamental capability in applications including AIGC content evaluation, creative assistance, and photography education.

Despite the growing demand, current IAA methods face notable limitations. Most existing approaches rely on simplistic score predictions without capturing the inherent subjectivity, multidimensionality, and nuanced interpretations of aesthetics. Moreover, available datasets are often small in scale, coarse in granularity, and lack professionally curated annotations based on established aesthetic theories. This gap severely limits the ability of state-of-the-art multimodal large models (MLLMs)~\cite{internvl3, qwen2.5} to understand and reason about aesthetics.
% thereby impeding their effectiveness in advanced applications such as AIGC quality control, intelligent art critique, and user-centered creative feedback.

\begin{figure}[htbp]
  \centering
  \vspace{-5pt}
    \includegraphics[width=0.99\linewidth]{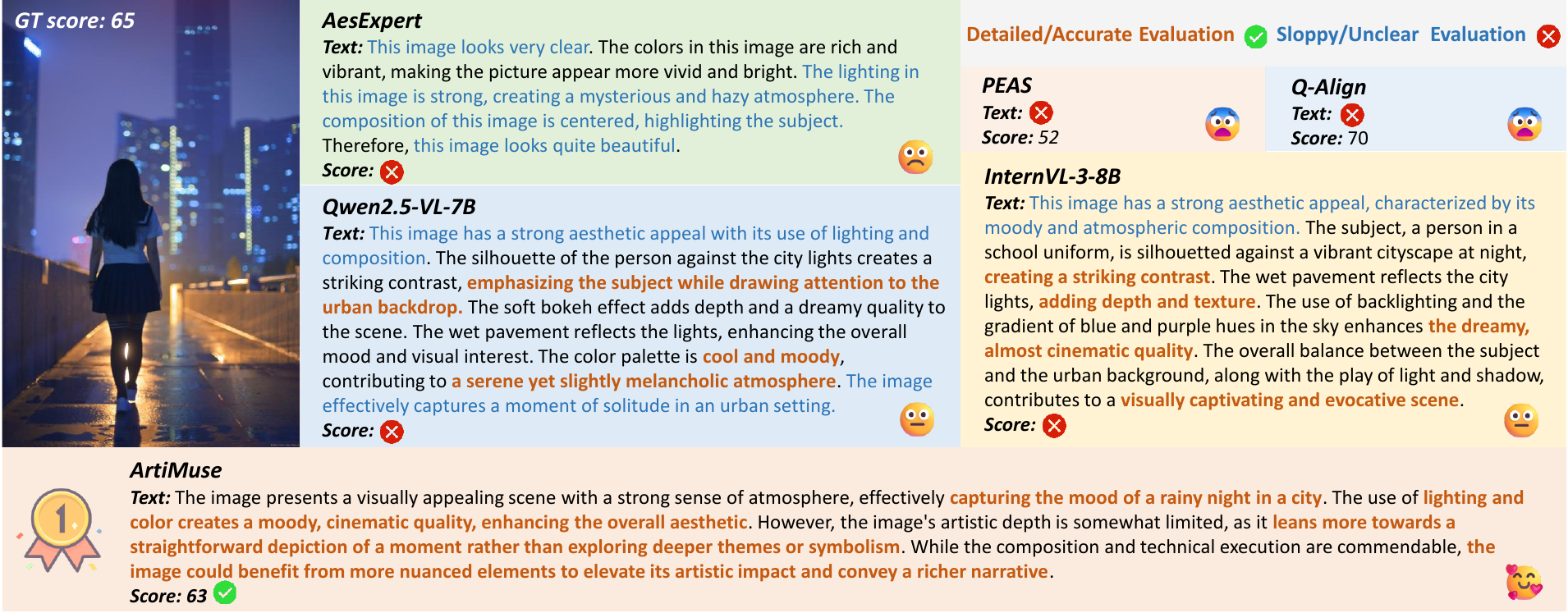}
   \vspace{-10pt}
   \caption{In comparison with existing models, ArtiMuse outperforms them by simultaneously achieving both accurate evaluation and precise aesthetics scoring in multi-dimensional assessments.}
   % \caption{Comparison with existing models. ArtiMuse surpasses other models by simultaneously achieving accurate evaluation and precise aesthetics scoring in multi-dimensional comparisons.}
   \label{fig:intro_comp}
   \vspace{-7pt}
\end{figure}
% 我们的模型可以同时输出Accurate Evaluation和准确的aesthetics score，而其他模型都无法同时完成好这两方面

% Graphic Design, 3D Design, AIGC,}\\\textbf{Photography, Painting \& Calligraphy

% To address these challenges, we introduce \textbf{ArtiMuse}, a professional aesthetic understanding multimodal large language model (MLLM), along with \textbf{ArtiMuse-10K}, a meticulously curated expert-annotated dataset. In collaboration with professional experts with extensive training experience ranging from 3 to over 30 years, we systematically define eight explainable and fine-grained aesthetic attributes, encompassing various visual modalities such as graphic design, 3D design, AIGC images, photography, painting \& calligraphy. ArtiMuse-10K is a large and most comprehensive fine-grained image aesthetics dataset, featuring both \textbf{aesthetics scores} and \textbf{expert-written textual annotations}. Based on large-scale training, we develop ArtiMuse, a powerful aesthetic understanding MLLM capable of jointly generating aesthetics scores and fine-grained expert-level aesthetic analysis, marking a significant step towards a new paradigm of aesthetic AI that moves beyond score prediction to holistic understanding, reasoning, and user-interpretable feedback. Notably, ArtiMuse achieves state-of-the-art (SOTA) performance across multiple widely used public benchmarks, demonstrating its robust generalization and superior aesthetic assessing capabilities.

To address these challenges, we introduce \textbf{ArtiMuse}, a multimodal large language model (MLLM) for professional aesthetic understanding, together with \textbf{ArtiMuse-10K}, a meticulously curated, expert-annotated dataset. Collaborating with domain experts in aesthetics, each with 3 to over 30 years of experience, we systematically define eight explainable and fine-grained aesthetic attributes, covering aspects such as Composition \& Design, Visual Elements \& Structure, and Originality \& Creativity, among others. Based on these attributes, we construct ArtiMuse-10K, the largest and most comprehensive fine-grained image aesthetics dataset to date, featuring both quantitative aesthetic scores and expert-written textual analyses across diverse visual domains, including graphic design, 3D design, AIGC-generated images, photography, and painting \& calligraphy.

Leveraging this dataset, ArtiMuse is trained to jointly predict aesthetic scores and generate expert-level, fine-grained textual feedback, advancing aesthetic AI from mere score prediction toward holistic aesthetic reasoning and user-interpretable analysis. Notably, ArtiMuse achieves state-of-the-art performance across multiple widely used public aesthetics benchmarks, demonstrating its robust generalization ability and superior performance in both quantitative assessment and qualitative explanation, as shown in Fig.~\ref{fig:teaser}.

In addition, a core technical challenge in aesthetics modeling lies in continuous score prediction using MLLMs, which are inherently designed for discrete token generation. Existing methods such as Q-Align~\cite{qalign} attempt to transform continuous scores into discrete ratings, and then reconstruct continuous values by weighted averaging over rating logits. However, this discretization inevitably incurs significant information loss and often leads to inaccurate predictions. To overcome this limitation, we propose a novel \textit{Token As Score} strategy that densely maps predefined discrete tokens to continuous values. Specifically, we utilize existing tokens within the native LLM tokenizer to represent numeric values, thus eliminating the need to expand the vocabulary or retrain the tokenizer. This lightweight yet effective technique enables precise and robust modeling of continuous values within the MLLM framework, substantially improving the fidelity of aesthetics scoring.

Our main contributions can be summarized as follows:

% (1) A systematic framework for image aesthetics assessment (IAA) developed by professional experts, comprising 8 fine-grained aesthetic attributes that comprehensively evaluate various aspects of image aesthetics. This framework enables multi-perspective analysis of aesthetic qualities through carefully designed evaluation metrics.

\textbf{(1) ArtiMuse-10K}, a comprehensive and meticulously annotated image aesthetic assessment dataset containing 10,000 images spanning over 5 main categories and 15 subcategories. Each image is manually annotated by professional experts with detailed textual evaluations across 8 aesthetic attributes, accompanied by an overall aesthetics score. As far as we know, This dataset represents the most extensive expert-curated resource for aesthetics assessment to date.

\textbf{(2) ArtiMuse}, a novel image aesthetics assessment model, is capable of performing fine-grained expert-level textual analysis and providing accurate aesthetic scores. ArtMuse exhibits significantly superior aesthetic assessment expertise and fine-grained analysis compared to other IAA models and general-purpose MLLMs. 
% It is the first IAA model to achieve consistently strong performance across both textual description and numerical scoring tasks.

\textbf{(3) Token As Score}, which enables precise continuous aesthetics scoring in MLLMs by mapping existing tokens to numeric values, avoiding quantization loss and tokenizer changes. It offers a lightweight, effective solution for accurate and stable score prediction.

% \begin{itemize}[noitemsep, left=0pt] % 同时取消缩进和条目间距

% \item A systematic framework for image aesthetics assessment (IAA) developed by professional experts, comprising 8 fine-grained aesthetic attributes that comprehensively evaluate various aspects of image aesthetics. This framework enables multi-perspective analysis of aesthetic qualities through carefully designed evaluation metrics.

% \item ArtiMuse-10K, a comprehensive and meticulously annotated image aesthetic evaluation dataset containing 10,000 images spanning over 5 main categories and 15 subcategories. Each image is manually annotated by professional experts with detailed textual evaluations across 8 aesthetic attributes, accompanied by an overall aesthetic quality score. As far as we know, This dataset represents the most extensive expert-curated resource for aesthetics assessment to date.

% \item ArtiMuse, a novel image aesthetics assessment model capable of performing comprehensive fine-grained text evaluations and accurate scores. 

% \item A complete technical pipeline encompassing the entire workflow from data collection and processing to model training and evaluation. The systematic approach integrates multiple stages including data collection \& processing, annotation generation and model training, providing a reproducible framework for IAA research.

% \end{itemize}

\section{Related Work}
\label{sec:related_work}

\subsection{Multi-modality Large Language Models}
\vspace{-5pt}
% GPT series, Gemini series, Qwen-VL series, InternVL series

% 1. 偏p 2.与人类专家描述风格有gap
% With the development of MLLMs~\cite{gpt4,gemini,qwen2.5,internvl3}, their capabilities have gradually evolved from basic image-text matching to understanding high-level semantic information in images, offering new possibilities for image aesthetics assessment. However, current MLLMs still face clear limitations in this task. They tend to give overly positive evaluations, lacking objective judgment and detailed analysis of image quality. Moreover, the language they generate differs significantly from the professional descriptions used by human experts, making them less suitable for high-quality automated aesthetic evaluation. Therefore, it is necessary to systematically optimize and guide these models through fine-tuning strategies.

With the advancement of MLLMs~\cite{gpt4,gemini,qwen2.5,internvl3}, their ability has expanded from basic image-text matching to understanding high-level semantic content, offering new possibilities for image aesthetics assessment. However, current MLLMs still struggle with objective evaluation, often producing overly positive and superficial judgments. Moreover, the text they generate differs significantly from the professional descriptions used by human experts, making them less suitable for high-quality automated aesthetic evaluation. Therefore, it is necessary to systematically optimize and guide these models through fine-tuning strategies.

\subsection{Image Aesthetics Assessment}
\vspace{-5pt}
% 可以引一下intro部分的数据集对比表格
% 总结:当前的IAA数据集的标注往往过于简单,大部分仅包含常见图像评估属性的程度判定,缺少从美学角度的细致评价
% Although several IAA datasets have been proposed, as shown in Table~\ref{tab:Dataset_Comparison}, they still exhibit notable limitations in terms of annotation dimensions and content diversity. First, regarding score and textual descriptions, some datasets~\cite{tad66k,aadb,para,ava} provide overall aesthetics scores but lack evaluative descriptions of image quality, while others~\cite{artemis,impressions} offer only general comments such as emotional impressions without clear numerical ratings, making it difficult to derive quantitative assessments of image aesthetics. Second, in terms of evaluation dimensions, most datasets~\cite{apddv2,rpcd} only include “overall impression” comments, lacking fine-grained annotations of aesthetic attributes, which limits their utility for more comprehensive aesthetics assessment. Third, in terms of image content diversity, existing datasets~\cite{aadb,para,ava} are mainly focused on photographic images, with only a few~\cite{apddv2,artemis,baid} including artworks, and there is a serious lack of coverage for AIGC-generated images and everyday scenes. Therefore, current IAA datasets are insufficient for supporting high-quality aesthetic modeling across diverse scenarios, highlighting the urgent need for a high-quality benchmark dataset with broader content coverage, complete score annotations, and fine-grained aesthetic attributes.
\noindent\textbf{Datasets.} As summarized in Table~\ref{tab:Dataset_Comparison}, existing IAA datasets suffer from three key limitations: (1) Many~\cite{tad66k,aadb,para,ava} offer overall aesthetics scores but lack detailed evaluative descriptions, while others~\cite{artemis,impressions} provide only vague comments without numerical ratings; (2) Most~\cite{apddv2,rpcd} focus solely on overall impressions, lacking fine-grained aesthetic attribute annotations; (3) In terms of content, datasets~\cite{aadb,para,ava} are mainly photographic, with limited inclusion of artworks~\cite{apddv2,artemis,baid} and little to no AIGC or everyday scene coverage. These gaps hinder comprehensive aesthetic modeling, underscoring the need for a more diverse, well-annotated benchmark.

\noindent\textbf{Models.} IAA models have evolved from simple regression to multimodal generative evaluation with integrated language understanding. Existing approaches fall into two categories: (1) Regression-based models (e.g., TANet~\cite{tad66k}, AesMamba~\cite{AesMamba}) directly predict aesthetics scores from image features but lack interpretability and generalization; (2) MLLM-based generative models leverage vision-language understanding to align better with human perception. Instruction-tuned models~\cite{Q-instruct,yun2024scaling} improve text generation but with limited granularity. AesExpert~\cite{aesexpert} produces expert-style descriptions but lacks score prediction. Q-Align~\cite{qalign} and UNIAA~\cite{Uniaa} combine text and discrete scores, yet lack fine-grained dimension-level evaluation. To overcome these gaps, we introduce ArtiMuse, a unified model that generates expert-level analysis and accurate aesthetics scores.

\section{ArtiMuse-10K Dataset}
\label{sec:dataset}

\subsection{Dataset Overview} 
\vspace{-5pt}
As shown in Tab.~\ref{tab:Dataset_Comparison}, ArtiMuse-10K far exceeds existing IAA datasets in diversity and granularity. It contains 10,000 images across 5 main categories (Design, AIGC, photography, etc.) with 15 fine-grained subcategories. Each image is annotated by professional experts on eight aesthetic attributes and an overall score, offering superior professional rigor and annotation granularity.

% As shown in Tab.~\ref{tab:Dataset_Comparison}, ArtiMuse-10K demonstrates significantly superior diversity and granularity compared to existing IAA datasets. The dataset comprises 10,000 images spanning 5 main categories (design, AIGC, photography, painting, etc.), which are further divided into 15 fine-grained subcategories. Each image is meticulously annotated by experts on eight distinct aesthetic attributes plus an overall aesthetics score, making ArtiMuse-10K far exceed current IAA datasets in both professional rigor and annotation granularity.

% 如Tab.1所示，相较于之前的IAA数据集，ArtiMuse-10K展现出了非常优越的多样性和细致程度。图像来自包含disign、AIGC、photography、painting等的五个大类，并且分为多达15个subcategories，总数量达到10000张。我们采用纯expert手工标注，每张图像都标注了8 aesthetic attributes的评价和一个overall aesthetics score，在专业性和细致程度上都远远超过现有的IAA数据集。

\begin{table}[ht]
\centering
\setlength{\tabcolsep}{2pt}     
\vspace{-7pt}
\caption{A Comparison between ArtiMuse-10K dataset and existing IAA datasets.}
\vspace{-5pt}
\label{tab:Dataset_Comparison}
\resizebox{\textwidth}{!}{%
\begin{tabular}{l c c c c c c c c}
\toprule[1.5pt]
\textbf{Dataset} & \textbf{Main Categories} & \textbf{Subcategories} & \textbf{\# Image} & \textbf{Score} & \textbf{Text Caption} & \textbf{\# Attribute}  & \textbf{Attribute Categories} &  \textbf{Annotators} \\ 
\midrule
AVA~\cite{ava} & Photography & -- & 255,528  & \textcolor{green}{\faCheck} & \textcolor{red}{\faTimes} & -- & -- & Non-Experts   \\ 
AADB~\cite{aadb} & Photography & -- & 10,000  & \textcolor{green}{\faCheck} & \textcolor{red}{\faTimes} & -- & -- & Non-Experts \\
FLICKR-AES~\cite{flickr} & Photography & 9 Categories & 40,499  & \textcolor{green}{\faCheck} & \textcolor{red}{\faTimes} & -- & -- & Non-Experts   \\ 
SPAQ~\cite{spaq} & Photography & -- & 111,125  & \textcolor{green}{\faCheck} & \textcolor{red}{\faTimes} & -- & -- & Non-Experts   \\ 
KonIQ-10K~\cite{koniq} & Photography & -- & 10,073  & \textcolor{green}{\faCheck} & \textcolor{red}{\faTimes} & -- & -- & Non-Experts   \\ 
ArtEmis~\cite{artemis} & Painting & -- & 81,446  & \textcolor{red}{\faTimes} & \textcolor{green}{\faCheck} & 1 Attribute & Emotional Analysis & Non-Experts \\
RPCD~\cite{rpcd} & Photography & -- & 73,965 & \textcolor{green}{\faCheck} & \textcolor{green}{\faCheck} & 1 Attribute & Overall Comment & Non-Experts \\
PARA~\cite{para} & Photography & -- & 31,229  & \textcolor{green}{\faCheck} & \textcolor{red}{\faTimes} & -- & -- & Non-Experts  \\
TAD66K~\cite{tad66k} & Painting, Photography  & -- & 66,000  & \textcolor{green}{\faCheck} & \textcolor{red}{\faTimes} & -- & -- & Non-Experts   \\
Impressions~\cite{impressions} & Photography & -- & 1,440  & \textcolor{red}{\faTimes} & \textcolor{green}{\faCheck} & 3 Attributes & \makecell{Description, Perception,\\  Evaluation} & Non-Experts \\
BAID~\cite{baid} & Painting & -- & 60,337 & \textcolor{green}{\faCheck} & \textcolor{red}{\faTimes} & -- & -- & Non-Experts \\
APDDv2~\cite{apddv2} & Painting & \makecell{3 Categories} & 10,023  & \textcolor{green}{\faCheck} & \textcolor{green}{\faCheck} & 1 Attribute & Overall Comment & Professional Experts \\

\midrule
\makecell[l]{\textbf{ArtiMuse-10K} \\ \textbf{(Ours)}} & \makecell{\textbf{Graphic Design, 3D Design, }\\ \textbf{AIGC, Photography,} \\ \textbf{Painting \& Calligraphy}} & \makecell{\textbf{15 Detailed}\\ \textbf{Categories}} & \textbf{10,000} & \textbf{\textcolor{green}{\faCheck}} & \textbf{\textcolor{green}{\faCheck}} & \textbf{8 Attributes} & \makecell{\textbf{Fine-grained Attributes}\\ \textbf{(Composition \& Design,} \\ \textbf{Technical Execution, etc.)}} & \textbf{Professional Experts} \\
\bottomrule[1.5pt]
\end{tabular}}
\vspace{-10pt}
\end{table}
% As shown in Tab.~\ref{tab:Dataset_Comparison}, ArtiMuse-10K dataset相较于以往的IAA数据集更加细致和专业。在数据来源上，ArtiMuse

\subsection{Image Collection}
\vspace{-5pt}
% Prior studies~\cite{apddv2, qalign, aesexpert} have emphasized the critical importance of collecting diverse image types and expanding data domain coverage for enhancing dataset quality. Our ArtiMuse-10K dataset comprises 10,000 images systematically collected from five main categories: Graphic Design, 3D Design, AI-Generated Content (AIGC), Photography, and Painting \& Calligraphy. These images are further devided into 15 distinct subcategories including Chinese Painting, Sculpture, Architecture, Daily Photography, among others, ensuring comprehensive coverage of artistic expressions. 

% , cao2024grids, cao2025dualx, li2025diffvsr

\begin{figure}[!h]
    \centering
    \vspace{-10pt}
    % 左侧 example 图
    \begin{minipage}{.49\textwidth}
        \centering
        \includegraphics[width=\textwidth]{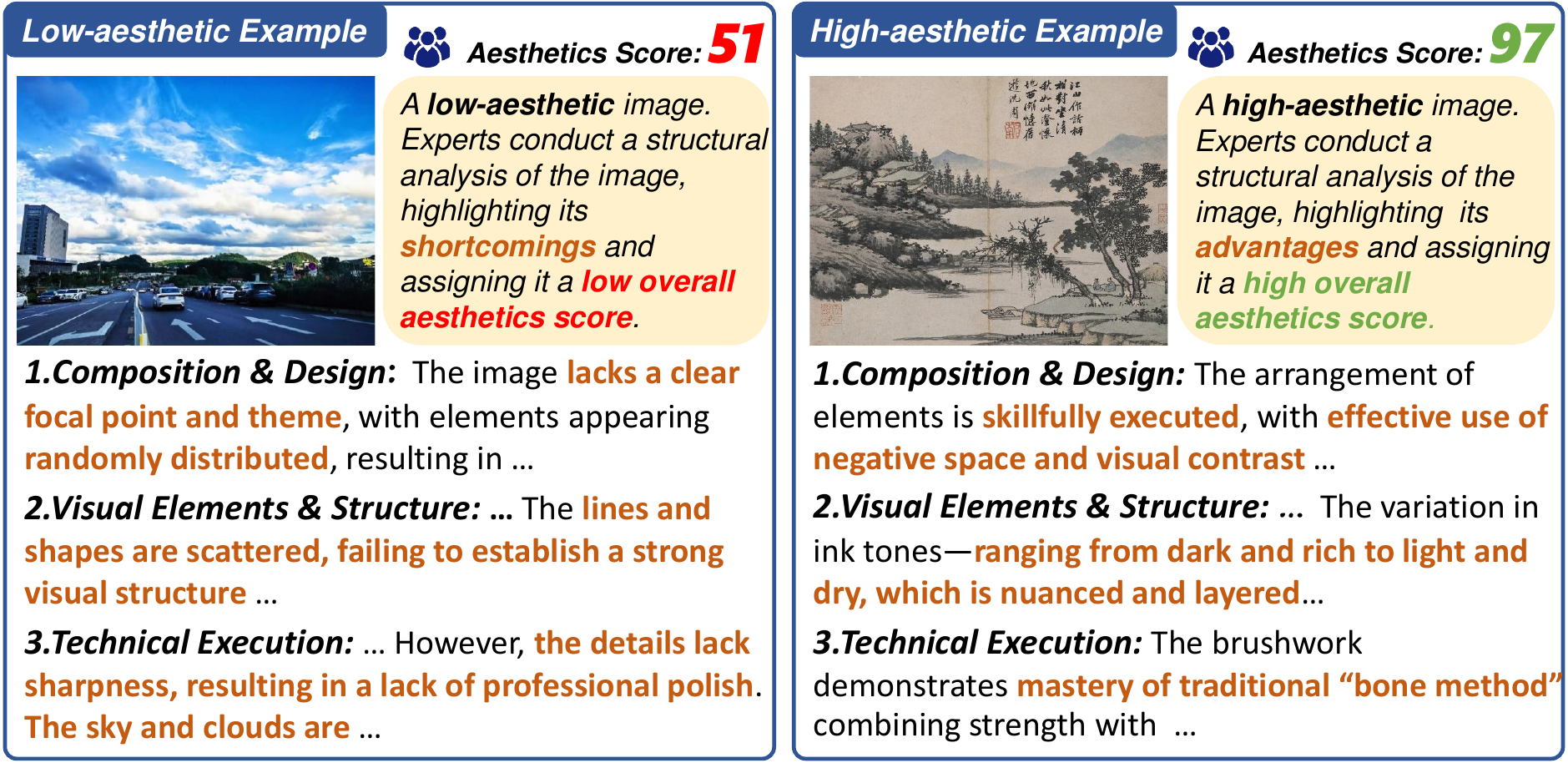}
        \vspace{-15pt}
        \caption{Data examples in ArtiMuse-10K.}
        \label{fig:dataset_example}
        \vspace{-5pt}
    \end{minipage}%
    \hfill
    \hspace{0.5pt}% ← 在这里增加 1em 宽度的空白
    \begin{minipage}{.49\textwidth}
        \centering
        \includegraphics[width=\textwidth]{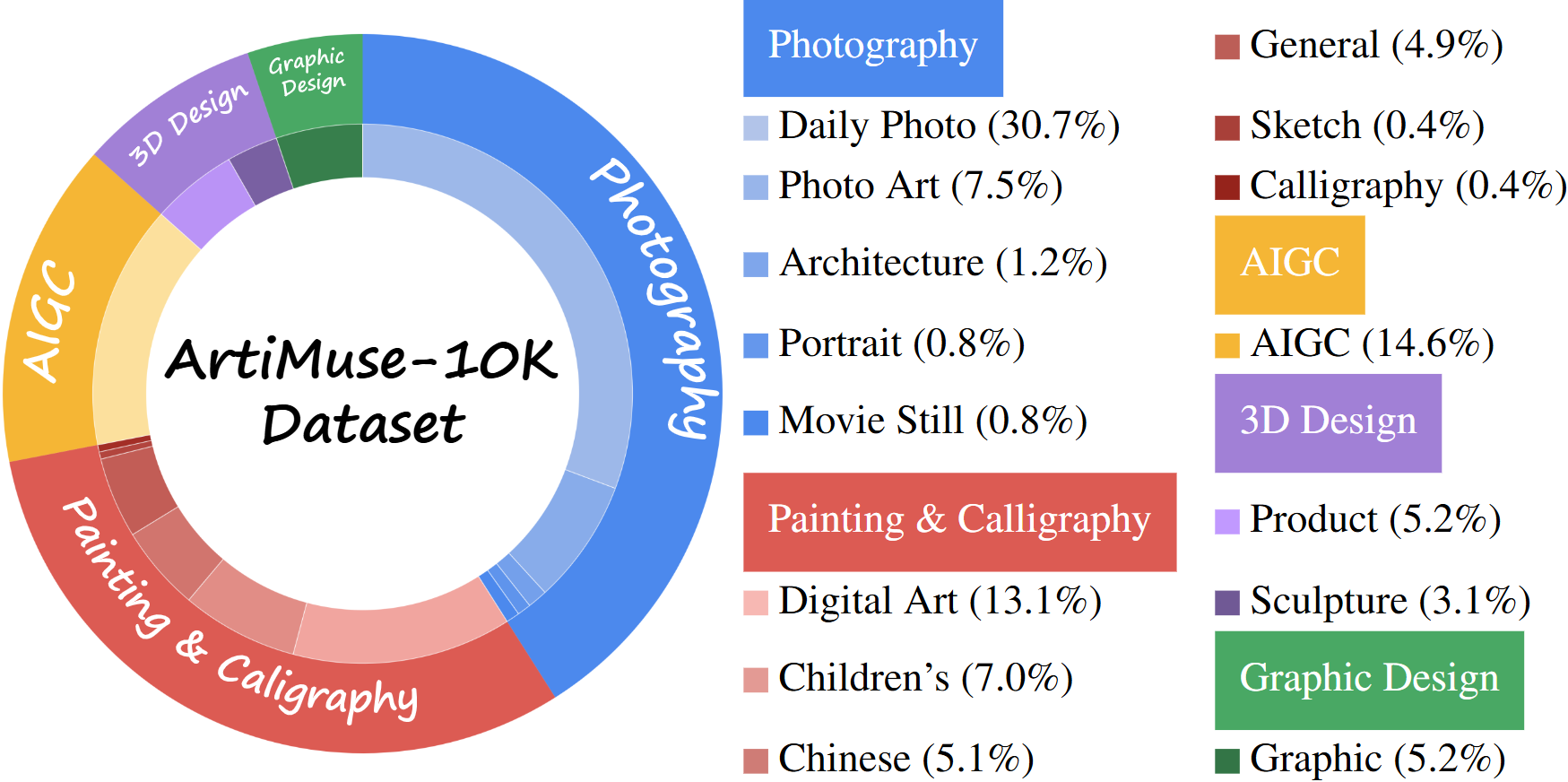}
        \vspace{-15pt}
        \caption{Composition of ArtiMuse-10K.}
        \label{fig:dataset_composition}
        \vspace{-5pt}
    \end{minipage}
    \vspace{-11pt}
    % \caption{ArtiMuse-10K Dataset.}
    % \label{fig:photosis_dataset_composition}
\end{figure}

Previous studies~\cite{apddv2, qalign, aesexpert, cao2024grids, li2025diffvsr, cao2025dualxvsrdualaxialspatialtimestemporal} have emphasized the importance of ensuring dataset diversity and extending domain coverage to enhance the quality and robustness of aesthetic assessment models. Building upon these insights, we construct ArtiMuse-10K, a high-quality dataset comprising 10,000 carefully curated images spanning five primary categories: Graphic Design, 3D Design, AIGC-generated images, Photography, and Painting \& Calligraphy. These categories are subdivided into 15 distinct subcategories, such as Chinese Painting, Sculpture, and Daily Photography, ensuring comprehensive representation of diverse artistic expressions. The internal data samples and overall dataset composition are illustrated in Fig.~\ref{fig:dataset_example} and Fig.~\ref{fig:dataset_composition}, respectively.

\noindent\textbf{Non-AIGC Images.} For non-AIGC images, we collaborate with domain experts to curate professionally created artworks sourced from academic settings, including student assignments and competition entries. To ensure the dataset reflects contemporary trends, we also collect a wide range of artistic and photographic works from reputable online art and photography platforms.

\noindent\textbf{AIGC Images.} We utilize state-of-the-art generative models (Stable Diffusion series~\cite{sd}, Dreamlike Photoreal 2.0~\cite{dreamlike_photoreal_2023}, FLUX~\cite{flux}, etc.) to systematically produce synthetic images. We further augment this core dataset with open-source community contributions produced using comparable architectures.
% This core synthetic dataset is augmented with open-source community contributions generated using equivalent architectures, ensuring diversity in style and technical implementation.

% 很多previous work都指出，需要广泛收集各种类型的图像、扩大数据集包含的数据域对于提高数据集质量至关重要。我们从包含了艺术作品、摄影作品、AIGC图像等类型图像的各类数据源中收集了32000张图像，并最终筛选出16967张图像构成ArtiMuse Dataset。ArtiMuse Dataset的数据构成如图1所示，数据包含4个大类，几乎涵盖了有美学评估价值的所有图像。此外每个大类的图像也有多个细分小类组成，共包含35个小类，充分展示了数据来源的丰富程度。

% Prior studies~\cite{apddv2, qalign, aesexpert} have emphasized the critical importance of collecting diverse image types and expanding data domain coverage for enhancing dataset quality. Through comprehensive curation from multiple sources including artistic works, professional photography, and AI-generated content (AIGC) images, we initially gathered 32,000 candidate images. Following rigorous quality screening and deduplication processes, we ultimately selected 16,967 high-quality images to form the ArtiMuse Dataset. As illustrated in Figure 1, our dataset architecture comprises 4 major categories that comprehensively cover all image types with recognized aesthetic evaluation value. These categories are further subdivided into 35 fine-grained subclasses, systematically demonstrating the dataset's exceptional diversity across multiple dimensions.

\definecolor{PhotographyMain}{RGB}{76,137,237}  % #4C89ED
\definecolor{AIGCMain}{RGB}{245,182,53}        % #F5B635
\definecolor{PaintingMain}{RGB}{220,91,83}     % #DC5B53
\definecolor{GraphicDesignMain}{RGB}{73,168,100} % #49A864
\definecolor{3DDesignMain}{RGB}{161,128,214}   % #A180D6
\definecolor{DailyPhoto}{RGB}{178,196,233}  % #b2c4e9
\definecolor{AIGC}{RGB}{245,182,53}         % #f5b635
\definecolor{DigitalArt}{RGB}{247,184,179}  % #f7b8b3
\definecolor{PhotographicArt}{RGB}{152,181,234} % #98b5ea
\definecolor{ChildrensPainting}{RGB}{227,154,148} % #e39a94
\definecolor{GraphicDesign}{RGB}{51,117,70}  % #337546
\definecolor{ProductDesign}{RGB}{193,153,255} % #c199ff
\definecolor{ChinesePainting}{RGB}{207,124,118} % #cf7c76
\definecolor{GeneralPainting}{RGB}{188,94,87} % #bc5e57
\definecolor{Sculpture}{RGB}{112,89,149}    % #705995
\definecolor{Architecture}{RGB}{127,166,235} % #7fa6eb
\definecolor{Portrait}{RGB}{101,151,236}    % #6597ec
\definecolor{MovieStill}{RGB}{76,137,237}   % #4c89ed
\definecolor{Sketch}{RGB}{168,64,57}        % #a84039
\definecolor{Calligraphy}{RGB}{149,35,27}   % #95231b

\subsection{Aesthetic Attributes}
\vspace{-5pt}
 % as presented in Tab.~\ref{tab:attributes}
%% 要建立一个细粒度标注的美学评估数据集，首先需要制定一套完善的细粒度评估标准。我们与美术领域专家就当前图像美学质量的细分维度进行了讨论，并最终提出了如表格1所示的美学评估体系。这一体系中共有7个细分美学属性和一个总体评价，从构图、设计、视觉元素、技术执行、创意、主题、情绪等方面详细定义了图像的各个美学维度。这套评价体系不依赖于图像类型和图像具体内容，无论是艺术绘画、摄影还是AIGC图像均可以使用。
To establish a fine-grained annotated dataset for image aesthetics assessment, the primary task involves developing a comprehensive assessment system. Through systematic consultations with artistic experts, we have formulated a novel aesthetic assessment system. This system comprises 8 specific aesthetic attributes and an overall aesthetics score, systematically defining key dimensions of image aesthetics including Composition \& Design, Visual Elements \& Structure, Technical Execution, Originality \& Creativity, Theme \& Communication, Emotion \& Viewer Response, Overall Gestalt and Comprehensive Evaluation. Notably, our system is content-agnostic and universally applicable to image types from natural to AIGC.

\subsection{Human Annotations}
\vspace{-5pt}
% 在完成数据的收集和Aesthetic attributes的定义后，我们开始进行数据的标注。如今的MLLM虽然已经非常强大，拥有进行美学评估的相应知识和判别能力，但是进行评估时仍然容易出现各种错误，例如无论图像美学质量好坏总是倾向于输出正面评价等。为了确保数据的准确性，我们邀请了大量美术领域专家对数据集按照定义好的Aesthetic attributes进行标注，并且还对整体美学质量进行了评分。最终每张图像得到了共计9条高质量人工标注（包含美学评分在内），整个数据集共计152703条高质量人工标注。

\begin{figure}[t]
  \centering
    \includegraphics[width=0.99\linewidth]{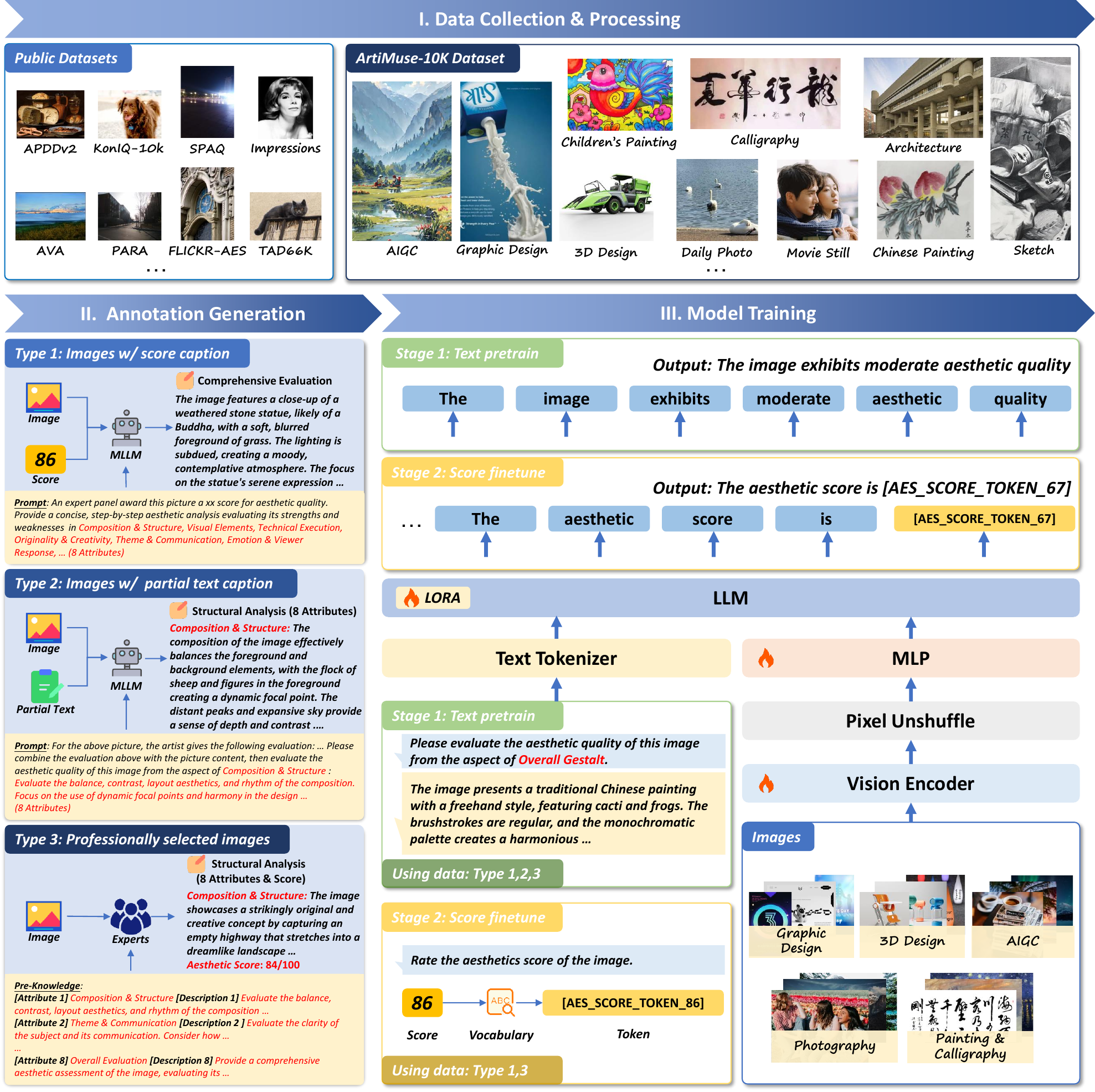}
   \vspace{-10pt}
   \caption{Overview of ArtiMuse. ArtiMuse encompasses a multi-stage pipeline spanning data collection \& processing, annotation generation, and model training, systematically enhancing its text evaluation capabilities and score assessment proficiency across multiple dimensions. }
   \label{fig:pipeline}
   \vspace{-15pt}
\end{figure}

Based on the predefined aesthetic attributes, we invite professional experts to meticulously annotate images in the ArtiMuse-10K dataset. We collaborate with domain experts whose professional experience spans a broad spectrum, ranging from at least three years to over three decades, including distinguished authorities in the field. The entire annotation process is illustrated in Fig.~\ref{fig:pipeline} as Type 3: Professionally Selected Images. Each image in ArtiMuse-10K is ultimately annotated with textual analysis describing eight distinct aesthetic attributes and an overall aesthetics score. Our comprehensive annotation framework enhances dataset quality and model performance by integrating multi-dimensional aesthetic attributes for fine-grained visual analysis, expert-curated scores for reliable aesthetic assessment, and rich semantic annotations for improving training  robustness.
% This comprehensive annotation framework significantly enhances both dataset quality and model capability through three key advantages: (1) the multi-dimensional aesthetic attributes enable fine-grained visual analysis, (2) the expert-curated scores provide reliable aesthetics assessments, and (3) the rich semantic annotations facilitate more accurate and robust training across.

% Following data collection and operational definition of aesthetic attributes, we implemented a rigorous multi-stage annotation protocol to ensure assessment reliability. While modern multimodal large language models (MLLMs) demonstrate remarkable aesthetic discernment capabilities, our preliminary analysis revealed systematic biases in automated evaluations, particularly an optimistic rating tendency regardless of actual aesthetic merit (see in Appendix). To establish ground truth annotations, we engaged domain experts from fine arts and computational aesthetics fields through a three-phase process: 1) Attribute-specific annotation using our defined taxonomy, 2) Holistic quality scoring (1-10 scale), and 3) Cross-validation rounds to resolve discrepancies. Each image received 9 independent human evaluations (comprising both attribute assessments and composite scores), yielding 152,703 validated annotations across the dataset - representing an unprecedented annotation density in aesthetic computation research.

% 补一个说明MLLM美学评价犯错误的图？ 放到supp

\vspace{-5pt}
\section{Methodology}
\label{sec:methodology}
\vspace{-5pt}
% \subsection{Overview}
% \vspace{-5pt}
% % 我们采用InternVL-3作为base model，
% We propose a systematic pipeline encompassing three key components: Data Collection \& Processing, Annotation Generation, and Model Training. By integrating these components, ArtiMuse is able to simultaneously generate high-quality textual critiques and produce accurate aesthetics scores, effectively bridging the gap between qualitative artistic evaluation and quantitative aesthetic measurement. 

% The process is divided into three main stages: Data Collection \& Processing, where the raw image data is gathered and refined; Annotation Generation, where detailed textual and score-based annotations are created for the images; and Model Training, which utilizes the prepared data and annotations through a multi-stage strategy involving large language models and vision encoders to build the final aesthetic evaluation capability.

\subsection{Dataset Collection \& Processing}
\vspace{-5pt}
% 更丰富的数据源和更细致的人工标注对于数据集的质量至关重要.因此在ArtiMuse-10K dataset之外,我们还从现有的公开数据集中筛选出了一批具有较高质量标注信息的数据.筛选时我们主要关注数据集的标注是否有和美学相关的有价值标注,例如美学评分,文字分析,和美学相关的属性tag等.我们共从超过8个公开数据集筛选了超过350,000张图像.更详细的细节信息参见Supp..

Richer data sources and more meticulous manual annotations are crucial for enhancing dataset quality. In addition to the ArtiMuse-10K dataset, we carefully curate over 350,000 high-quality annotated images from existing datasets, including APDDv2~\cite{apddv2}, PARA~\cite{para}, Impressions~\cite{impressions} and so on. 

% Additional details are in the Supp

\noindent\textbf{Aesthetic Caption Quality.} 
% 我们收集数据时特别关注了数据集本身的Aesthetic annotation quality, Our selection criteria primarily focus on whether the datasets contain valuable aesthetic-related annotations, such as aesthetics scores, textual analyses, and aesthetic attribute tags. 这些信息都会在Annotation Generation部分被利用以提升数据集的质量.
We place particular emphasis on the aesthetic caption quality. Our selection criteria prioritize datasets that include valuable aesthetic-related captions such as aesthetics scores, comprehensive textual analyses, and aesthetic attribute tags. These captions are subsequently utilized in the Annotation Generation phase to systematically enhance dataset quality.

\noindent\textbf{Aesthetic Quality Diversity.} 
Our collection specifically incorporates images with varying aesthetic qualities, including intentionally retained lower-quality samples, to address both dataset diversity requirements and mitigate the prevalent preference bias observed in contemporary LLMs. This carefully balanced composition strategy enhances model training through controlled inclusion of suboptimal visual materials, thereby improving discriminative capabilities in aesthetic assessment.
% This foundational stage involves assembling the image data that will be used throughout the pipeline. The process draws from two primary sources. Firstly, various public datasets known to contain aesthetic information, such as AVA, PARA, Flickr-AES, TAD66K, PVPL, KonIQ-10k, SPAQ, and Impressions, are incorporated. Secondly, a specialized PhotolysisAesthetic Dataset is utilized, which includes a diverse range of content categories like Calligraphy, Modern painting, Buildings, Chinese painting, AIGC, Portrait, Crayon, Engraving, and Oil painting. After collecting data from these sources, a crucial Filtering step is applied. This processing step is essential for refining the raw data, ensuring its quality and relevance before it proceeds to the subsequent stages, thereby yielding the "Raw Data" used for annotation generation.
\vspace{-3pt}
\subsection{Annotation Generation}
\vspace{-5pt}
% The Annotation Generation stage focuses on enriching the dataset with detailed descriptive and evaluative information. This involves creating different types of annotations tailored to the nature of the available information for each image. 
The Annotation Generation stage aims to enrich the dataset with detailed descriptive and evaluative annotations, illustrated in Fig.~\ref{fig:pipeline}. This process involves creating distinct annotation types based on the available information for each image. \textbf{Type 1:} For images with only score caption, we leverage this global quality assessment to generate holistic analyses. We design a prompt to guide the MLLM in producing a comprehensive evaluation based on predefined aesthetic attributes, while incorporating both the score and visual input. \textbf{Type 2:} For images with partial text captions containing specific aesthetic descriptions, we employ a prompt to instruct the MLLM to generate fine-grained evaluations. For each image, the model produces a structural analysis across 8 aesthetic attributes, utilizing both the textual and visual inputs. \textbf{Type 3:} For professionally selected images, we engage experts to conduct structural analysis based on pre-defined aesthetic attributes, along with providing an overall aesthetics score. More details are in the Supp.

\noindent\textbf{Importance of Manual Annotations.} 
Although MLLMs demonstrate strong aesthetic evaluation capabilities, our empirical analysis reveals a systematic bias: they tend to generate overwhelmingly positive assessments regardless of the actual image quality, as shown in Fig.~\ref{fig:text_qualitative_combined}. This positivity bias leads to annotations that poorly reflect true aesthetic merit. To address this limitation, we incorporate professional human evaluations to provide balanced and reliable ground-truth annotations.
% While modern MLLMs demonstrate remarkable aesthetic discernment capabilities, our analysis revealed systematic biases in automated evaluations, particularly an optimistic rating tendency regardless of actual aesthetic merit, as shown in Fig~\ref{fig:text_qualitative}.
% While MLLMs demonstrate remarkable aesthetic discernment capabilities, our empirical analysis reveals that MLLMs sometimes provide annotations inadequately correlate with actual aesthetic merit as illustrated in Fig~\ref{fig:text_qualitative}. To address this limitation, we invite professional experts to evaluate and annotate images, ensuring more reliable ground-truth references.

% Although MLLMs exhibit impressive aesthetic judgment capabilities, our empirical analysis demonstrates that their annotations occasionally show limited correlation with genuine aesthetic quality, as evidenced in Fig~\ref{fig:text_qualitative}. To overcome this limitation, we employ professional experts to assess and annotate images, thereby establishing more accurate and reliable ground-truth references.

\subsection{Training Strategy}
\label{Sec:training_strategy}
\vspace{-5pt}
ArtiMuse is built on InternVL-3-8B~\cite{internvl3}. We modify the dynamic resolution strategy to a fixed-resolution approach while retaining the remaining components. The training process consists of two distinct phases: text pretraining and score fine-tuning, as illustrated in Fig.~\ref{fig:pipeline}. In both stages, we jointly train the vision encoder, MLP, and LLM components, with the LLM undergoing LoRA-based fine-tuning. The ArtiMuse uses common GPT loss~\cite{gptloss}, i.e. minimizing the cross-entropy loss between the predicted logits and target tokens.

% The final stage outlines the multi-faceted training approach employed to develop the aesthetic evaluation model. This strategy involves several sequential steps designed to build different capabilities. Initially, a Stage 1 Text Pretrain focuses solely on language understanding, training a Large Language Model (LLM), potentially with LORA, on text inputs processed by a Text Tokenizer. This stage aims to enable the model to process and generate coherent text, as exemplified by generating descriptive phrases about image quality. Building upon this, Stage 2 Score Pretrain integrates aesthetics scores by training the model to associate specific text prompts with numerical aesthetics scores represented as tokens, like \verb|[Aes_Score_Token_86]|. Stage 3 then introduces visual information through Text Pretrain with images. In this step, a Vision Encoder processes image data (including Artworks, Photography, and AIGC), potentially after a Pixel Unshuffle operation, and these visual features are combined with text inputs via an MLP before being fed to the LLM. This teaches the model to generate text descriptions and evaluations based on image content. The final Stage 4, Score Pretrain with images, combines all elements: the model takes an image and a text prompt asking for an aesthetics score as input and is trained to predict the correct aesthetics score token, leveraging its learned understanding of both language and visual aesthetics to provide a final evaluation score.
% resolution strategy\

\noindent\textbf{Text Pretrain.} 
The text pretraining phase utilizes our complete collected image dataset, where each image is paired with its corresponding aesthetic analysis caption generated during the annotation generation stage. This phase aims to equip the model with accurate structural aesthetic analysis capabilities while largely preserving the MLLM's pretrained knowledge. To achieve this balance, we apply LoRA fine-tuning specifically to the LLM component.

\noindent\textbf{Score Finetune.} 
After establishing foundational aesthetic understanding through pretraining, we proceed to score fine-tuning. In this phase, we convert each image's overall aesthetics score into a specialized scoring token designed exclusively for aesthetics scoring, which then serves as the training caption. Inspired by previous works~\cite{qalign, Uniaa, nexttoken}, we propose a novel score prediction strategy called \textit{Token As Score}, which eliminates the need for vocabulary expansion or tokenizer retraining. Specifically, we designate 101 existing tokens as \verb|[Aes_Score_Token]|s, each corresponding to integer scores ranging from 0 to 100. We select tokens that are concise and inherently carry ordinal semantic information from the vocabulary. In our implementation, we employ twin-letter combinations as tokens (e.g., Score 1 is represented as \verb|[Aes_Score_Token_1]|, where the actual token is \verb|ab|. See supp. for more details). During data preprocessing, we first normalize aesthetics scores to the [0,100] range and then map them to their corresponding tokens. This methodology enables the construction of training data where continuous scores are discretized into token representations. The model is subsequently fine-tuned to predict these discrete tokens. During inference, we convert the predicted tokens back to their numerical values, and the final aesthetics score is derived by computing the expectation over the probability distribution of all possible score tokens. Specifically, we denote $l_i$ and $p_i$ for logits and probability of \verb|[Aes_Score_Token_i]|, the final aesthetics score $S_{\text{Aes}}$ is compute as: 
$ S_{\text{Aes}} = \sum_{i=0}^{100} i \times p_{i}  = \sum_{i=0}^{100} i \times \frac{e^{l_i}}{\sum_{j=0}^{100} e^{l_j}} $.
% Inspired by previous works~\cite{qalign, Uniaa, nexttoken}, we propose a score prediction strategy namely \textit{Token As Score}. Specifically,  we predefine 101 distinct \verb|[Aes_Score_Token]|s corresponding to integer scores from 0 to 100. During data preparation, we first normalize the aesthetics scores to the [0,100] range and map them to their respective tokens (e.g., a score of 86 becomes \verb|[Aes_Score_Token_86]|). This approach allows us to construct training data where numerical scores are replaced by their tokenized representations. The model is then fine-tuned to predict these discrete tokens. For inference, we map the predicted tokens back to their numerical values. The final aesthetics score is computed based on the probability distribution over all possible scoring tokens. Specifically, we denote $l_i$ and $p_i$ for logits and probability of \verb|[Aes_Score_Token_i]|, the final aesthetics score $S_{\text{Aes}}$ is compute as: 
% $ S_{\text{Aes}} = \sum_{i=0}^{100} i \times p_{i}  = \sum_{i=0}^{100} i \times \frac{e^{l_i}}{\sum_{j=0}^{100} e^{l_j}} $.
\vspace{-5pt}
\begin{figure}[h]
  \centering
    \includegraphics[width=0.99\linewidth]{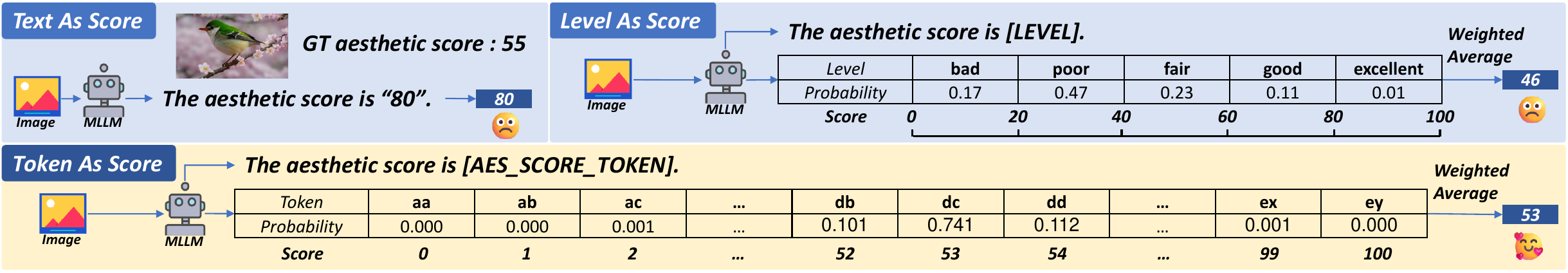}
   \vspace{-10pt}
   \caption{Comparison of score prediction methods. Token As Score features a more rational design and delivers more precise results. }
   \label{fig:token_as_score}
   \vspace{-15pt}
\end{figure}

% \begin{equation}
%     S_{\text{Aes}} = \sum_{i=0}^{100} i \times p_{i}  = \sum_{i=0}^{100} i \times \frac{e^{l_i}}{\sum_{j=0}^{100} e^{l_j}}
% \end{equation}

% S_{\text{Aes}} = \sum_{i=0}^{100} i \times p_{i}  = \sum_{i=0}^{100} i \times \frac{e^{l_i}}{\sum_{j=0}^{100} e^{l_j}}
% $$

\noindent\textbf{Why Token As Score?} Current approaches for scoring with MLLMs primarily fall into two categories: (1) directly prompting the LLM to output scores as text (\textit{Text As Score}), or (2) predefining discrete levels corresponding to specific score intervals and computing the final score based on the model's predicted token distribution (\textit{Level As Score}). Previous works~\cite{qalign, nexttoken, Uniaa} demonstrate that directly generating scores as text leads to severe hallucination issues. Thus, we adopt the Token As Score approach and investigate the impact of token granularity on model performance. A comparison of these score prediction methods is shown in Fig.~\ref{fig:token_as_score}. Further experiments in Tab.~\ref{tab:ablation-study} show that 100 aesthetics score tokens achieve optimal results.

\noindent\textbf{Maintaining Text Ability.} 
A widely recognized challenge in IAA and IQA tasks is that MLLMs often struggle to simultaneously preserve their textual understanding and scoring capabilities~\cite{depictqa_v1, deqa_score}. Since the training data in the score fine-tuning phase is significantly more monotonous than in text pretraining, full fine-tuning of the LLM can easily degrade its structural aesthetic analysis ability. To mitigate this issue while maintaining textual proficiency, we employ LoRA-based fine-tuning for the LLM, enabling the model to retain both its linguistic and scoring capabilities.
% 放个对比图展示如何保留annotation的能力

\vspace{-5pt}
\section{Experiments}
\label{sec:Experiments}
\vspace{-5pt}
\subsection{Implementation Details} 
\vspace{-5pt}
In our experiments, we adopt InternVL-3-8B~\cite{internvl3} as the base model initialized with its pretrained weights. During text pretraining, we implement a batch size of 128 and learning rate of $4e-5$ with a cosine annealing schedule~\cite{cosine}, training for one epoch to balance convergence with prior knowledge preservation. For the score fine-tuning, we maintain the batch size at 128 while adjusting the learning rate to $2e-5$ across 2 training epochs. We maintain identical configurations across all experiments, with all training conducted on 4 * NVIDIA A100 80GB GPUs. The text pretraining phase typically takes 5 hours, while the score fine-tuning duration varies between 10 minutes to 4 hours depending on dataset size, demonstrating efficient convergence across different scales.

% \noindent\textbf{Efficiency Analysis.}

% \subsection{Aesthetics Assessment Performance}

\subsection{Structural Aesthetic Analysis}
\label{Sec:structural_aesthetic_analysis}
\vspace{-5pt}
\noindent\textbf{Judging by MLLM.} To evaluate current models' ability of structural aesthetic analysis, we design a judgement framework leveraging the superior comprehension power of MLLM. An image is presented to both experts and various models to generate aesthetic analysis on 8 aesthetics attributes. Then a judging MLLM selects which model performs best across each attribute, using the human expert's description as a reference. The results in Tab.~\ref{tab:image-evaluation} show ArtiMuse outperforms other models across 8 aesthetic attributes, demonstrating superior structural aesthetic analysis capability.
% The final results is shown in Tab.~\ref{tab:image-evaluation}, indicating that ArtiMuse achieves the highest proportion of best performances across different attributes, demonstrating its superior capability in structural aesthetic analysis.

% 此外,我们还做了一个user study,邀请volunteers投票选出他们认为输出文本质量更好的模型,每个模型被选择的比例展示在Tab.~\ref{tab:image-evaluation}的Human Rate中,我们的模型在user preference方面也有着显著的优势.

% \begin{figure}[h]
%   \centering
%     \includegraphics[width=0.99\linewidth]{figures/5.2.1.pdf}
%    \vspace{-10pt}
%    \caption{Overview of the structural aesthetic analysis ability judgment pipeline.}
%    \label{fig:Aesthetic Annotation}
%    \vspace{-5pt}
% \end{figure}

\noindent\textbf{Judging by Human.} In addition, we conduct a user study where participants are asked to compare and vote for the model they perceive as producing higher-quality aesthetic analysis. The proportion of selections for each model, presented as Human Rate in Table~\ref{tab:image-evaluation}, demonstrates that our approach achieves a significantly higher preference rate compared to other methods.
\vspace{-10pt}
% 不同模型被选取率的统计.The first 8 aesthetic attributes are judged by Gemini-2.0-flash, while Human Rate is 由人类志愿者选取.
\begin{table}[htbp]
\centering
\caption{The selection rates of different models. For the first 8 aesthetic attributes, evaluations are performed by Gemini-2.0-flash, while Human Rate is provided by volunteer participants.}
\vspace{-5pt}
\label{tab:image-evaluation}
\resizebox{\textwidth}{!}{%
\fontsize{6}{6}\selectfont
\renewcommand{\arraystretch}{0.8} % 调整行距，值越小行距越窄
\begin{tabular}{l|c c c c }
\toprule[1pt]
Aesthetic Attributes & AesExpert~\cite{aesexpert} & Qwen-2.5-VL-7B~\cite{qwen2.5}  & InternVL3-8B~\cite{internvl3} & ArtiMuse \\
\midrule
1. Composition \& Design & 0.0\% & 12.7\% & 10.4\% & \textbf{76.9\%} \\
2. Visual Elements \& Structure & 0.0\% & 19.3\% & 16.5\% & \textbf{64.2\%} \\
3. Technical Execution & 0.0\% & 9.9\% & 10.4\% & \textbf{79.7\%} \\
4. Originality \& Creativity & 0.0\% & 13.7\% & 8.5\% & \textbf{77.8\%}\\
5. Theme \& Communication & 0.9\% & 17.5\% & 24.1\% & \textbf{58.5\%} \\
6. Emotion \& Viewer Response & 0.0\% & 17.5\% & 24.1\% & \textbf{58.5\%} \\
7. Overall Gestalt & 0.0\% & 14.6\% & 9.4\% & \textbf{75.9\%} \\
8. Comprehensive Evaluation & 0.0\% & 17.5\% & 10.8\% & \textbf{71.7\%} \\
\rowcolor{gray!30} Attributes Average & 0.1\% & 14.3\% & 14.5\% & \textbf{71.1\%} \\
\midrule
\rowcolor{gray!30} Human Rate & 1.5\% & 11.5\% & 19.2\% & \textbf{67.8\%} \\
\bottomrule[1pt]
\end{tabular}
\vspace{-15pt}
}
\end{table}

\begin{figure}[h]
  \centering
    \includegraphics[width=\linewidth]{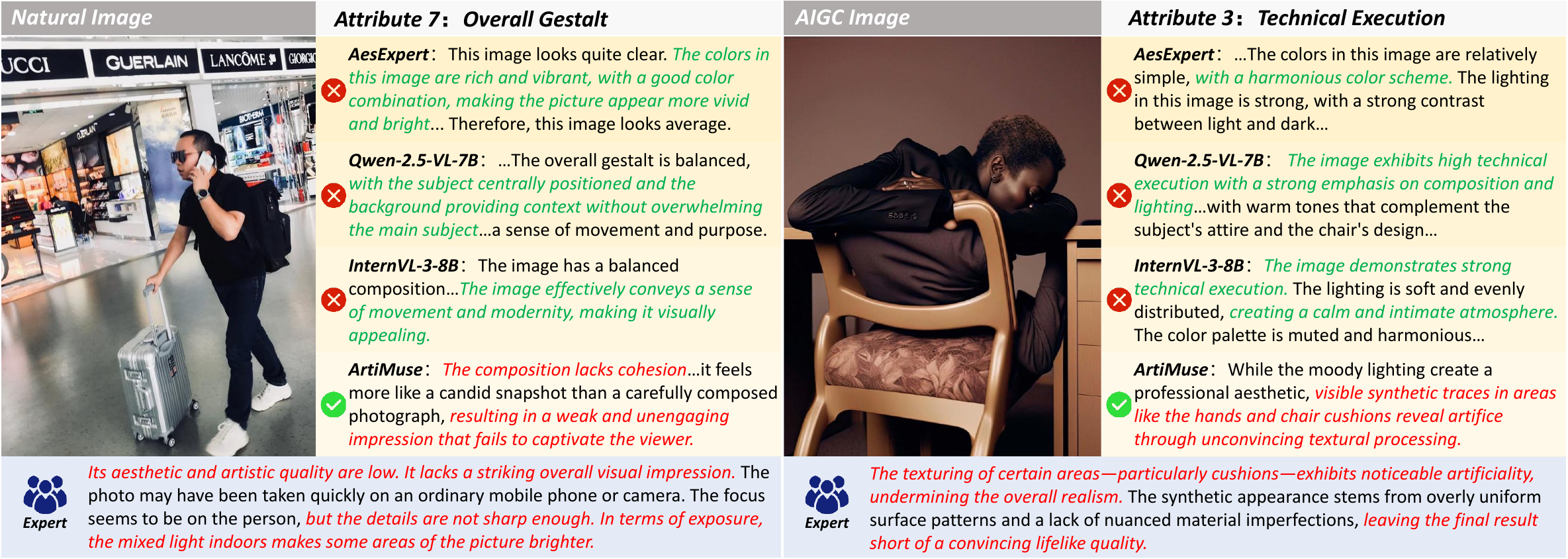}
    \vspace{-5pt}
    \includegraphics[width=\linewidth]{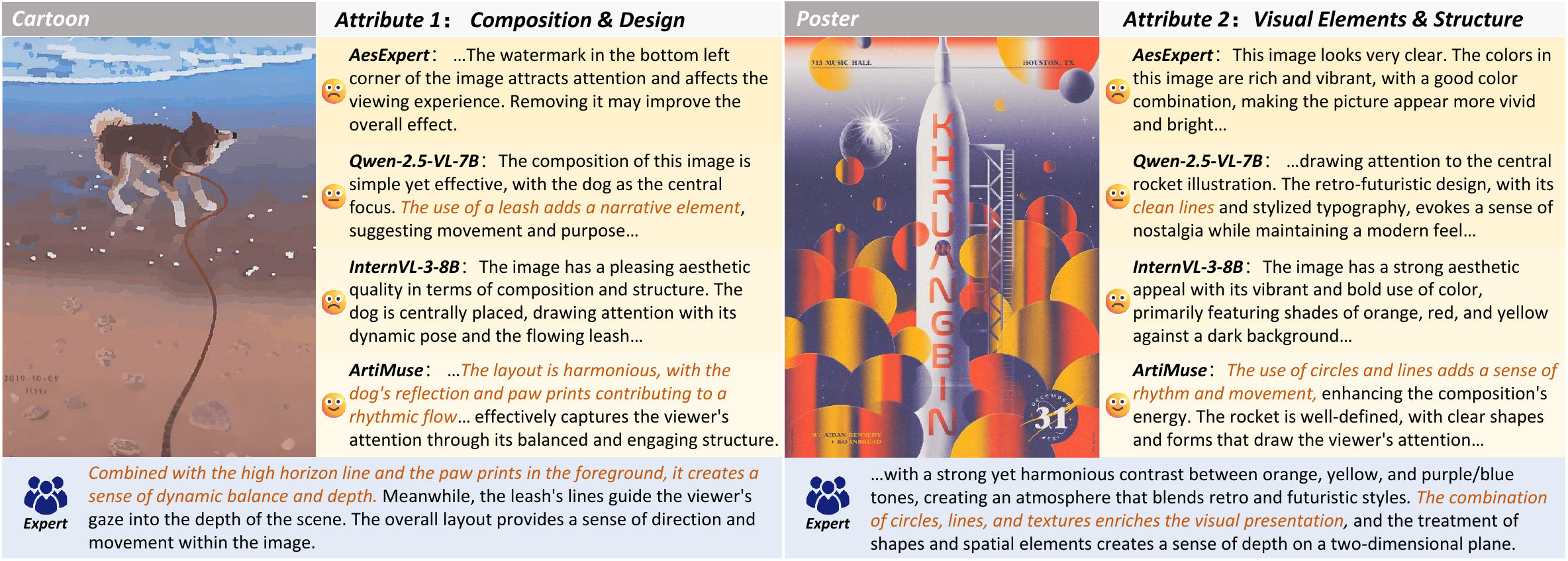}
    \vspace{-15pt}
  
  \caption{Structural aesthetic analysis results. \textcolor{red}{Red}, \textcolor{green}{green}, and \textcolor{brown}{brown} denote positive, negative, and expert-level analyses, respectively. ArtiMuse uniquely identifies flaws in low-aesthetic images while providing professional assessment of high-aesthetic images, capabilities absent in other models.}
  \label{fig:text_qualitative_combined}
  \vspace{-20pt}
\end{figure}

\noindent\textbf{Qualitative comparison.} Fig.~\ref{fig:text_qualitative_combined}presents a systematic evaluation of aesthetic analysis performance across different models. Our approach demonstrates consistent superiority in analyzing both natural and AIGC images, with particular strengths in identifying key aesthetic elements such as compositional cohesion and characteristic AIGC artifacts. More results are provided in the Supp.

% Structural aesthetic analysis of images. The \textcolor{red}{red}, \textcolor{green}{green}, and \textcolor{brown}{brown} colors denote positive, negative, and expert-level analyses, respectively. ArtiMuse uniquely identifies flaws in low-aesthetic images while providing professional assessment of high-aesthetic images, capabilities absent in other models.

% The \textcolor{brown}{brown} color represents expert-level analysis. ArtiMuse can analyze the merits of high-aesthetic images from a professional perspective, whereas other models fail to achieve this.

\subsection{Aesthetics Scoring}
\vspace{-5pt}
\noindent\textbf{Comparison across Multiple Image Aesthetics Scoring Datasets.} 
We evaluate the performance of ArtiMuse against other models across multiple Image aesthetics scoring datasets. For models unable for test (TANet~\cite{tad66k}, AesMamba~\cite{AesMamba}, UNIAA-LLaVA~\cite{Uniaa}, Next Token Is Enough~\cite{nexttoken}), we directly adopt the test results reported in their original papers. For models with unclear training protocols or those trained on general scenarios (MUSIQ~\cite{musiq}, VILA~\cite{vila}, mPLUG-Owl2~\cite{mPLUG-Owl2}, ShareGPT-4V~\cite{sharegpt4v}, Qwen-2.5-VL-7B~\cite{qwen2.5}, InternVL3-8B~\cite{internvl3}, Q-Instruct~\cite{qinstruct}, PEAS~\cite{peas}), we test their official released models. Both Q-Align~\cite{qalign} and our proposed model are fine-tuned on each target dataset. As shown in Tab.~\ref{tab:score_comp}, ArtiMuse demonstrates superior performance, achieving nearly the highest metrics across all datasets. Notably, it outperforms other models by over 0.05 PLCC on the PARA~\cite{para} and ArtiMuse-10K datasets, demonstrating its accurate aesthetics scoring capability.

\noindent\textbf{Generalization Ability.} 
We compare the generalization capabilities of ArtiMuse and Q-Align, the top-performing baseline model in our comparison. Both models are fine-tuned solely on the largest AVA dataset~\cite{ava} and subsequently evaluated on out-of-distribution datasets without additional adaptation. As presented in Table~\ref{tab:score_comp}, ArtiMuse consistently achieves superior performance over Q-Align across all benchmark datasets. Remarkably, ArtiMuse's zero-shot transfer performance exceeds that of several specialized IAA models, highlighting its exceptional generalization ability.
% We conduct a generalization capability comparison between ArtiMuse and Q-Align, which represents the best-performing baseline model in our evaluation. Both models are fine-tuned exclusively on the largest AVA~\cite{ava} dataset and then evaluated on other datasets without further adaptation. As shown in Tab.~\ref{tab:score_comp}, ArtiMuse consistently outperforms Q-Align across all datasets except ArtiMuse-10K. Notably, ArtiMuse's zero-shot performance surpasses even some specialized IAA models, demonstrating its superior generalization capabilities.

\noindent\textbf{Discussion of Image Aesthetics Scoring Datasets.} Prior work~\cite{qalign, deqa_score, Uniaa, nexttoken} has consistently demonstrated that IAA remains a challenging task due to the subjective nature of aesthetic perception and the substantial distributional shifts across different datasets. Our results in Tab.~\ref{tab:score_comp} further corroborate this observation: while models can achieve strong performance when fine-tuned on a single dataset, their accuracy often degrades significantly when evaluated on unseen datasets. 

\begin{table}[htbp]
\vspace{-10pt}
\centering
\caption{Comparison on aesthetics scoring. The best and second-best performances are highlighted in \textcolor{red}{red} and \textcolor{blue}{blue}, respectively. † Results are taken directly from original papers as these models cannot be tested. * Results are trained only on AVA to compare the generalization ability. For models without scoring capability, we prompt them to directly output scores as text for evaluation.}
\label{tab:score_comp}
\resizebox{\textwidth}{!}{%
\fontsize{8}{9}\selectfont
\renewcommand{\arraystretch}{0.8} % 调整行距，值越小行距越窄
\begin{tabular}{lcccccccccc}
\toprule
\multirow{2}{*}{\textbf{Model}} & \multicolumn{2}{c}{\textbf{AVA}~\cite{ava}} & \multicolumn{2}{c}{\textbf{PARA}~\cite{para}} & \multicolumn{2}{c}{\textbf{TAD66K}~\cite{tad66k}} & \multicolumn{2}{c}{\textbf{FLICKR-AES}~\cite{flickr}} & \multicolumn{2}{c}{\textbf{ArtiMuse-10K}} \\
\cmidrule(lr){2-3} \cmidrule(lr){4-5} \cmidrule(lr){6-7} \cmidrule(lr){8-9} \cmidrule(lr){10-11}
 & SRCC & PLCC & SRCC & PLCC & SRCC & PLCC & SRCC & PLCC & SRCC & PLCC \\
\midrule
\multicolumn{11}{c}{\textit{\textbf{Traditional Models}}} \\
MUSIQ~\cite{musiq} & 0.225 & 0.258 & 0.490 & 0.600 & 0.099 & 0.149 & 0.150 & 0.216 & -0.060 & -0.074 \\
TANet~\cite{tad66k} † & 0.758 & 0.765 & -- & -- & \textcolor{red}{0.513} & \textcolor{blue}{0.531} & -- & -- & -- & -- \\
VILA~\cite{vila} & 0.776 & 0.775 & 0.651 & 0.658 & 0.418 & 0.444 & 0.616 & 0.645 & 0.273 & 0.268 \\
AesMamba~\cite{AesMamba} † & 0.774 & 0.769 & \textcolor{red}{0.936} & \textcolor{blue}{0.902} & 0.511 & 0.483 & -- & -- & -- & -- \\
\midrule
% \multicolumn{11}{c}{\textit{\textbf{General Multimodal Large Language Models}}} \\
\multicolumn{11}{c}{\textit{\textbf{MLLMs for General-Purpose Applications}}} \\
mPLUG-Owl2~\cite{mPLUG-Owl2} & 0.206 & 0.211 & 0.376 & 0.372 & 0.089 & 0.106 & 0.382 & 0.359 & 0.159 & 0.145 \\
ShareGPT-4V~\cite{sharegpt4v} & 0.213 & 0.199 & 0.509 & 0.417 & 0.097 & 0.091 & 0.335 & 0.289 & 0.076 & 0.057 \\
Qwen-2.5-VL-7B~\cite{qwen2.5} & 0.391 & 0.371 & 0.721 & 0.743 & 0.240 & 0.242 & 0.621 & 0.578 & 0.256 & 0.179 \\
InternVL3-8B~\cite{internvl3} & 0.364 & 0.332 & 0.667 & 0.693 & 0.203 & 0.191 & 0.553 & 0.459 & 0.187 & 0.157 \\
\midrule
% \multicolumn{11}{c}{\textit{\textbf{Aesthetics Assessment Multimodal Large Language Models}}} \\
\multicolumn{11}{c}{\textit{\textbf{MLLMs for Image Aesthetics Assessment}}} \\
Q-Instruct~\cite{qinstruct} & 0.318 & 0.338 & 0.569 & 0.724 & 0.122 & 0.159 & 0.259 & 0.299 & -0.045 & -0.056 \\
PEAS~\cite{peas} & 0.748 & 0.748 & 0.686 & 0.700 & 0.415 & 0.444 & 0.577 & 0.613 & 0.306 & 0.293 \\
Q-Align~\cite{qalign} & 0.822 & 0.817 & \textcolor{blue}{0.913} & 0.888 & 0.501 & \textcolor{blue}{0.531} & \textcolor{blue}{0.798} & \textcolor{blue}{0.818} & \textcolor{blue}{0.551} & \textcolor{blue}{0.573} \\
UNIAA-LLaVA~\cite{Uniaa} † & 0.713 & 0.704 & 0.864 & 0.895 & 0.411 & 0.425 & 0.724 & 0.751 & -- & -- \\
Next Token Is Enough~\cite{nexttoken} † & \textcolor{red}{0.828} & \textcolor{blue}{0.825} & -- & -- & 0.413 & 0.444 & -- & -- & -- & -- \\
\rowcolor{gray!30} \textbf{ArtiMuse (Ours)} & \textbf{\textcolor{blue}{0.827}} & \textbf{\textcolor{red}{0.826}} & \textbf{\textcolor{red}{0.936}} & \textbf{\textcolor{red}{0.958}} & \textbf{\textcolor{blue}{0.510}} & \textbf{\textcolor{red}{0.543}} & \textbf{\textcolor{red}{0.814}} & \textbf{\textcolor{red}{0.837}} & \textbf{\textcolor{red}{0.614}} & \textbf{\textcolor{red}{0.627}} \\
% \midrule
% \multicolumn{11}{c}{\textit{\textbf{Comparison of Generalization Ability}}} \\
% Q-Align (ICML 2024) * & 0.822 & 0.817 & - & - & - & - & 0.643 & 0.664 & -- & -- \\
\midrule
\multicolumn{11}{c}{\textit{\textbf{Comparison of Generalization Ability}}} \\
Q-Align * & 0.822 & 0.817 & 0.694 & 0.711 & 0.417 & 0.445 & 0.643 & 0.664 & 0.337 & 0.320 \\
ArtiMuse (Ours) * & 0.827 & 0.826 & 0.697 & 0.725 & 0.419 & 0.451 & 0.647 & 0.676 & 0.395 & 0.376 \\
\bottomrule
\end{tabular}
}
\vspace{-10pt}
\end{table}

\subsection{Ablation Studies} 

\noindent\textbf{Datasets Variants.} We conduct 4 experiments (a)-(c) to systematically validate the contribution of each dataset component, as shown in Tab.~\ref{tab:ablation-study}. The results demonstrate consistent performance drop when any component is removed, with the most significant drop occurring upon exclusion of the Images w/ score caption subset, for the subset's inclusion of data from AVA. The results underscore the critical impact of dataset composition on model performance.

\noindent\textbf{Training Strategy.} Comparative analysis between (d) and (i) reveals that full fine-tuning significantly impacts model performance, primarily due to the loss of fundamental aesthetic priors acquired during the text pretraining phase. This finding is further substantiated by the comparison between (e) and (i), which conclusively demonstrates the effectiveness of our proposed 2-stage training paradigm. The results indicate that preserving pretrained text representations while adapting to score prediction tasks yields superior performance compared to end-to-end joint training approaches.

\noindent\textbf{Score Prediction.} Our systematic exploration of score prediction strategies is presented in (f)-(i). Exp.(f), which directly converts scores to text for both training and inference, demonstrates suboptimal performance. The introduction of aesthetics score Tokens yields significant improvements, with analysis revealing that (g) suffers from insufficient token granularity while (h) is hampered by excessive token complexity. Configuration (i) achieves the optimal balance between precision and learnability, establishing it as our final choice. More experiments are provided in the Supp. 

\begin{table}[htbp]
\centering
\caption{Ablation studies. The table compares different combinations of dataset variants, training, and training methods, with evaluation metrics SRCC and PLCC reported for AVA dataset.}
\label{tab:ablation-study}
\resizebox{\textwidth}{!}{%
\begin{tabular}{l|ccc|c|c|cc}
\toprule[1.5pt]
Exp. &
\makecell{\textbf{Images w/} \\ \textbf{Score Caption}} & 
\makecell{\textbf{Images w/} \\ \textbf{Partial Text Caption}} & 
\makecell{\textbf{Professionally } \\ \textbf{Selected Images}} & 
% \textbf{Images w/ partial text caption} & 
% \textbf{Professionally selected images} & 
% \textbf{LLM finetuning} & 
% \textbf{Training stages} & 
\textbf{Training Strategies} & 
\textbf{Score Prediction} & 
% \textbf{Training Strategy} & 
\textbf{SRCC} & 
\textbf{PLCC} \\
\midrule
(a) & $\checkmark $ & $\checkmark$ & -- & LLM LoRA / 2-Stage Training & 100 Aesthetics Score Tokens & 0.824 & 0.825 \\
(b) &  -- & $\checkmark$ & $\checkmark$ & LLM LoRA / 2-Stage Training & 100 Aesthetics Score Tokens & 0.621 &  0.627 \\
(c) & $\checkmark$ &  -- & $\checkmark$ & LLM LoRA / 2-Stage Training & 100 Aesthetics Score Tokens & 0.825 & 0.824 \\
\midrule
(d) & $\checkmark$ & $\checkmark$ & $\checkmark$ & LLM Full-finetune / 2-Stage Training  & 100 Aesthetics Score Tokens & 0.816 & 0.814 \\
(e) & $\checkmark$ & $\checkmark$ & $\checkmark$  &  LLM LoRA / Joint Training & 100 Aesthetics Score Tokens  & 0.821 & 0.820 \\
\midrule
(f) & $\checkmark$ & $\checkmark$ & $\checkmark$  & LLM LoRA / 2-Stage Training & Output Score As Text & 0.820 & 0.819 \\
(g) & $\checkmark$ & $\checkmark$ & $\checkmark$  & LLM LoRA / 2-Stage Training & 5 Aesthetics Score Tokens  & 0.823 & 0.821 \\
(h) & $\checkmark$ & $\checkmark$ & $\checkmark$  & LLM LoRA / 2-Stage Training & 200 Aesthetics Score Tokens & 0.823 & 0.819 \\

\rowcolor{gray!30} (i) & $\checkmark$ & $\checkmark$ & $\checkmark$  & LLM LoRA / 2-Stage Training 
 & 100 Aesthetics Score Tokens & \textbf{0.827} & \textbf{0.826} \\

\bottomrule[1.5pt]
\end{tabular}
}
\vspace{-10pt}
\end{table}

\section{Conclusion}
\vspace{-5pt}
\label{sec:Conclusion}
We introduce ArtiMuse-10K, a large expert-annotated dataset for image aesthetics assessment, and ArtiMuse, the first model to achieve expert-level textual evaluation and precise aesthetics scoring. Additionally, we propose Token As Score, a lightweight yet effective method enabling precise continuous score prediction in MLLMs. Together these contributions will advance the field of image aesthetics assessment by providing more comprehensive dataset, more superior model, and more efficient scoring paradigm.
% We present ArtiMuse, a multimodal model that advances image aesthetics assessment through fine-grained understanding and interpretable feedback. By introducing the expert-annotated ArtiMuse-10K dataset and a novel score-as-token technique, we overcome key limitations in existing IAA methods, including coarse-grained evaluation and continuous score prediction challenges. ArtiMuse achieves state-of-the-art performance across benchmarks, demonstrating robust aesthetic reasoning. Our framework, with eight explainable attributes, bridges technical and artistic evaluation, enabling applications in AIGC quality control and creative assistance. The proposed score-as-token method provides an efficient solution for continuous value modeling in MLLMs. This work establishes a foundation for future research in interpretable aesthetic AI, with potential extensions to dynamic content and personalized feedback. 

\noindent\textbf{Limitations.} The current model is limited to understanding and analyzing, and is unable to provide professional aesthetic enhancement recommendations, which will be addressed in future work.
% 当前的模型局限在understanding和analyzing，无法给出专业的美学质量提升建议，会在future work中完成

{
    \small
    % Generated by IEEEtran.bst, version: 1.14 (2015/08/26)

    % \bibliographystyle{IEEEtran} \bibliography{neurips_2025}
}

%%%%%%%%%%%%%%%%%%%%%%%%%%%%%%%%%%%%%%%%%%%%%%%%%%%%%%%%%%%%

\appendix
\clearpage % 确保目录从新页开始

\begin{center}
    \vspace*{2cm} % 垂直间距，让标题更居中偏上
    {\Large\bfseries Appendix} % 目录标题，可以根据需要调整字体大小和粗细
    \vspace*{1cm} % 标题下方的间距
\end{center}

\tableofcontents % 生成目录

\clearpage % 确保目录结束后，正文从新页开始

\section{ArtiMuse-10K Dataset Details}
% 公开数据集具体的收集情况，过滤方式

\subsection{Details of Aesthetic Attributes}
The ArtiMuse-10K dataset employs structural analysis for textual annotations, with each image evaluated across eight fine-grained aesthetic attributes. These attributes were rigorously defined by a panel of domain experts, all of whom possess at least \textbf{3 years} of formal training in aesthetics, with the most senior member boasting \textbf{over 30 years} of professional experience in the field. This ensures comprehensive coverage of key image aesthetics dimensions while maintaining robust generalizability across diverse image types—including designs, photographs, paintings, calligraphy, and AI-generated content (AIGC) images. The detailed of these attributes are presented in Tab.~\ref{tab:attributes}.

% ArtiMuse-10K数据集中的文字标注采用结构化分析策略，从8个细分美学属性进行标注，这8个属性是由领域专家定义的，能够涵盖图像美学的重要方面，并且通用于包含设计、摄影、绘画、书法、AIGC等各种类型图像的美学评价中。其具体属性介绍如Tab.1所示。

\begin{table*}[htbp]
\centering
% \vspace{-13pt}
\label{tab:attributes}
\caption{Aesthetic attributes and their descriptions of ArtiMuse-10K dataset.}
\resizebox{0.99\textwidth}{!}{%
% \fontsize{8}{9}\selectfont
% \renewcommand{\arraystretch}{0.6} % 调整行距，值越小行距越窄
\begin{tabularx}{\linewidth}{c l l}
\toprule[1pt]
\textbf{No.} & \textbf{Attribute} & \textbf{Description} \\
\midrule
1 &
Composition \& Design &
\makecell[l]{Evaluate the balance, contrast, layout aesthetics, and \\ rhythm of the composition. Focus on the use of dynamic \\focal points, unity, and harmony in the design.}
\\
\midrule
2 &
Visual Elements \& Structure & 
\makecell[l]{Analyze the interplay of color, geometry,\\ spatial organization, and illumination to optimize visual\\ contrast and structural clarity.}
\\

\midrule
3 &
Technical Execution &
\makecell[l]{Examine the mastery of medium and materials, including\\ brushstrokes, focus, exposure, light handling, as well as \\clarity and resolution of the image.}
\\

\midrule
4 &
Originality \& Creativity & 
\makecell[l]{Analyze the uniqueness of the concept and execution,\\ focusing on how the work exceeds common styles with  \\imagination, and creative breakthroughs.}
\\

\midrule
5 &
Theme \& Communication & 
\makecell[l]{Evaluate the clarity of the subject and its communication.\\ Consider how effectively the narrative, cultural \\significance, and societal context are conveyed.}
\\

\midrule
6 &
Emotion \& Viewer Response & 
\makecell[l]{Assess how well the work evokes an emotional response, \\engages the viewer, and creates lasting impressions\\ with personal significance.}
\\

\midrule
7 &
Overall Gestalt & 
\makecell[l]{Evaluate the overall visual appeal and artistic impact of \\ the image, considering how well the elements combine\\ to create an engaging, meaningful impression.}
\\

\midrule
8 &
Comprehensive Evaluation &
\makecell[l]{Provide a comprehensive aesthetics assessment of \\the image, evaluating its effectiveness in visual impact,\\ theme communication, and artistic depth.}
\\

\midrule
-- &
Overall Aesthetics Score &
\makecell[l]{Overall aesthetics score derived from \\ multi-dimensional evaluation.}
\\

\bottomrule[1pt]
\end{tabularx}}
% \vspace{-15pt}
\end{table*}

\subsection{Charateristics of ArtiMuse-10K} 

\begin{figure}[h]
  \centering
    \includegraphics[width=0.99\linewidth]{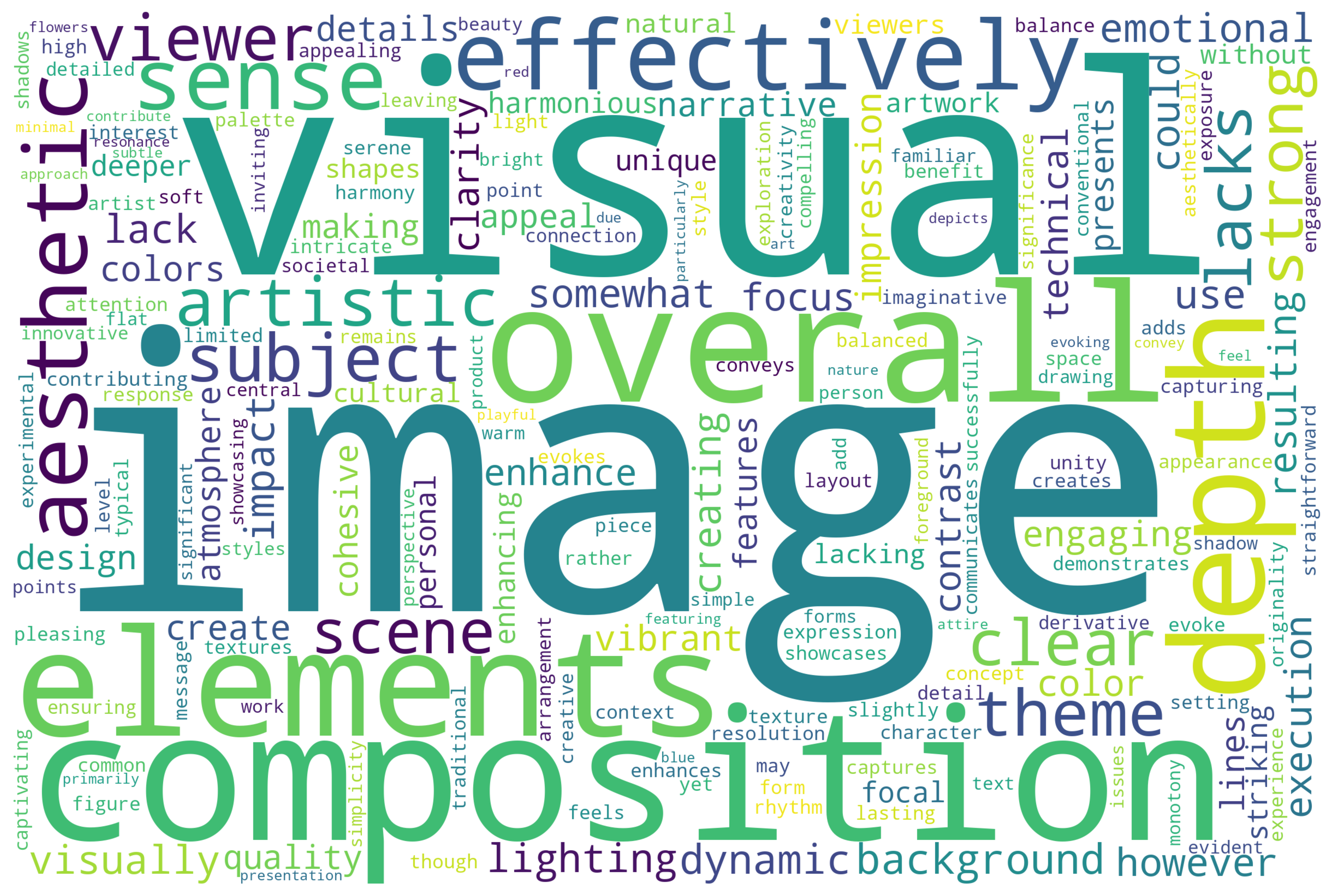}
   % \vspace{-10pt}
   \caption{Wordcloud of ArtiMuse-10K dataset. The most frequent words in ArtiMuse-10K dataset are all highly relevant to the aesthetic assessment of images.}
   \label{fig:wordcloud}
   \vspace{-10pt}
\end{figure}

\noindent\textbf{WordCloud.}  WordCloud of our introduced ArtiMuse-10K dataset is depicted in Fig.~\ref{fig:wordcloud} . We analyze the textual annotations of ArtiMuse across eight aesthetic attributes and find that the most frequently occurring terms—such as "image," "visual," "composition," "overall," and "elements"—are strongly correlated with image aesthetic quality. This observation suggests that human experts primarily focus on fundamental visual characteristics when assessing artistic merit.
% 我们统计了ArtiMuse的8个aesthetic attributes上的文字标注，其中出现最频繁的词如Image,visual, composition, overall, elements等词都和图像的美学质量强相关。

\begin{figure}[htbp]
  \centering
    \includegraphics[width=0.95\linewidth]{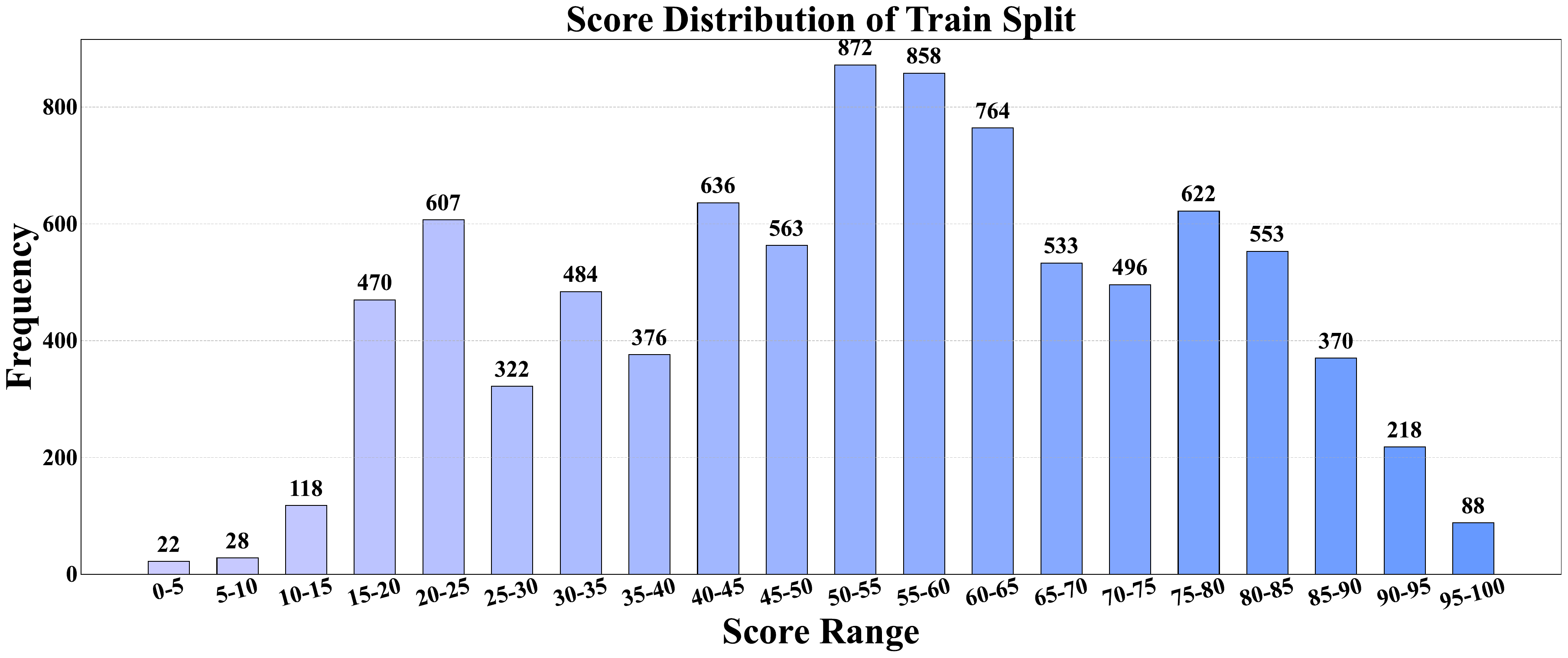}
    \includegraphics[width=0.95\linewidth]{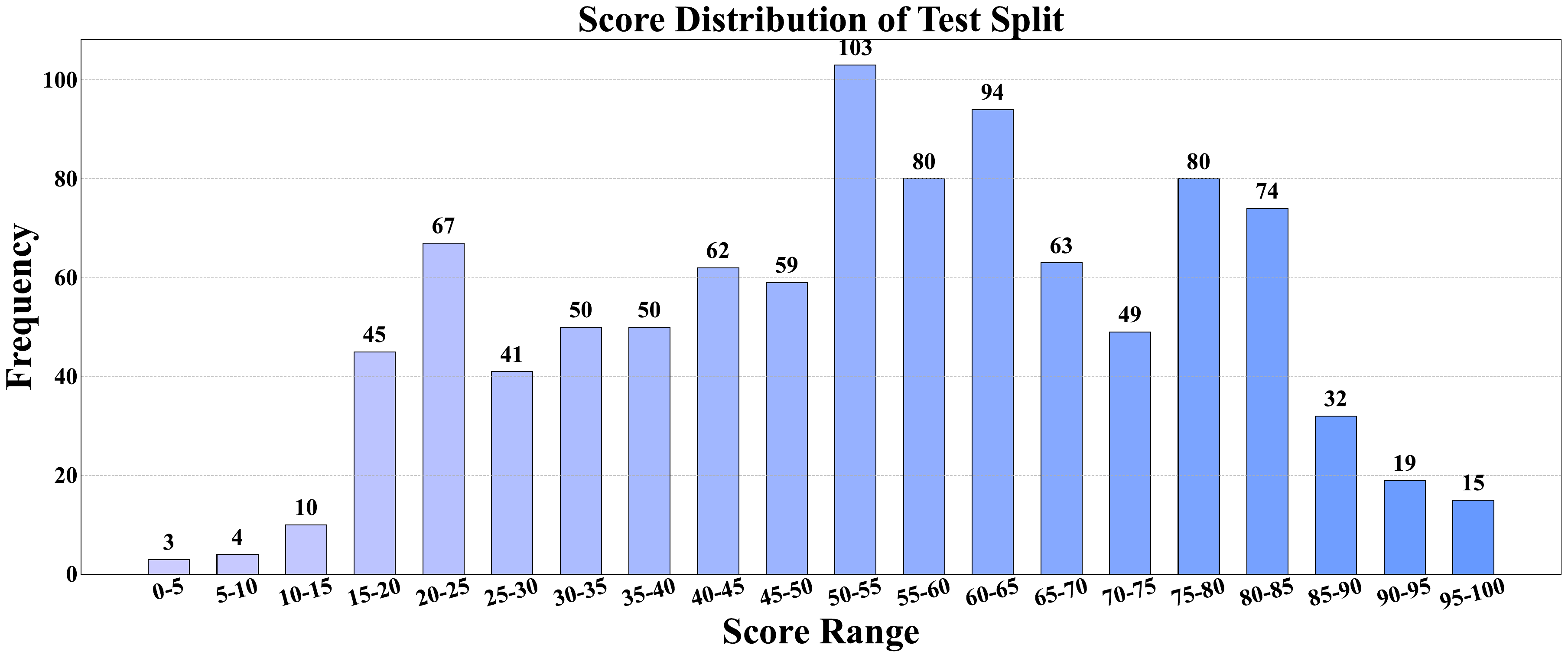}
   \vspace{-10pt}
   \caption{Score distribution of training and test splits in ArtiMuse-10K.}
   \label{fig:train_test_dist_artimuse10k}
   \vspace{-10pt}
\end{figure}

\begin{figure}[htbp]
  \centering
    \includegraphics[width=0.95\linewidth]{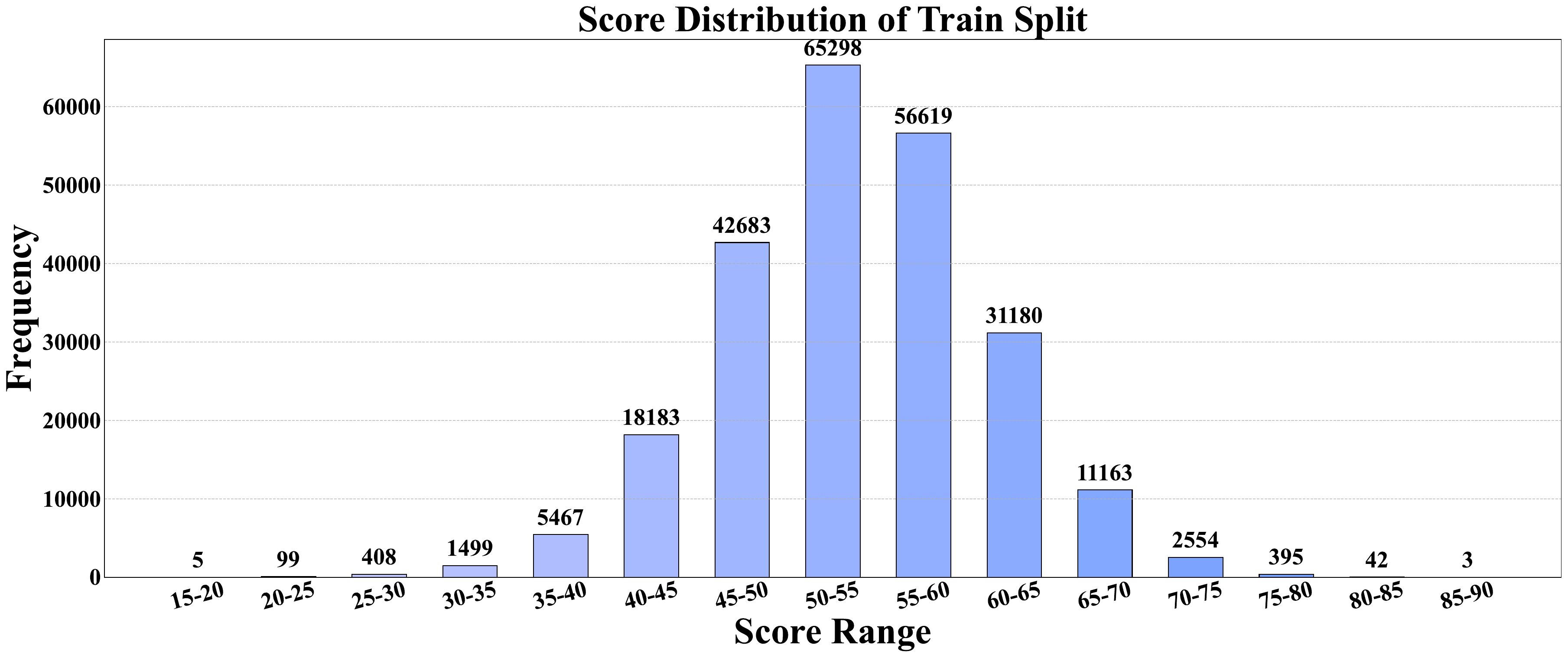}
    \includegraphics[width=0.95\linewidth]{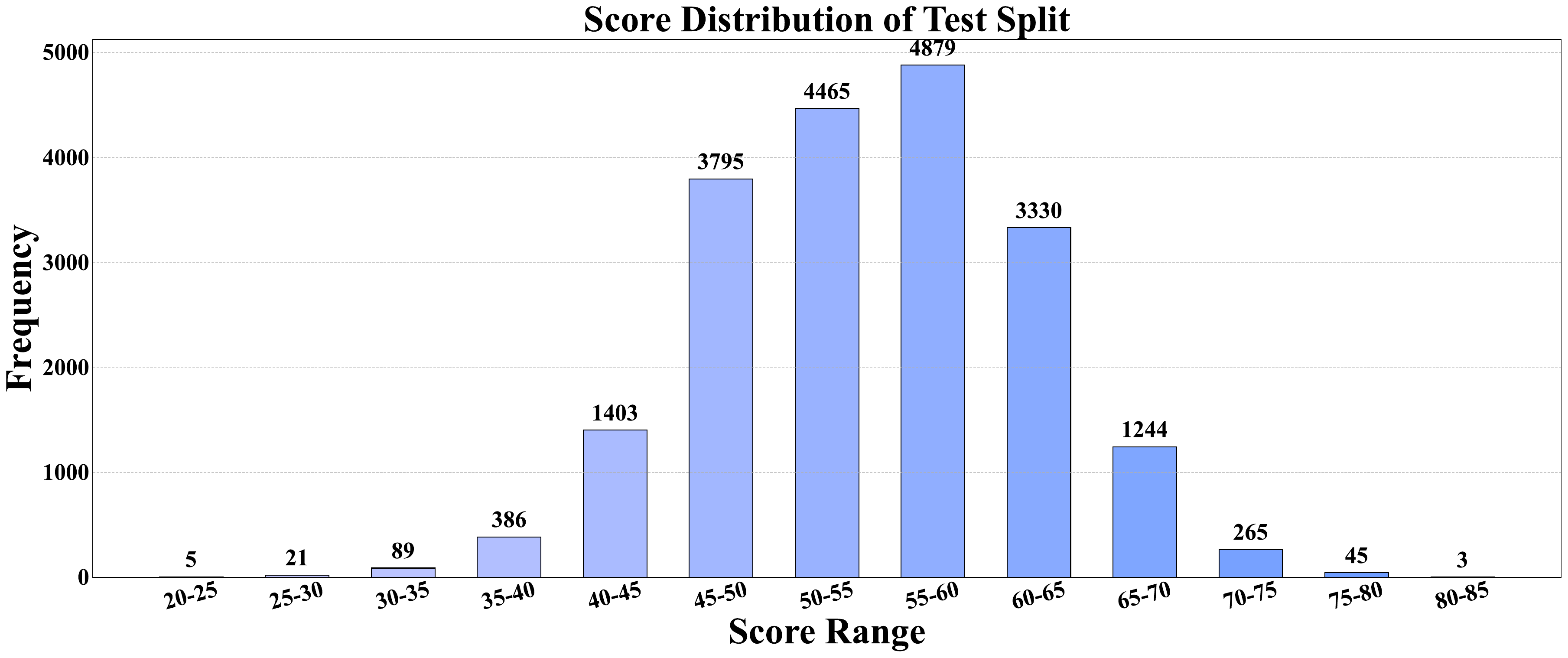}
   \vspace{-10pt}
   \caption{Score distribution of training and test splits in AVA.}
   \label{fig:train_test_dist_ava}
   \vspace{-10pt}
\end{figure}

\begin{figure}[htbp]
  \centering
    \includegraphics[width=0.95\linewidth]{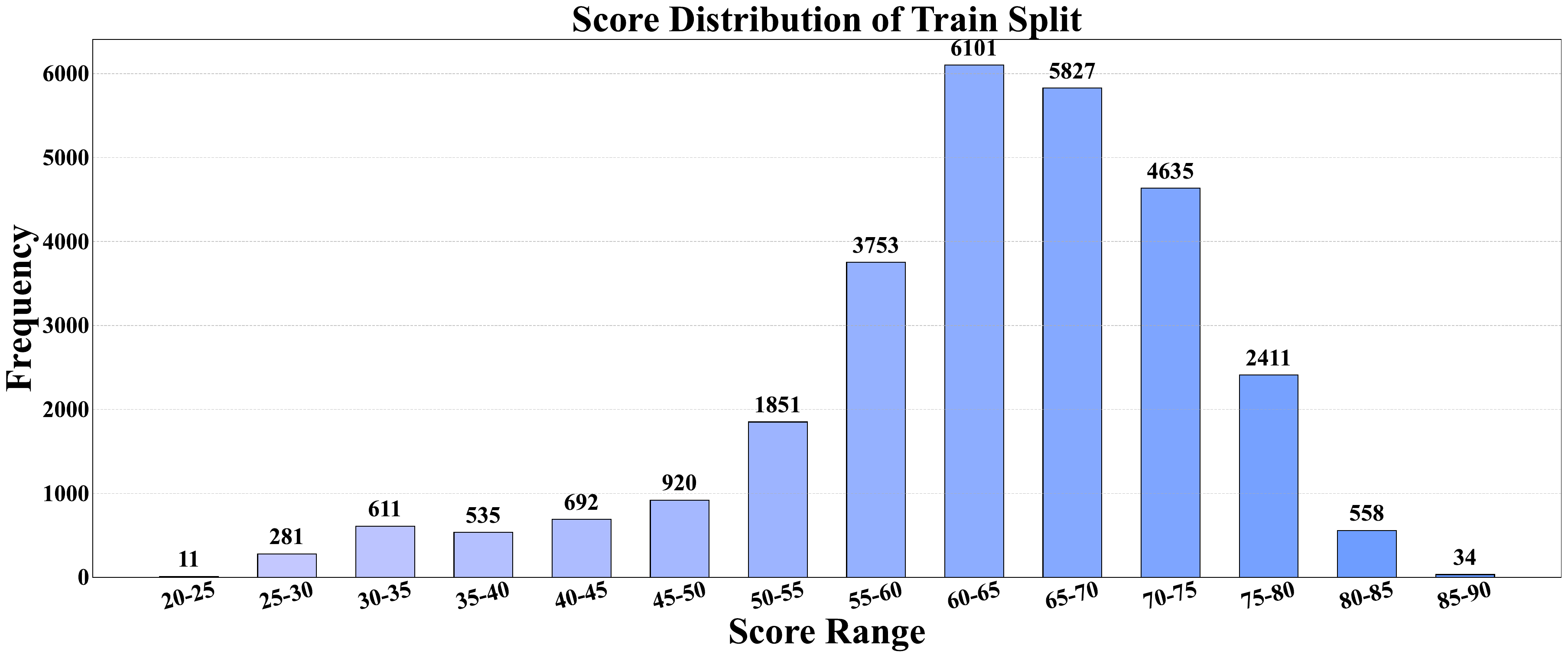}
    \includegraphics[width=0.95\linewidth]{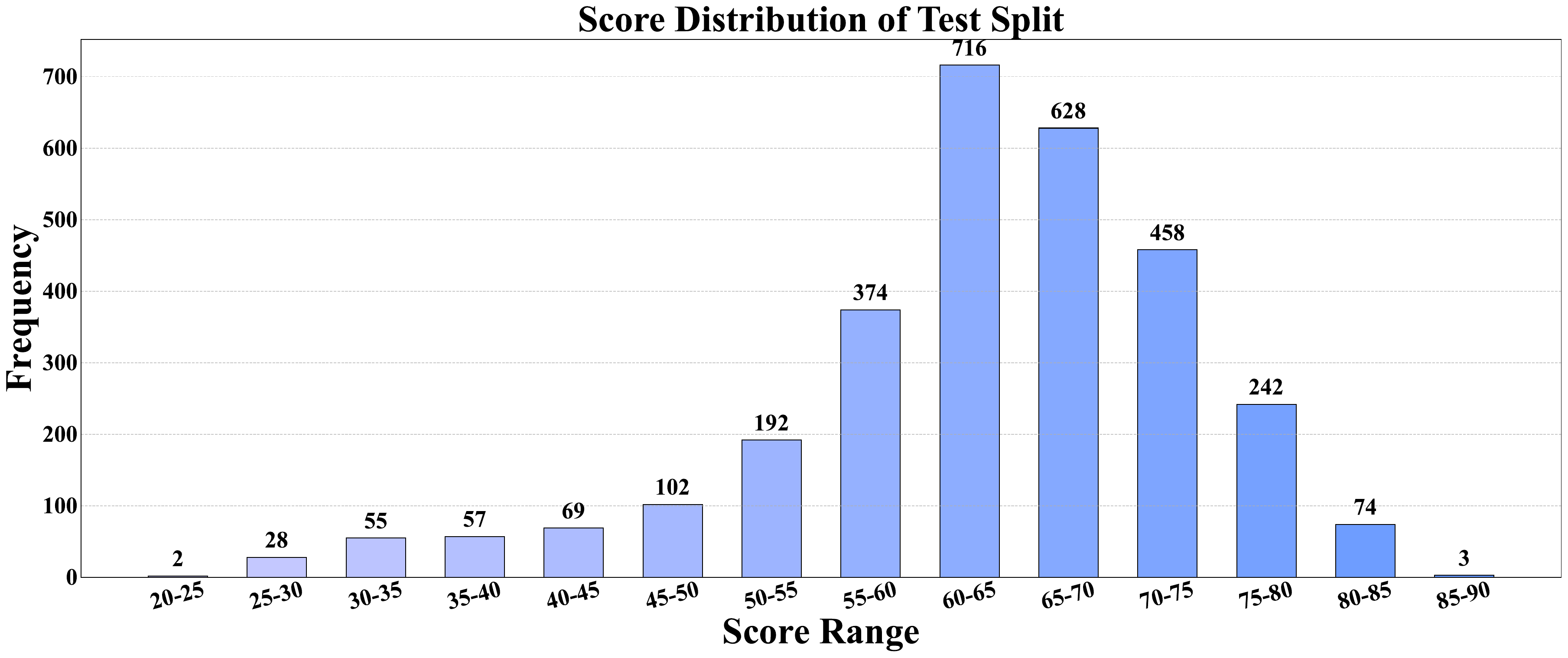}
   \vspace{-10pt}
   \caption{Score distribution of training and test splits in PARA.}
   \label{fig:train_test_dist_para}
   \vspace{-10pt}
\end{figure}

\begin{figure}[htbp]
  \centering
    \includegraphics[width=0.95\linewidth]{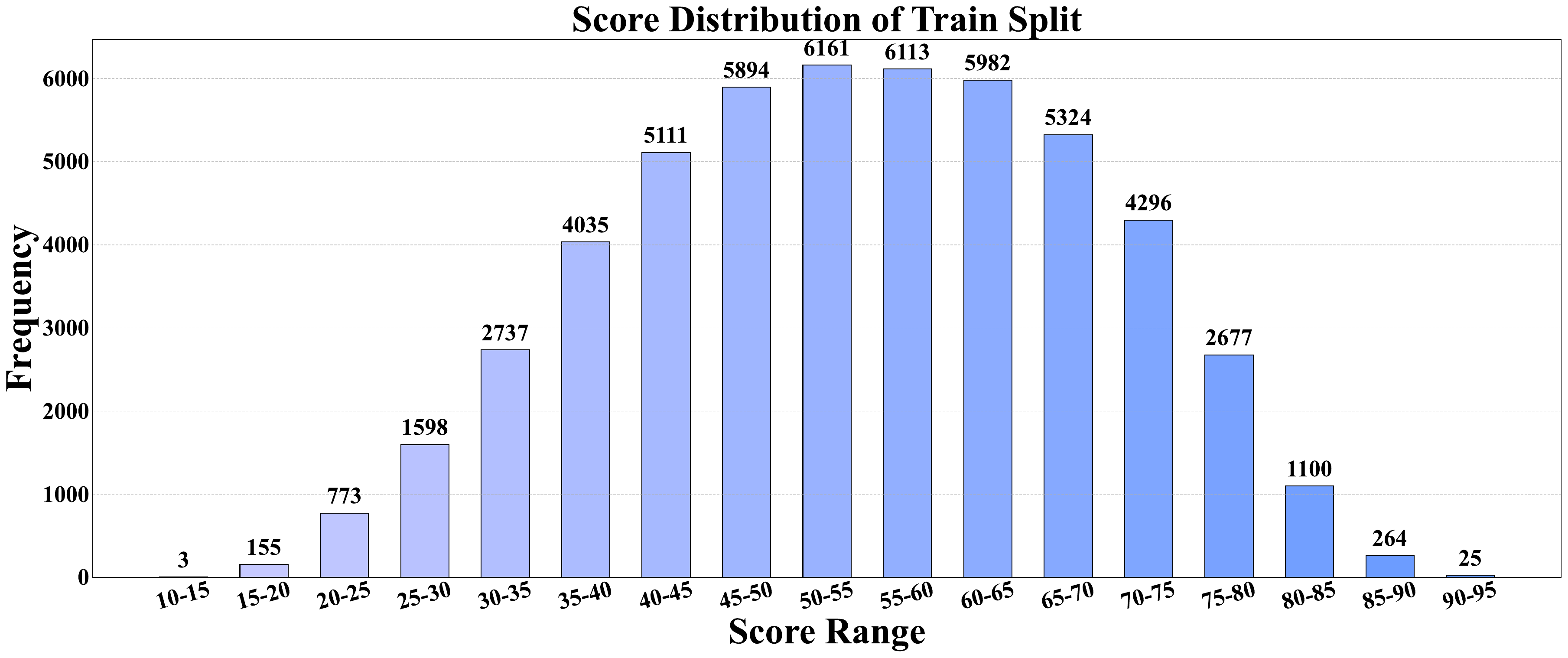}
    \includegraphics[width=0.95\linewidth]{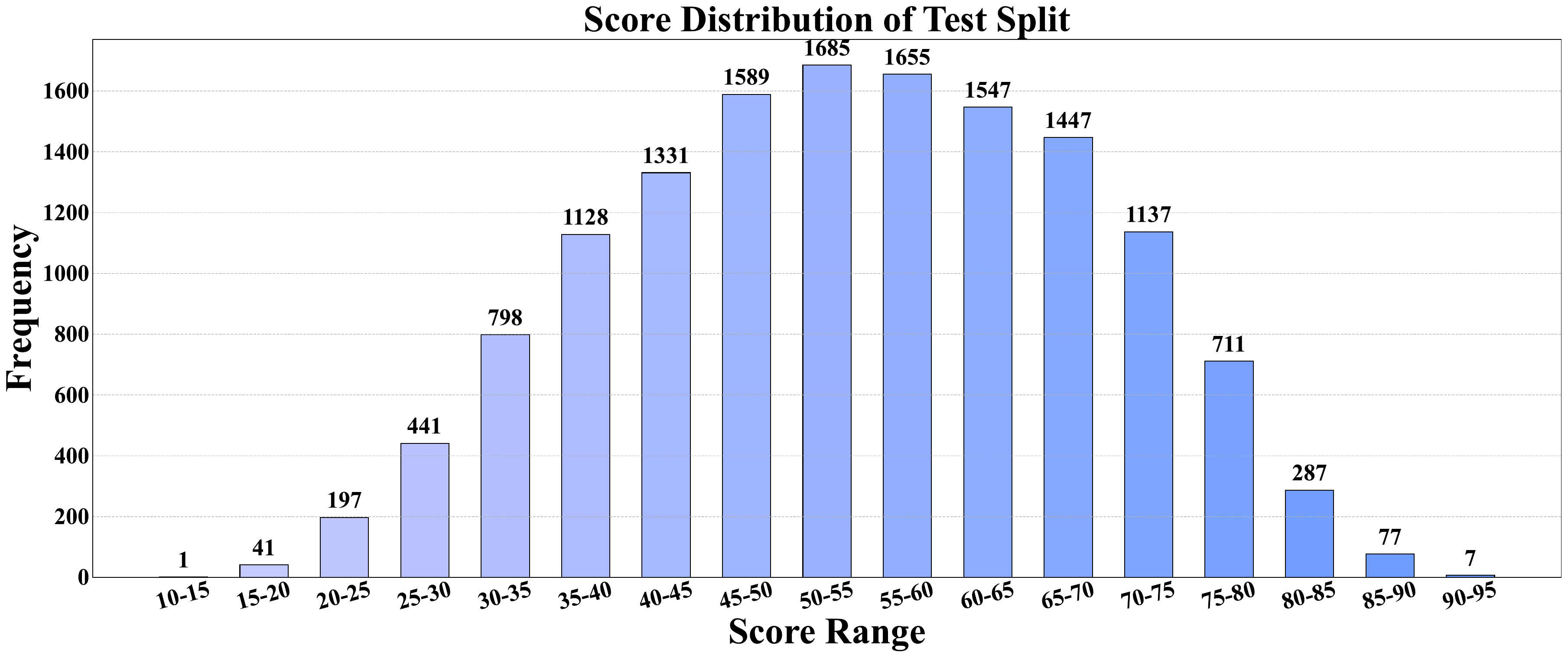}
   \vspace{-10pt}
   \caption{Score distribution of training and test splits in TAD66K.}
   \label{fig:train_test_dist_tad66k}
   \vspace{-19pt}
\end{figure}

\begin{figure}[htbp]
  \centering
    \includegraphics[width=0.95\linewidth]{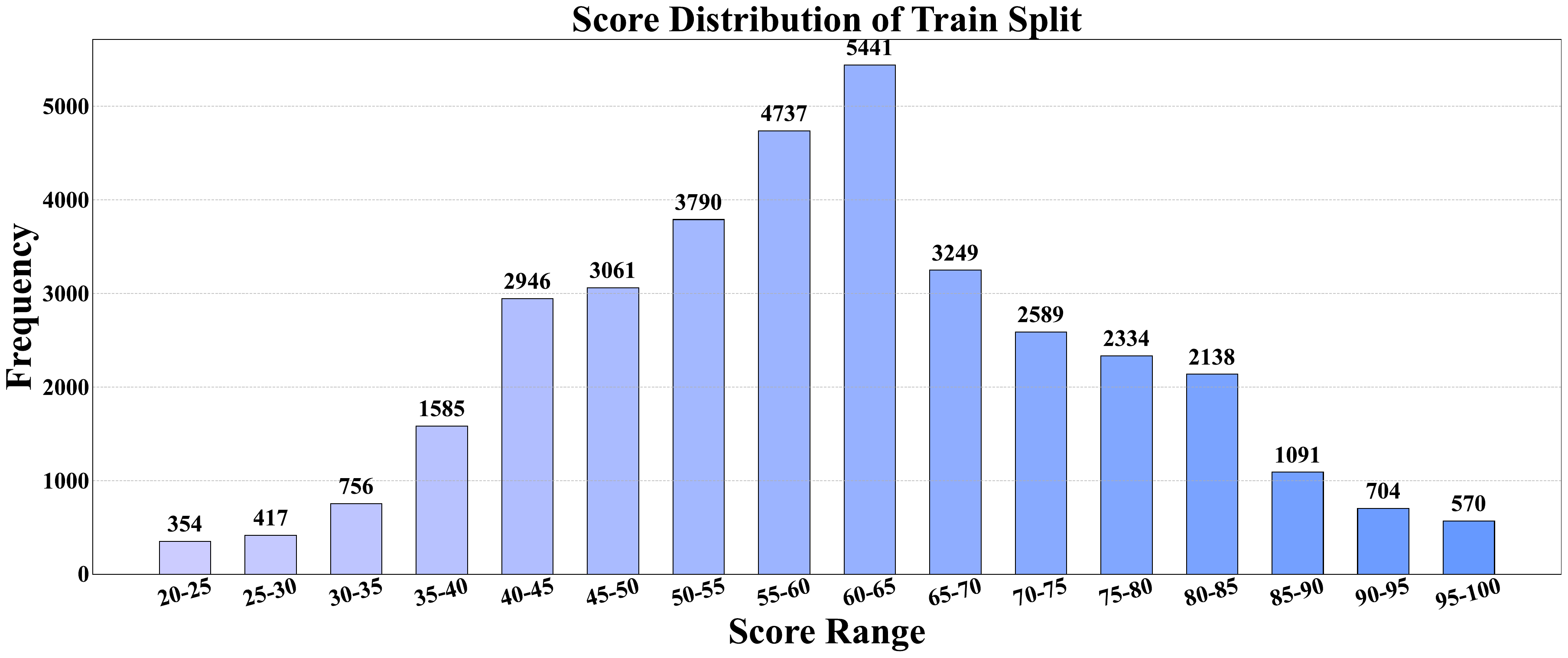}
    \includegraphics[width=0.95\linewidth]{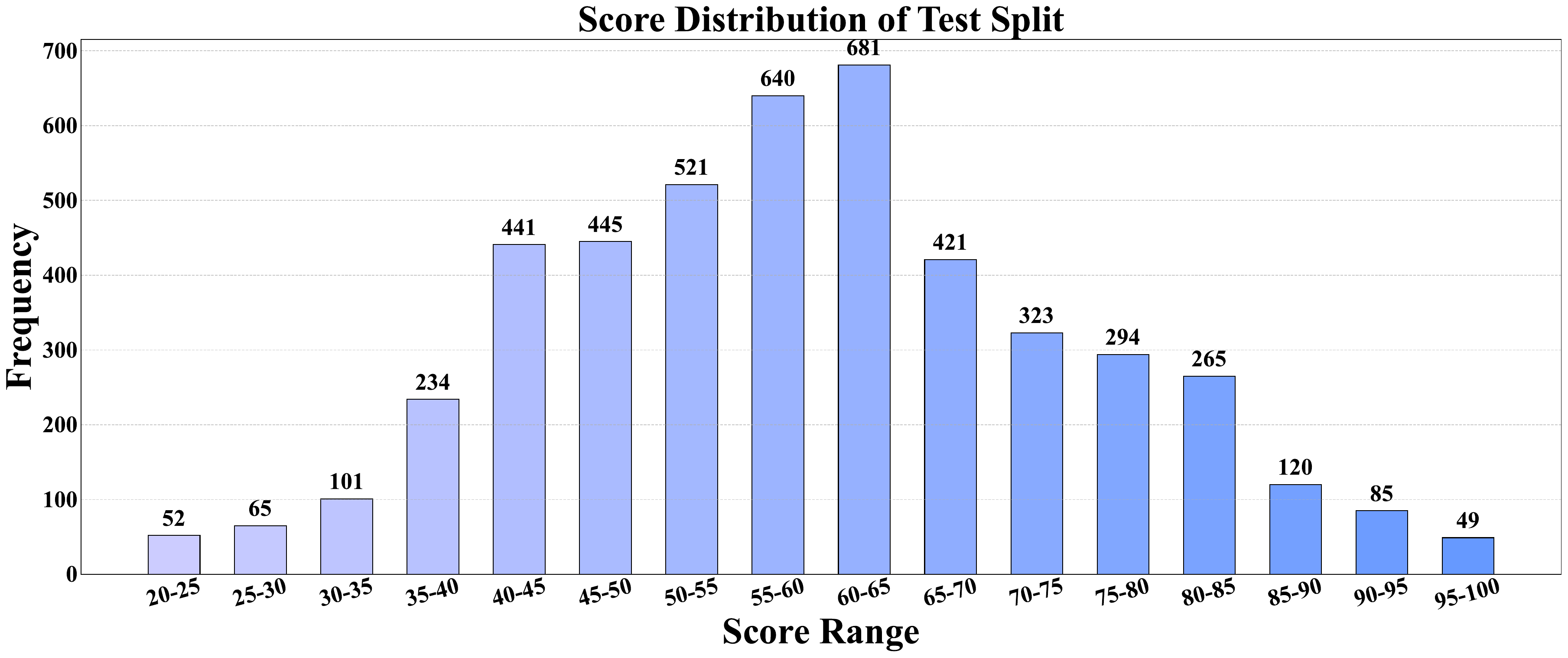}
   \vspace{-10pt}
   \caption{Score distribution of training and test splits in FLICKR-AES.}
   \label{fig:train_test_dist_flickr}
   \vspace{-10pt}
\end{figure}

% 我们将ArtiMuse-10K dataset的10000张图像分成了train split和test split，分别包含9k和1k张图片。我们分别统计了train dataset和test dataset的分数分布，展示在Fig.1和Fig.2中，他们都近似服从高斯分布。
% 为了对比不同数据集的分布差异，我们将AVA，PARA, TAD66K, FLICKR-AES的分数均归一化到0-100区间，并进行了score 分布的统计，结果依次展示在a,b,c,d中。分析表明，包括ArtiMuse在内的各个aesthetics scoring数据集均近似服从高斯分布，且ArtiMuse有着更好的分数多样性。
\noindent\textbf{Score Distributions.} 
We divide the 10,000 images in the ArtiMuse-10K dataset into a training split (9,000 images) and a test split (1,000 images). The score distributions for both the training and test datasets are shown in Fig.~\ref{fig:train_test_dist_artimuse10k}. To compare the distribution differences across datasets, we normalize the scores of AVA~\cite{ava}, PARA~\cite{para}, TAD66K~\cite{tad66k}, and FLICKR-AES~\cite{flickr} to the $[0, 100]$ range and analyze their score distributions, with results shown in Fig.~\ref{fig:train_test_dist_ava}, Fig.~\ref{fig:train_test_dist_para}, Fig.~\ref{fig:train_test_dist_tad66k}, and Fig.~\ref{fig:train_test_dist_flickr} respectively. Our analysis reveals that all aesthetic scoring datasets, including ArtiMuse, approximately follow Gaussian distributions. Notably, ArtiMuse demonstrates superior score diversity compared to other datasets. 

% \begin{figure}[h]
%   \centering
%     \includegraphics[width=0.99\linewidth]{figures/supp/test_dist.pdf}
%    % \vspace{-10pt}
%    \caption{Test split distribution of ArtiMuse-10K.}
%    \label{fig:test_dist}
%    % \vspace{-5pt}
% \end{figure}

% ArtiMuse-10K的图像来源丰富，共包括5 main categories和15 subcategories，不同类别的图像数目详见Tab.1
\noindent\textbf{Statistical Analysis across Categories.} 
The images in ArtiMuse-10K are sourced from diverse origins and encompass a total of 5 main categories and 15 subcategories. The detailed distribution of image counts across these categories is presented in Tab.~\ref{tab:categoriy_counts}.

\begin{table}[htbp]
    \centering
    \caption{Statistics of ArtiMuse-10K across main categories and subcategories.}
    \label{tab:categoriy_counts}
    \resizebox{\textwidth}{!}{
    \begin{tabular}{llp{6cm}cc}
        \toprule[1pt]
        \textbf{Main Category} & \textbf{Subcategory} & \textbf{Description} & \textbf{\# Image} \\
        \midrule
        \multirow{6}{*}{Photography} 
            & Daily Photo       & Casual photos capturing daily scenes & 3071 \\
            & Photographic Art  & Photos with artistic processing & 758 \\
            & Architecture      & Photos of buildings and structures & 119 \\
            & Portrait          & Portrait photography & 82 \\
            & Movie still       & Screenshots from films or TV shows & 81 \\
            & \textbf{\textit{Total}} & -- & \textbf{4111} \\
        \midrule
        \multirow{7}{*}{Painting \& Calligraphy} 
            & Digital Art       & \makecell[l]{Computer-aided digital paintings} & 1314 \\
            & Children's Painting & Paintings created by children & 699 \\
            & Chinese Painting  & Chinese ink wash paintings & 511 \\
            & General Painting  & General paintings with diverse scopes & 485 \\
            & Sketch            & Pencil/charcoal sketches & 43 \\
            & Calligraphy       & Artistic handwriting and lettering & 43 \\
            & \textbf{\textit{Total}} & --  & \textbf{3095} \\
        \midrule
        \multirow{2}{*}{AIGC} 
            & AIGC              & \makecell[l]{AI-generated content \\(particularly generative models)} & 1453 \\
            & \textbf{\textit{Total}} & --  & \textbf{1453} \\
        \midrule
        \multirow{3}{*}{3D Design} 
            & Product Design    & 3D model snapshots for products & 516 \\
            & Sculpture         & Sculpting artwork snapshots & 307 \\
            & \textbf{\textit{Total}} & --  & \textbf{823} \\
        \midrule
        \multirow{2}{*}{Graphic Design} 
            & Graphic Design    & Posters/logos/visual designs & 518 \\
            & \textbf{\textit{Total}} & --  & \textbf{518} \\
        \midrule
        \textbf{\textit{Total}} & -- & -- & \textbf{10000} \\
        \bottomrule[1pt]
    \end{tabular}}
\end{table}

% ArtiMuse-10K中由professional experts对每张图像从8个aesthetic attributes进行了细致的文字评价，并给出一个总体评分。我们在这里展示数据集中每个数据的完整形式，

% \section{Method Details}
% 我们从现有的公开数据集中选择了包含较高质量aesthetic caption的部分数据集进行sample，采样的时候特别保证了采样结果的 AestheticCaptionQuality和AestheticQualityDiversity，具体采样数量如Tab.1所示。
\section{Details of Public Dataset Collection \& Processing}
We select and sample a subset of high-quality aesthetic captions from existing public datasets, with particular emphasis on ensuring both aesthetic caption quality and diversity in the sampling process. The specific sampling statistics are presented in Tab.~\ref{tab:dataset_public}.

\begin{table}[htbp]
    \centering
    \caption{Collection \& processing results of public datasets.}
    \label{tab:dataset_public}
    \begin{tabular}{c | c c}
        \toprule[1.5pt]
        \textbf{Public Dataset} & \textbf{Dataset Type} & \textbf{Sampled Size} \\
        \midrule
        % \multirow{4}{*} 
        APDDv2~\cite{apddv2} & w / partial text caption  &  4,898 \\
        SPAQ~\cite{spaq} & w / partial text caption & 1,537 \\
        KonIQ-10k~\cite{koniq} & w / partial text caption & 1,488 \\
        Impressions~\cite{impressions} & w / partial text caption & 1,443 \\
        \midrule
        AVA~\cite{ava} & w / score caption & 235,598 \\
        TAD66K~\cite{tad66k} & w / score caption & 52,248 \\
        PARA~\cite{para} & w / score caption & 28,220 \\
        FLICKR-AES~\cite{flickr} & w / score caption & 35,762 \\
        \midrule
        \textbf{\textit{Total}} & -- & \textbf{361,194} \\
            % \hline
            % Impression~\cite{impressions} & Photographs & 1443 \\
            
        % \midrule
        % \multirow{3}{*}{} 
        %     & BAID [65]          & 2970 \\
        %     & CAD [31]           & 59   \\
        %     & ArtEmis [1]        & 1957 \\
        \bottomrule[1.5pt]
    \end{tabular}
\end{table}

% 对于只有aesthetic score的数据集，因为缺少不同维度的详细描述，所以我们只能用分数作为引导用MLLM生成对于图像的总体评价。我们采用如下的prompt模板:

\subsection{Datasets w/ Score Caption}

\noindent\textbf{AVA~\cite{ava}, TAD66K~\cite{tad66k}, PARA~\cite{para}, FLICKR-AES~\cite{flickr}.} For datasets containing only aesthetic scores without multi-dimensional annotations, we employ the scores as the primary guidance for MLLM to generate comprehensive image evaluations. The following prompt template is adopted:

\begin{promptbox}
An expert panel award this picture a \textit{<score>} score out of \textit{<range>} for aesthetic quality. 
Provide a concise, step-by-step aesthetic analysis evaluating its strengths and 
weaknesses in Composition \& Design, Visual Elements \& Structure, Technical Execution, Originality \& Creativity,  Theme \& Communication, Emotion \& Viewer Response, Overall Gestalt and Comprehensive Evaluation. 
\end{promptbox}

Here, \textit{\texttt{<score>}} and \textit{\texttt{<range>}} represents the scores and their value ranges, extracted from the dataset with score caption, serve as quantitative indicators to guide the MLLM's image analysis process.
% 代表从数据集中获取的分数及其数值范围，用于引导MLLM对于图像进行分析。

\subsection{Datasets w/ Partial Text Caption} 

% APDDv2数据集中包含了10023张图像，其属性包括:filename,Artistic Categories,Total aesthetic score,Theme and logic,Creativity,Layout and composition,Space and perspective,The sense of order,**Light and shadow,**Color,Details and texture,The overall,Mood,Language Comment，其中最重要的信息是Language Comment。我们从原始数据中过滤掉总评文字过短或者无的数据，剩余4898条。对于过滤后的数据，我们设计了如下prompt对包括总评，分类，各个子类的分数等重要信息进行了利用:
\noindent\textbf{APDDv2~\cite{apddv2}.} The APDDv2 dataset comprises 10,023 images, each annotated with 
multiple attributes, including: filename, Artistic Categories,  Total aesthetic score,  Theme and logic,  Creativity,  Layout and composition,  Space and perspective,  The sense of order,  Light and shadow,  Color,  Details and texture,  The overall,  Mood, and Language Comment (the most critical attribute for our study). We filter out samples with excessively short or missing Language Comment entries, retaining 4,898 valid instances. For the filtered data, we design a structured prompt template:

\begin{promptbox}
For the above picture, the artist gave the following evaluation: \textit{<language\_comment>}. For other aesthetic attributes: 

This image has a artistic category of \textit{<artistic\_categories>}. 

The total aesthetic score is \textit{<total\_aesthetic\_score>} out of 100.

The score for theme and logic is \textit{<theme\_and\_logic>} out of 10.

The score for creativity is \textit{<creativity>} out of 10.

The score for layout and composition is \textit{<layout\_and\_composition>} out of 10.

The score for space and perspective is \textit{<space\_and\_perspective>} out of 10.

The score for sense of order is \textit{<sense\_of\_order>} out of 10.

The score for light and shadow is \textit{<light\_and\_shadow>} out of 10.

The score for color is \textit{<color>} out of 10.

The score for details and texture is \textit{<details\_and\_texture>} out of 10.

The score for overall is \textit{<overall>} out of 10.

The score for mood is \textit{<mood>} out of 10.

Please combine the evaluation above with the picture content, then evaluate the aesthetic quality of this image from the attribute of \textit{<attribute>}. \textit{<description>}.
Limit the assessment to one paragraph (<=100 words), avoiding markdown formatting. Answer in English. Do not repeat contents in artist's evaluation (like scores).
\end{promptbox}

which incorporates key information such as the overall comment, category labels, and subcategory scores to ensure comprehensive utilization of the available annotations. Here, words enclosed in angle brackets (<>) denote referenced phrases or statements. For instance, \textit{\texttt{<language\_comment>}}, \textit{\texttt{<artistic\_categories>}}, \textit{\texttt{<total\_aesthetic\_score>}}, ..., and \textit{\texttt{<mood>}} refer to the corresponding captions in the dataset, while \textit{\texttt{<attribute>}} and \textit{\texttt{<description>}} represent the specific attribute and its description listed in Tab.~\ref{tab:attributes}.

% \verb||
% 这里带有<>的单词代表相应的引用词组或语句，例如\verb|<language\_comment>|, \verb|<artistic\_categories>|, \verb|<total\_aesthetic\_score>|, ... , \verb|<mood>|代指数据集中的相应caption，\verb|<attribute>| 和 \verb|<description>|则是Tab.1中的相应attribute及其description。

% 原始数据集中的图像属性包括EXIF tags，MOS，Image attribute scores，Scene category labels等属性。我们从中选择了与图像美学相关的属性，包括MOS分数和部分与美学相关的属性分数，并设计了如下的prompt模板:
% SPAQ共有11125张图像，我们从中进行过滤，约束为:(1)过滤后的数据中80%满足MOS分值和后4列分数均值（brightness, colorfulness, contrast, sharpness）落在[0,25]或者[75,100]区间，确保能够得到足够多的低美学质量和高美学质量的图像；(2)categories属性不是空

\noindent\textbf{SPAQ~\cite{spaq}.} 
The original dataset contains various image attributes, including EXIF tags, mean opinion scores (MOS), image attribute scores, and scene category labels. The SPAQ dataset comprises 11,125 images, which we filter according to two key criteria: (1) 80\% of the filtered subset must have either MOS (Mean Opinion Score) or the average of four quality metrics (brightness, colorfulness, contrast, and sharpness) falling within the extreme ranges of $[0,25]$ or $[75,100]$, ensuring sufficient representation of both low and high aesthetic quality samples; (2) all selected images must contain valid entries for the "categories" attribute. From these, we select attributes relevant to visual aesthetics—specifically, MOS ratings and a subset of aesthetic-related attribute scores—and designed the following prompt template:

\begin{promptbox}
The score for overall quality is \textit{<mos>} out of 100, with a high degree (if \textit{<mos>} > 75) / low degree (if \textit{<mos>} < 25) of aesthetic appeal. \\
The score for brightness is \textit{<brightness>} out of 100. \\
The score for colorfulness is \textit{<colorfulness>} out of 100. \\
The score for contrast is \textit{<contrast>} out of 100. \\
The score for sharpness is \textit{<sharpness>} out of 100. \\
The image content belongs to the following categories: \textit{<categories>}.
Please combine the evaluation above with the picture content, then evaluate the aesthetic quality of this image from the attribute of \textit{<attribute>}. \textit{<description>}.
Limit the assessment to one paragraph (<=100 words), avoiding markdown formatting. Answer in English. Do not repeat contents in artist's evaluation (like scores).
\end{promptbox}

Here, The classification into high-degree and low-degree categories is governed by the MOS threshold: instances with MOS > 75 are designated as high-degree, while those with MOS < 25 are categorized as low-degree. The placeholders \textit{\texttt{<mos>}}, \textit{\texttt{<brightness>}}, \textit{\texttt{<colorfulness>}}, ..., and \textit{\texttt{<categories>}} correspond to the respective captions from the SPAQ dataset, while \textit{\texttt{<attribute>}} and \textit{\texttt{<description>}} refer to the specific aesthetic attributes and their detailed descriptions as presented in Table~\ref{tab:attributes}.

% 其中high degree/low degree的选择依据\textit{\texttt{<mos>}}进行，如果\textit{\texttt{<mos>}}超过75则判定为high degree，如果\textit{\texttt{<mos>}}低于25则判定为low degree

% 其中\verb|<mos>|,\verb|<brightness>|,\verb|<colorfulness>|,...,  \verb|<categories>|来自SPAQ数据集的caption，while \verb|<attribute>| and \verb|<description>| represent the specific attribute and its description listed in Tab.~\ref{tab:attributes}.

% KonIQ-10K数据集中包含10,000条数据，我们从中挑选如下与美学相关的属性进行过滤:MOSz，brightness,contrast,colorfulness,sharpness,quality_factor，过滤条件为:过滤后的数据中80%满足MOSz分值落在[0,25]或者[75,100]区间，确保能够得到足够多的低美学质量和高美学质量的图像。最终我们得到了1488张过滤后图像，使用如下prompt引导MLLM进行标注:
\noindent\textbf{KonIQ-10K~\cite{koniq}.} The KonIQ-10K dataset comprises 10,000 images, from which we select the following aesthetic-relevant attributes for filtering: MOSz, brightness, contrast, colorfulness, sharpness, and quality\_factor. Our filtering criteria requires that 80\% of the selected images must have MOSz scores falling within either the $[0,25]$ or $[75,100]$ ranges, ensuring balanced representation of both low and high aesthetic quality samples. Through this process, we obtain 1,488 filtered images, which are then annotated by the MLLM using the following prompt template:

\begin{promptbox}
The score for overall quality is \textit{<MOSz>} out of 100, with a high degree (if \textit{<MOSz>} > 75) / low degree (if \textit{<MOSz>} < 25) of aesthetic appeal. \\
The score for brightness is \textit{<brightness>} out of 1. \\
The score for contrast is \textit{<contrast>} out of 1. \\
The score for colorfulness is \textit{<colorfulness>} out of 1. \\
The score for sharpness is \textit{<sharpness>} out of 100. \\
Please combine the evaluation above with the picture content, then evaluate the aesthetic quality of this image from the attribute of \textit{<attribute>}. \textit{<description>}.
Limit the assessment to one paragraph (<=100 words), avoiding markdown formatting. Answer in English. Do not repeat contents in artist's evaluation (like scores).
\end{promptbox}

Here, The classification into high-degree and low-degree categories is governed by the MOSz threshold: instances with MOSz > 75 are designated as high-degree, while those with MOSz < 25 are categorized as low-degree. The placeholders \textit{\texttt{<MOSz>}}, \textit{\texttt{<brightness>}}, \textit{\texttt{<contrast>}}, ..., and \textit{\texttt{<sharpness>}} correspond to the respective captions from the KonIQ-10K dataset, while \textit{\texttt{<attribute>}} and \textit{\texttt{<description>}} refer to the specific aesthetic attributes and their detailed descriptions as presented in Table~\ref{tab:attributes}.

% 数据集中原始1400+张图片，每张图片对应了多个annotator的回答（包括image_description，image_impression，image_aesthetic_eval三个方面），同时给出了这些annotator的个人信息（教育经历，美学经验等等），共有4800+条数据。我们按照如下条件进行过滤:同一张图片仅根据评估人的美学经验保留最有经验者的一条评价，最终过滤出1443条数据。
\noindent\textbf{Impressions~\cite{impressions}.} The original dataset contains over 1,400 images, each accompanied by multiple annotations (including image descriptions, impressions, and aesthetic evaluations) from different annotators, resulting in more than 4,800 data entries in total. Along with these annotations, Impressions also collects detailed annotator metadata such as educational background and aesthetic experience. To ensure annotation quality, we apply the following filtering criterion: for each image, we retain only the evaluation from the most aesthetically experienced annotator. This filtering process yields a refined dataset of 1,443 high-quality annotations, which are then annotated by the MLLM using the following prompt template:

\begin{promptbox}
This image's caption is: \textit{<caption>}.\\
What is happening in the image: \textit{<image\_description>}.\\
The emotions/thoughts/beliefs that the photograph may inspire: \textit{<image\_impression>}.\\
The aesthetic elements that elicited the expressed impression: \textit{<image\_aesthetic\_eval>}. \\
Please combine the evaluation above with the picture content, then evaluate the aesthetic quality of this image from the attribute of \textit{<attribute>}. \textit{<description>}.
Limit the assessment to one paragraph (<=100 words), avoiding markdown formatting. Answer in English. Do not repeat contents in artist's evaluation (like scores).
\end{promptbox}

Here, the placeholders \textit{\texttt{<caption>}}, \textit{\texttt{<image\_description>}}, \textit{\texttt{<image\_impression>}}, and \textit{\texttt{<image\_aesthetic\_eval>}} correspond to the respective captions from the Impressions dataset, while \textit{\texttt{<attribute>}} and \textit{\texttt{<description>}} refer to the specific aesthetic attributes and their detailed descriptions as presented in Table~\ref{tab:attributes}.

% 我们详细对比了不同的Score Prediction Strategy，实验结果展示在Tab.1中。所有的实验中都仅有预测分数的方式不同，训练数据、训练设置、模型结构等完全相同。各类strategy详细解释如下:
% 为了实验结论准确鲁棒，我们在AVA和ArtiMuse-10K上都进行了实验
\section{Details of Token As Score Strategy}
We conducted a comprehensive comparison of various score prediction strategies, and the experimental results are presented in Tab.~\ref{tab:score_prediction}. Across all experiments, the prediction score methodology was the sole differentiating factor, while the training data, training configurations, and model architecture remained consistent. To ensure robust and reliable experimental conclusions, we conduct comprehensive evaluations on both AVA (the largest image aesthetics scoring dataset)~\cite{ava} and ArtiMuse-10K (ours).

% follow qalign, 我们尝试了通过预测5个不同的离散level单词来预测分数。具体来说，我们按照如下方法将训练集中的分数转换为对应的level，然后让模型预测这些离散的level。推理时则通过计算模型在这五个level上的预测概率分布来加权求和得到最终的分数预测结果。
\subsection{Level As Score} 
Following Q-Align~\cite{qalign}, we predict scores by predicting five distinct discrete levels. Specifically, during training, we convert the continuous scores in the dataset into corresponding levels based on a predefined mapping and train the model to predict these discrete levels. This mapping scheme involves uniformly dividing the range between the maximum score (M) and the minimum score (m) into five distinct intervals, with scores within each interval being assigned to a corresponding discrete level:
\begin{equation}
L(s) = l_i \text{ if } m + \frac{i - 1}{5} \times (M - m) < s \leq m + \frac{i}{5} \times (M - m)
\end{equation}
where 
\begin{equation}
\{l_i\}_{i=1}^{5} = \{\text{bad, poor, fair, good, excellent}\}
\end{equation}
which are the standard text rating levels as defined by ITU~\cite{ITU}. During inference, the final score prediction was derived by computing a weighted sum of the predicted probability distribution across these five levels.

\noindent\textbf{Discussions.} The comparison between Exp. (a) and (i) in Tab.~\ref{tab:score_prediction}, along with other experimental groups, demonstrates that the Level As Score approach exhibits a significant performance degradation compared to the Token As Score. This decline can be attributed to the overly coarse-grained level partitioning scheme, which fails to achieve fine-grained score mapping. Furthermore, the adopted vocabulary lacks proper alignment with the LLM's lexical table design, collectively contributing to the suboptimal outcomes.

% 在表格的(a)和(k)及其他实验组的对比可以看出，Level As Score的做法性能相较于Token As Score方法有明显的下降，这是因为这里的level划分方法还是过于粗糙，无法细粒度的进行分数映射，同时所采用的单词也没有进行匹配与LLM词汇表的设计，因此导致了较差的结果。

% 在这一探索中，我们尝试扩展了llm的词汇表，向其中添加了额外的token用于aesthetics score的预测。例如对于Expanding 25 Tokens，最终我们会在词汇表中新增如下token:\verb|[AES_0]|,\verb|[AES_1]|, \verb|[AES_2]|,...\verb|[AES_25]|，分别对应于预测的分数0,4,8,...,100。我们同样训练模型去预测这些token，并在推理时通过其预测token的概率分布进行加权求和得到分数。
\subsection{Token As Score w/ Expanding Tokens} 
We provide a detailed exposition of the \textbf{\textit{Token As Score}} strategy, as referenced in the  Sec.~\ref{Sec:training_strategy}, Score Finetune of the main paper. In this investigation, we explore the expansion of the LLM vocabulary by incorporating additional tokens specifically for aesthetics score prediction. For instance, in the "Expanding 25 Tokens" configuration, we augment the vocabulary with the following tokens: \verb|[AES_SCORE_TOKEN_0]|, \verb|[AES_SCORE_TOKEN_1]|, \verb|[AES_SCORE_TOKEN_2]|, ..., \verb|[AES_SCORE_TOKEN_25]|. These tokens correspond to predicted scores of 0, 4, 8, ..., 100, respectively. The model is trained to predict these specialized tokens, and during inference, the final aesthetic score is derived by computing a weighted sum based on the predicted probability distribution over these tokens.

% 对比(b)(c)(d)(e)(f)实验可以发现，采用Token As Score策略时的效果随着引入token数目大致呈现先上升后下降的效果，在token数目达到100时效果最好。这是由于，token数目过少时，会导致无法建立准确的token-score映射；token数目过多时又会导致数据量或者模型容量本身不足以拟合。
% 在ArtiMuse-10K上可以看到Token As Score w/ Expanding Tokens的实验表现都很差，说明数据本身难度较高或数量不足够多时这种方法无法正常收敛。
\noindent\textbf{Discussions.} A comparison of experiments (b)-(f) on AVA reveals that the performance of the Token As Score strategy initially improves and then declines as the number of introduced tokens increases, peaking at 100 tokens. This trend occurs because an insufficient number of tokens fails to establish an accurate token-score mapping, while an excessive number exceeds the available data or model capacity, leading to underfitting. Experimental results on ArtiMuse-10K demonstrate that the Token As Score approach with expanding tokens performs poorly, suggesting this method fails to converge properly when either the dataset is inherently challenging or insufficient in size.

% 我们为在main paper的Sec.4.3的Score Finetune部分提到的Token As Score策略提供一个详细说明。这里我们尝试从LLM已有的可显示token中选择一部分用于aesthetics score prediction。在选择这些已有token时，我们尽量保证他们简短、有序、容易收敛且不会和分数本身产生歧义。如Tab.1所示，我们的各类设置具体如下:
% Existing 5 Tokens: 选择token为a,b,c,d,e，依次对应分数0，25，50，75，100
% Existing 25 Tokens: 选择token为a,b,c,...，y,依次对应分数0,4,8,...,100
% Existing 50 Tokens: 选择token为a,b,c,...，y,A,B,C,...,Y,依次对应分数0,2,4...,100
% Existing 100 Tokens (non-ordered): 选择token为qwen2.5-7B LLM词汇表中从字符0开始的100个字符，展示于Tab.2中,依次对应分数0,1,2，..,100
% Existing 100 Tokens (ordered): 这是ArtiMuse中的最终方案。我们使用小写字母拼接出100个在qwen2.5-7B LLM词汇表中按顺序排列的token，展示于Tab.3中,依次对应分数0,1,2，..,100
\subsection{Token As Score w/ Existing Tokens} 
We futher explore the selection of a subset of the LLM's existing displayable tokens for aesthetics score prediction. Our selection criteria prioritize brevity, inherent order, ease of convergence during training, and minimal ambiguity with numerical scores. As illustrated in Tab.~\ref{tab:score_prediction}, our specific configurations in experiments are as follows:

\noindent\textbf{Existing 25 Tokens.} We select the tokens \verb|a|, \verb|b|, \verb|c|, ... , \verb|y|, which are sequentially mapped to scores ranging from 0 to 100 with an interval of 4 (i.e., 0, 4, 8, ..., 100).

\noindent\textbf{Existing 50 Tokens.} We select the tokens \verb|a|, \verb|b|, \verb|c|, ... , \verb|y|, \verb|A|, \verb|B|, \verb|C|, ... , \verb|Y|, which are sequentially mapped to scores ranging from 0 to 100 with an interval of 2 (i.e., 0, 2, 4, ..., 100).

\noindent\textbf{Existing 100 Tokens (non-ordered).} We select the first 100 character tokens starting from \verb|0| within the vocabulary of the Qwen2.5-7B LLM, as detailed in Tab.~\ref{tab:token_mapping_100_no_order}. These tokens are sequentially mapped to scores from 0 to 100.

% \begin{promptbox}
%     (1) 0-0 (1) 0-0 (1) 0-0 (1) 0-0 (1) 0-0 
% \end{promptbox}

% 

\begin{table}[htbp]
\centering
\caption{Token-score mapping table for existing 100 tokens (non-ordered).}
\label{tab:token_mapping_100_no_order}
\resizebox{\textwidth}{!}{
\begin{tabular}{l|*{20}{c}}
\hline
% \multicolumn{21}{c}{Token Mapping Table (Part 1)} \\
% \hline
\textbf{Token ID} & 15 & 16 & 17 & 18 & 19 & 20 & 21 & 22 & 23 & 24 & 25 & 26 & 27 & 28 & 29 & 30 & 31 & 32 & 33 & 34 \\
\textbf{Token} & \texttt{0} & \texttt{1} & \texttt{2} & \texttt{3} & \texttt{4} & \texttt{5} & \texttt{6} & \texttt{7} & \texttt{8} & \texttt{9} & \texttt{:} & \texttt{;} & \texttt{<} & \texttt{=} & \texttt{>} & \texttt{?} & \texttt{@} & \texttt{A} & \texttt{B} & \texttt{C} \\
\textbf{Score} & 0 & 1 & 2 & 3 & 4 & 5 & 6 & 7 & 8 & 9 & 10 & 11 & 12 & 13 & 14 & 15 & 16 & 17 & 18 & 19 \\
\hline
\end{tabular}}

% \vspace{5mm}

\resizebox{\textwidth}{!}{
\begin{tabular}{l|*{20}{c}}
\hline

\textbf{Token ID} & 35 & 36 & 37 & 38 & 39 & 40 & 41 & 42 & 43 & 44 & 45 & 46 & 47 & 48 & 49 & 50 & 51 & 52 & 53 & 54 \\
\textbf{Token} & \texttt{D} & \texttt{E} & \texttt{F} & \texttt{G} & \texttt{H} & \texttt{I} & \texttt{J} & \texttt{K} & \texttt{L} & \texttt{M} & \texttt{N} & \texttt{O} & \texttt{P} & \texttt{Q} & \texttt{R} & \texttt{S} & \texttt{T} & \texttt{U} & \texttt{V} & \texttt{W} \\
\textbf{Score} & 20 & 21 & 22 & 23 & 24 & 25 & 26 & 27 & 28 & 29 & 30 & 31 & 32 & 33 & 34 & 35 & 36 & 37 & 38 & 39 \\
\hline
\end{tabular}}

% \vspace{5mm}

\resizebox{\textwidth}{!}{
\begin{tabular}{l|*{20}{c}}
\hline

\textbf{Token ID} & 55 & 56 & 57 & 58 & 59 & 60 & 61 & 62 & 63 & 64 & 65 & 66 & 67 & 68 & 69 & 70 & 71 & 72 & 73 & 74 \\
\textbf{Token} & \texttt{X} & \texttt{Y} & \texttt{Z} & \texttt{[} & \texttt{\textbackslash} & \texttt{]} & \texttt{\^{}} & \texttt{\_} & \texttt{\`} & \texttt{a} & \texttt{b} & \texttt{c} & \texttt{d} & \texttt{e} & \texttt{f} & \texttt{g} & \texttt{h} & \texttt{i} & \texttt{j} & \texttt{k} \\
\textbf{Score} & 40 & 41 & 42 & 43 & 44 & 45 & 46 & 47 & 48 & 49 & 50 & 51 & 52 & 53 & 54 & 55 & 56 & 57 & 58 & 59 \\
\hline
\end{tabular}}

% \vspace{5mm}

\resizebox{\textwidth}{!}{
\begin{tabular}{l|*{20}{c}}
\hline

\textbf{Token ID} & 75 & 76 & 77 & 78 & 79 & 80 & 81 & 82 & 83 & 84 & 85 & 86 & 87 & 88 & 89 & 90 & 91 & 92 & 93 & 94 \\
\textbf{Token} & \texttt{l} & \texttt{m} & \texttt{n} & \texttt{o} & \texttt{p} & \texttt{q} & \texttt{r} & \texttt{s} & \texttt{t} & \texttt{u} & \texttt{v} & \texttt{w} & \texttt{x} & \texttt{y} & \texttt{z} & \texttt{\{} & \texttt{|}| & \texttt{\}} & \texttt{\textasciitilde} & \texttt{¡} \\
\textbf{Score} & 60 & 61 & 62 & 63 & 64 & 65 & 66 & 67 & 68 & 69 & 70 & 71 & 72 & 73 & 74 & 75 & 76 & 77 & 78 & 79 \\
\hline
\end{tabular}}

% \vspace{5mm}

\resizebox{\textwidth}{!}{
\begin{tabular}{l|*{21}{c}}
\hline

\textbf{Token ID} & 95 & 96 & 97 & 98 & 99 & 100 & 101 & 102 & 103 & 104 & 105 & 106 & 107 & 108 & 109 & 110 & 111 & 112 & 113 & 114 & 115 \\
\textbf{Token} & \texttt{¢} & \texttt{£} & \texttt{¤} & \texttt{¥} & \texttt{¦} & \texttt{§} & \texttt{¨} & \texttt{©} & \texttt{ª} & \texttt{«} & \texttt{¬} & \texttt{®} & \texttt{¯} & \texttt{°} & \texttt{±} & \texttt{²} & \texttt{³} & \texttt{´} & \texttt{µ} & \texttt{¶} & \texttt{·} \\
\textbf{Score} & 80 & 81 & 82 & 83 & 84 & 85 & 86 & 87 & 88 & 89 & 90 & 91 & 92 & 93 & 94 & 95 & 96 & 97 & 98 & 99 & 100 \\
\hline
\end{tabular}}
\end{table}

\noindent\textbf{Existing 100 Tokens (ordered).} This represents the final approach adopted in ArtiMuse. We construct 100 tokens by concatenating lowercase letters, ensuring these tokens are ordered within the vocabulary of the Qwen2.5-7B LLM, as presented in Tab.~\ref{tab:token_mapping_100_order}. These tokens are sequentially mapped to scores from 0 to 100.

\begin{table}[htbp]
\centering
\caption{Token-score mapping table for existing 100 tokens (ordered), which is used in ArtiMuse.}
\label{tab:token_mapping_100_order}
\small  % Make font smaller to fit more content

% Table 1: Scores 0-19
\resizebox{\textwidth}{!}{%
\begin{tabular}{l|*{20}{c}}
\hline
\textbf{Token} & \texttt{aa} & \texttt{ab} & \texttt{ac} & \texttt{ad} & \texttt{ae} & \texttt{af} & \texttt{ag} & \texttt{ah} & \texttt{ai} & \texttt{aj} & \texttt{ak} & \texttt{al} & \texttt{am} & \texttt{an} & \texttt{ao} & \texttt{ap} & \texttt{aq} & \texttt{ar} & \texttt{as} & \texttt{at} \\
\textbf{Score} & 0 & 1 & 2 & 3 & 4 & 5 & 6 & 7 & 8 & 9 & 10 & 11 & 12 & 13 & 14 & 15 & 16 & 17 & 18 & 19 \\
\hline
\end{tabular}%
}

% Table 2: Scores 20-39
\resizebox{\textwidth}{!}{%
\begin{tabular}{l|*{20}{c}}
\hline
\textbf{Token} & \texttt{au} & \texttt{av} & \texttt{aw} & \texttt{ax} & \texttt{ay} & \texttt{az} & \texttt{ca} & \texttt{cb} & \texttt{cc} & \texttt{cd} & \texttt{ce} & \texttt{cf} & \texttt{cg} & \texttt{ch} & \texttt{ci} & \texttt{cj} & \texttt{ck} & \texttt{cl} & \texttt{cm} & \texttt{cn} \\
\textbf{Score} & 20 & 21 & 22 & 23 & 24 & 25 & 26 & 27 & 28 & 29 & 30 & 31 & 32 & 33 & 34 & 35 & 36 & 37 & 38 & 39 \\
\hline
\end{tabular}%
}

% Table 3: Scores 40-59
\resizebox{\textwidth}{!}{%
\begin{tabular}{l|*{20}{c}}
\hline
\textbf{Token} & \texttt{co} & \texttt{cp} & \texttt{cq} & \texttt{cr} & \texttt{cs} & \texttt{ct} & \texttt{cu} & \texttt{cv} & \texttt{cw} & \texttt{cx} & \texttt{cy}  & \texttt{da} & \texttt{db} & \texttt{dc} & \texttt{dd} & \texttt{de} & \texttt{df} & \texttt{dg} & \texttt{dh} & \texttt{di} \\
\textbf{Score} & 40 & 41 & 42 & 43 & 44 & 45 & 46 & 47 & 48 & 49 & 50 & 51 & 52 & 53 & 54 & 55 & 56 & 57 & 58 & 59 \\
\hline
\end{tabular}%
}

% Table 4: Scores 60-79
\resizebox{\textwidth}{!}{%
\begin{tabular}{l|*{20}{c}}
\hline
\textbf{Token}  & \texttt{dj} & \texttt{dk} & \texttt{dl} & \texttt{dm} & \texttt{dn} & \texttt{do} & \texttt{dp} & \texttt{dq} & \texttt{dr} & \texttt{ds} & \texttt{dt} & \texttt{du} & \texttt{dv} & \texttt{dw} & \texttt{dx} & \texttt{dy} & \texttt{ea} & \texttt{eb} &  \texttt{ec} & \texttt{ed}\\
\textbf{Score} & 60 & 61 & 62 & 63 & 64 & 65 & 66 & 67 & 68 & 69 & 70 & 71 & 72 & 73 & 74 & 75 & 76 & 77 & 78 & 79 \\
\hline
\end{tabular}%
}

% Table 5: Scores 80-100
\resizebox{\textwidth}{!}{%
\begin{tabular}{l|*{21}{c}}
\hline
\textbf{Token}  & \texttt{ee} & \texttt{ef} & \texttt{eg} & \texttt{eh} & \texttt{ei} & \texttt{ej} & \texttt{ek} & \texttt{el} & \texttt{em} & \texttt{en} & \texttt{eo} & \texttt{ep} & \texttt{eq} & \texttt{er} & \texttt{es} & \texttt{et} & \texttt{eu} & \texttt{ev} & \texttt{ew} & \texttt{ex}  & \texttt{ey}\\
\textbf{Score} & 80 & 81 & 82 & 83 & 84 & 85 & 86 & 87 & 88 & 89 & 90 & 91 & 92 & 93 & 94 & 95 & 96 & 97 & 98 & 99 & 100 \\
\hline
\end{tabular}%
}

\end{table}

\noindent\textbf{Discussions.} 
The comparisons in (b)-(g), (c)-(h), and (d)-(j) demonstrate that when using the same number of tokens for prediction in the Token As Score, tokens from the existing vocabulary consistently yield better performance. This occurs because newly introduced tokens lack corresponding prior knowledge from the model's pretraining phase and do not possess inherent ordinal relationships with scores, making them less effective than tokens in the LLM vocabulary that carry clear semantic information and sequential relationships.

Furthermore, experiments (g), (h), and (j) reveal that when using existing tokens for Token As Score, model performance improves significantly as the number of tokens increases. Due to the limited number of displayable characters in the Qwen2.5-7B LLM vocabulary, we are currently unable to further increase this quantity, which will be explored in future work. Additionally, comparing (i) and (j) shows that the choice of tokens also affects performance—the token mapping scheme in (j), which has more explicit semantic and ordinal relationships, leads to better results.

% \multirow{2}{*}{\textbf{Model}} & \multicolumn{2}{c}{\textbf{AVA}~\cite{ava}} & \multicolumn{2}{c}{\textbf{PARA}~\cite{para}} & \multicolumn{2}{c}{\textbf{TAD66K}~\cite{tad66k}} & \multicolumn{2}{c}{\textbf{FLICKR-AES}~\cite{flickr}} & \multicolumn{2}{c}{\textbf{ArtiMuse-10K}} \\
% \cmidrule(lr){2-3} \cmidrule(lr){4-5} \cmidrule(lr){6-7} \cmidrule(lr){8-9} \cmidrule(lr){10-11}
%  & SRCC & PLCC & SRCC & PLCC & SRCC & PLCC & SRCC & PLCC & SRCC & PLCC \\

% 除了ArtiMuse-10K上expanding的策略收敛情况存在问题外，100 tokens的效果在各种数目的token中处于峰值

\begin{table}[h!]
    \centering
    \caption{Explorations on score prediction strategies. To ensure experimental validity, we conduct our experiments both on the AVA dataset and AriMuse-10K dataset. (j) represents the setting of Token As Score strategy in ArtiMuse. Beyond the convergence issues observed with the expanding strategy on ArtiMuse-10K, the 100-token configuration demonstrates peak performance across various token quantities.}
    \label{tab:score_prediction}
    \begin{tabular}{ll|cccc}
        \toprule[1pt]
        \multirow{2}{*}{\textbf{Exp.}} & \multirow{2}{*}{\textbf{Score Prediction}} & \multicolumn{2}{c}{\textbf{AVA}~\cite{ava}} &  \multicolumn{2}{c}{\textbf{ArtiMuse-10K}} \\ \cmidrule(lr){3-4} \cmidrule(lr){5-6}
        & & SRCC & PLCC & SRCC & PLCC \\
        \midrule
        (a) & 5 Levels & 0.820 & 0.818 & 0.571 & 0.551 \\
        \midrule
        (b) & Expanding 25 Tokens & 0.803 & 0.665 & 0.045 & 0.055 \\
        (c) & Expanding 50 Tokens & 0.822 & 0.821 & 0.018 & 0.027\\
        \rowcolor{gray!30} (d) & Expanding 100 Tokens & 0.824 & 0.822 & 0.029 & 0.027\\ 
        (e) & Expanding 250 Tokens & 0.823 & 0.821 & -0.012 & 0.002 \\
        (f) & Expanding 500 Tokens & 0.821 & 0.819 & 0.006 & 0.012\\
        \midrule
        % (g) & Existing 5 Tokens & 0.823 & 0.821 \\
        (g) & Existing 25 Tokens & 0.823 & 0.822 & 0.006 & 0.010 \\
        (h) & Existing 50 Tokens & 0.825 & 0.824 & 0.612 & 0.623 \\
        (i) & Existing 100 Tokens (non-ordered) & 0.826 & 0.825 & 0.582 & 0.541 \\
         \rowcolor{gray!30} (j) & Existing 100 Tokens (ordered) & \textbf{0.827} & \textbf{0.826} & \textbf{0.614} & \textbf{0.627} \\
        \bottomrule[1pt]
    \end{tabular}
\end{table}

% \begin{table}[h!]
%     \centering
%     \caption{More ablations on score prediction strategies. To ensure experimental validity, we conduct our training on the AVA dataset, the largest available benchmark for aesthetics scoring.  (k) represents the setting of Token As Score in ArtiMuse.}
%     \label{tab:score_prediction}
%     \begin{tabular}{ll|cc}
%         \toprule[1pt]
%         \textbf{Exp.} & \textbf{Score Prediction} & \textbf{SRCC} & \textbf{PLCC} \\ 
%         \midrule
%         (a) & 5 Levels & 0.820 & 0.818 \\
%         \midrule
%         (b) & Expanding 25 Tokens & 0.823 & 0.821 \\
%         (c) & Expanding 50 Tokens & 0.822 & 0.821 \\
%         (d) & Expanding 100 Tokens & 0.823 & 0.821 \\
%         (e) & Expanding 250 Tokens & 0.824 & 0.822 \\
%         (f) & Expanding 500 Tokens & 0.822 & 0.819 \\
%         \midrule
%         % (g) & Existing 5 Tokens & 0.823 & 0.821 \\
%         (g) & Existing 25 Tokens & 0.823 & 0.822 \\
%         (h) & Existing 50 Tokens & 0.825 & 0.824 \\
%         (i) & Existing 100 Tokens (non-ordered) & 0.826 & 0.825 \\
%         (j) & Existing 100 Tokens (ordered) & \textbf{0.827} & \textbf{0.826} \\
%         \bottomrule[1pt]
%     \end{tabular}
% \end{table}

% simple vs complex
\section{Implementation Details}

\subsection{Training Details}

% 我们采用InternVL-3-8B模型作为base model进行训练，并根据aesthetic Assessment任务的特点，在text pretrain和score finetune部分对超参数做了一些调整，详见Tab.1
\noindent\textbf{Hyperparameters.} 
We employ the InternVL-3-8B~\cite{internvl3} model as our base model and adopt its default hyperparameters for the aesthetic assessment task through two training stages: Text Pretrain and Score Finetune. The pre-trained models and specific hyperparameter configurations are detailed in Table~\ref{tab:hyperparameters}, with modifications carefully designed to address the unique requirements of visual aesthetic evaluation.

\begin{table}[h!]
\centering
\caption{Pre-trained models and hyperparameters used for ArtiMuse, including text pretraining and score finetuning.}
\label{tab:hyperparameters}
\resizebox{\textwidth}{!}{
\begin{tabular}{l|c c}
\toprule
\textbf{Pre-trained models / Hyperparameters} & \textbf{Text Pretrain} & \textbf{Score Finetune} \\
\midrule
Vison Encoder & InternViT-300M-448px-V2.5 & InternViT-300M-448px-V2.5 \\
Large Language Model & Qwen2.5-7B & Qwen2.5-7B \\
Large Language Model LoRA Rank &  16 & 128 \\
Image Resolution & $448 \times 448$ & $448 \times 448$ \\
Max Sequence Length & 8192 & 8192 \\
Batch Size & 128 & 128 \\
Warmup Epochs & 0.03 & 0.03   \\
Gradient Accuracy & 1 & 1 \\
Numerical Precision & Float16 & Float16 \\
LR Schedule & Cosine decay & Cosine decay \\
LR Max & 4e-5 & 2e-5 \\
Weight Decay & 0.05 & 0 \\
Epoch & 1 & 2 \\
\bottomrule
\end{tabular}}
\end{table}

% \begin{table}[h!]
% \centering
% \caption{Pre-trained models and hyperparameters used for ArtiMuse, including text pretraining and score finetuning.}
% \label{tab:hyperparameters}
% \begin{tabular}{l|c c}
% \toprule
% \textbf{Pre-trained models / Hyperparameters} & \textbf{Text Pretrain} & \textbf{Score Finetune} \\
% \midrule
% Vison Encoder & \multicolumn{2}{c}{InternViT-300M-448px-V2.5} \\
% Large Language Model & \multicolumn{2}{c}{Qwen2.5-7B} \\
% Large Language Model LoRA Rank &  16 & 128 \\
% Image resolution & \multicolumn{2}{c}{$448 \times 448$} \\
% Max Sequence Length & \multicolumn{2}{c}{8192} \\
% Batch size & \multicolumn{2}{c}{128} \\
% Warmup epochs & \multicolumn{2}{c}{0.03} \\
% Gradient acc. & \multicolumn{2}{c}{1} \\
% Numerical precision & \multicolumn{2}{c}{Float16} \\
% LR schedule & \multicolumn{2}{c}{cosine decay} \\
% LR max & 4e-5 & 2e-5 \\
% Weight decay & 0.05 & 0 \\
% Epoch & 1 & 2 \\
% \bottomrule
% \end{tabular}
% \end{table}

% 如表格所示，我们在训练过程中采用了fixed image resolution 策略，
\noindent\textbf{Resolution Strategy.} 
The original InternVL-3 model employs a dynamic high-resolution strategy~\cite{internvl3} to handle images of varying resolutions and attribute ratios. This approach involves three key steps: closest attribute ratio matching, image resizing and splitting, and optional thumbnail generation. Given an input image with dimensions \( W \times H \), the aspect ratio \( r = W/H \) is computed. The algorithm selects a target aspect ratio \( r_{\text{best}} \) from a predefined set \( \mathcal{R} \), which minimizes distortion while constraining the number of tiles \( n_{\text{tiles}} \) within a range \([n_{\text{min}}, n_{\text{max}}]\). The image is resized to dimensions \( S \times i_{\text{best}} \times S \times j_{\text{best}} \) (where \( S = 448 \)) and split into \( n_{\text{tiles}} = i_{\text{best}} \times j_{\text{best}} \) tiles of size \( S \times S \). If \( n_{\text{tiles}} > 1 \), a thumbnail of size \( S \times S \) is appended to preserve a global view.  

However, in ArtiMuse, we adopt a fixed-resolution strategy instead of the dynamic approach. Aesthetic evaluation relies heavily on holistic image features, such as composition, color harmony, and spatial relationships, which can be disrupted by splitting an image into localized tiles. The dynamic strategy’s tile-based processing risks fragmenting these global characteristics, thereby degrading performance in tasks requiring an integrated understanding of visual aesthetics. By resizing all images to a uniform resolution without tiling, we preserve the structural and semantic coherence of the entire image. This adjustment ensures that the model captures aesthetic qualities through a consistent, undistorted representation of the input, aligning better with the requirements of fine-grained aesthetic analysis. Our experiments demonstrate that employing the fixed-resolution strategy yields approximately 0.3 improvements in both SRCC and PLCC metrics for aesthetic scoring tasks compared to the dynamic high-resolution strategy, while simultaneously more than doubling training and inference efficiency.
% 我们的实验表明，使用fixed-resolution strategy相较于dynamic high-resolution strategy能够在aesthetics scoring任务上取得SRCC/PLCC上0.3左右的提升，且训练和推理效率也能提升超过一倍。

\subsection{Inference Details for Aesthetics Scoring} 
We present the implementation details for various models in the aesthetic scoring task. Note that certain models—including TANet~\cite{tad66k}, AesMamba~\cite{AesMamba}, UNIAA-LLaVA~\cite{Uniaa}, and Next Token Is Enough~\cite{nexttoken}—are excluded from this discussion due to testing constraints.

% 我们介绍在Aesthetics scoring任务上各个模型的Inference Details. TANet~\cite{tad66k}, AesMamba~\cite{AesMamba}, UNIAA-LLaVA~\cite{Uniaa}, Next Token Is Enough~\cite{nexttoken}等无法测试的模型在此不进行讨论。

% 对于有评分能力的模型，我们直接调用模型对图像进行评分。对于其中部分只能从general角度进行评分的模型，我们将其general评分作为结果。
\noindent\textbf{Models w/ Scoring Ability.} 
For models capable of generating aesthetic scores (Q-Instruct~\cite{qinstruct}, PEAS~\cite{peas}, Q-Align~\cite{qalign}), we directly utilize their scoring outputs. In cases where a model provides only general assessments (MUSIQ~\cite{musiq}), we adopt its general score as the final evaluation result. 

% 对于没有评分能力的模型，我们使用特殊的prompt引导其进行评分。设计的prompt如下:这样可以引导模型输出0-100范围内的整数分值，和ArtiMuse的评分输出形式一致，我们将此结果用于比较。
% 
\noindent\textbf{Models w/o Scoring Ability.} 
For models lacking inherent scoring capabilities (VILA~\cite{vila}, mPLUG-Owl2~\cite{mPLUG-Owl2}, ShareGPT-4V~\cite{sharegpt4v}, Qwen-2.5-VL-7B~\cite{qwen2.5}, InternVL3-8B~\cite{internvl3}), we employ carefully designed prompts to elicit numerical evaluations. The prompt structure is as follows:

\begin{promptbox}
    Please rate the aesthetic quality of this image and provide a score between 0 and 100, where 0 represents the lowest quality and 100 represents the highest. Your response should contain only an integer value.
\end{promptbox}

This prompt guides the model to output an integer score from 0 to 100, aligning with ArtiMuse’s scoring format. We use these prompted scores for comparative analysis, ensuring consistency across all evaluated models.

% 在测试模型的Textual Analysis能力时，我们为其他的模型设计了特殊的prompt，为他们提供相应的美学背景知识以保证公平。具体来说，对于ArtiMuse，我们测试时使用了如下的prompt:对于其他模型，我们则添加了相应的aspect description:更多的文字测试结果请见Sec.~\ref{sec:further_text}
\subsection{Inference Details for Textual Analysis} 
When evaluating the model's textual analysis capability, we design specialized prompts for comparative models by incorporating relevant aesthetic background knowledge to ensure fairness. Specifically, for ArtiMuse, we employ the following prompt format during testing:

\begin{promptbox}
    Please evaluate the aesthetic quality of this image from the attribute of \textit{<attribute>}.
\end{promptbox}

where \textit{\texttt{<attribute>}} represents the specific attribute listed in Tab.~\ref{tab:attributes}. For other models, we augment their inputs with corresponding attribute descriptions to maintain parity in contextual understanding:

\begin{promptbox}
    Background Knowledge: \textit{<attribute>}: \textit{<description>}. Please evaluate the aesthetic quality of this image from the attribute of \textit{<attribute>}. No more than 100 words.
\end{promptbox}

where \textit{\texttt{<attribute>}} and \textit{\texttt{<description>}} represent the specific attribute and its description listed in Tab.~\ref{tab:attributes}. Additional textual evaluation results and analysis are presented in Section~\ref{sec:further_text}.

\subsection{Comparison Details} 

\begin{figure}[h]
  \centering
    \includegraphics[width=0.99\linewidth]{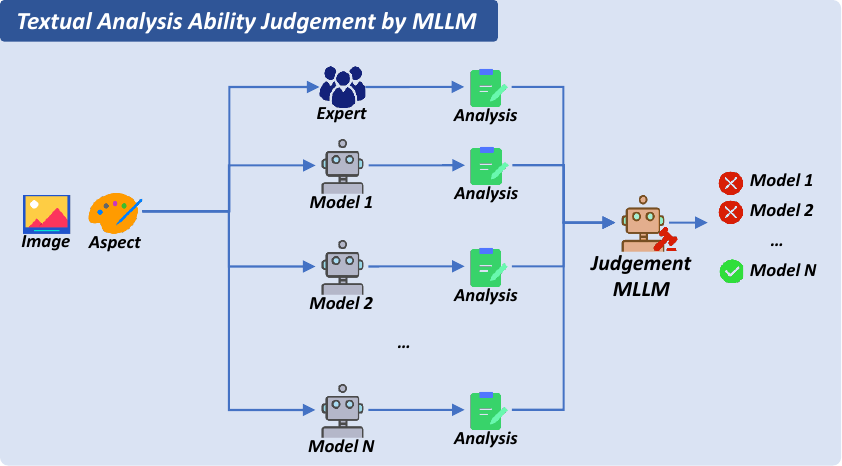}
   % \vspace{-10pt}
   \caption{Pipeline of the structural aesthetic analysis ability judgment by MLLM.}
   \label{fig:judgement}
   % \vspace{-5pt}
\end{figure}

\noindent\textbf{Judging by MLLM.}
We provide a detailed explanation of the methodology employed in Sec.~\ref{Sec:structural_aesthetic_analysis} of the main paper for using MLLMs to select among different models' structural aesthetic analysis results. As illustrated in Fig.~\ref{fig:judgement}, we first determine the input image and the corresponding aesthetic attributes, then guide the MLLM to generate textual evaluations using the following prompt template:

\begin{promptbox}
    You are an aesthetic evaluation expert. Please evaluate the aesthetic quality of this image from the attribute of \textit{<attribute>}. No more than 100 words.
\end{promptbox}

where \textit{\texttt{<attribute>}} corresponds to the specific aesthetic attributes listed in Tab.~\ref{tab:attributes}. For human experts, we also provide the attribute and invite them to provide textual evaluations. The image, attribute, expert evaluations, and the outputs from different models are then fed into a judgment MLLM (specifically, Gemini-2.0-flash) for assessment. We guide this MLLM to evaluate and select the highest-quality responses among the model outputs using a single-choice question format prompt (Taking 4 models as an example):

\begin{promptbox}
    You are an expert aesthetic evaluation judge. Your task is to evaluate the aesthetic analysis quality of each model's response, based on its alignment with the given human expert critique.
    There are four model-generated responses: 
    model1, model2, model3, and model4.
    Assess them independently for clarity, accuracy, insightfulness, and relevance, and identify the single best response overall.
    Output only the identifier of the best model (i.e., one of: model1, model2, model3, model4) — do not include any extra text, explanation, symbols, or formatting.
\end{promptbox}

which minimizes hallucinations, provides sufficient information for decision-making, and ensures consistent evaluation criteria across all model responses, thereby yielding relatively accurate and stable selection outcomes. The results is presented in Tab.~\ref{tab:image-evaluation} of the main paper.

% 我们详细说明Sec 5.2, main paper中采用的用MLLM对不同模型的strutural aesthetic analysis结果作选择的具体做法。如图所示，我们确定好输入图片和相应的美学维度，然后通过如下prompt引导模型输出文字评价:其中<attribute>对应于Tab.1中的相应aesthetic attributes。对于专家，我们也告知相应的attribute名称并邀请他们给出文字评价。我们将图片，attribute，专家评价和不同模型的评价一起输入到用于输出judgment的MLLM中，这里我们采用的是Gemini-2.0-flash. 我们通过如下prompt引导MLLM对于不同模型的回答质量进行评定和选择:这里我们采用单选题的形式，尽可能减少MLLM的幻觉，给予了MLLM足够的用于判定的信息，同时保证对于每个模型回答的评判标准相同，因此可以得到相对准确和稳定的选择结果。具体结构详见Tab.2,page 7 of the main paper.

% 在user study方面，我们从ArtiMuse-10K的测试集中随机选择了20张图像，保证他们覆盖了不同的categories和不同的美学质量，然后使用不同模型对每张图像从8 aesthetic attributes进行评价，记录其输出结果。我们将测试结果制作成20个选择题，每道题对应一张图像及不同模型在其某个attribute上的评价结果，并补充了关于相应attribute的详细描述作为背景。我们邀请了20位volunteers参与测试，受试者中既包含未经过专业训练的人，也包含有丰富美学评估经验的人，我们收集他们的答案并统计了各个模型被选择的比例。结果展示于Tab.2,page 7 of the main paper.
\noindent\textbf{Judging by Human.} For the user study, we randomly select 20 images from the ArtiMuse-10K test set, ensuring coverage across different categories and varying aesthetic qualities. Each image is evaluated by different models across 8 aesthetic attributes, with their outputs recorded. We compile these results into 20 multiple-choice questions, where each question corresponds to one image and the model-generated evaluations for a specific attribute, supplemented by a detailed description of that attribute for context. We recruit 20 volunteers, including both individuals without formal training and those with extensive aesthetic evaluation experience, to participate in the study. Their selections are collected, and the preference rates for each model are computed. The results are presented in Tab.~\ref{tab:image-evaluation} of the main paper.

\section{More Results}

% 我们将ArtiMuse和当前MLLM领域最先进的多模态大模型进行了比较，开源模型选择了Qwen-2.5-VL-72B-instruct~\cite{qwen2.5}和InternVL3-78B~\cite{internvl3}，闭源模型选择了GPT-4o~\cite{gpt4}和Gemini-2.0-flash~\cite{gemini}，结果展示在Tab.~\ref{tab:score_comp}中，可以看到闭源模型如GPT-4o比开源模型更优，但是ArtiMuse作为仅有8B的模型仍然在aesthetics score任务上比这些最先进的MLLM有显著的优势。
\subsection{Comparison with SOTA Open-Source \&  Closed-Source MLLMs}
We benchmark ArtiMuse against state-of-the-art multimodal large language models (MLLMs), including both open-source (Qwen-2.5-VL-72B-instruct~\cite{qwen2.5} and InternVL3-78B~\cite{internvl3}) and closed-source models (GPT-4o~\cite{gpt4} and Gemini-2.0-Flash~\cite{gemini}). As shown in Tab.~\ref{tab:score_comp_supp}, closed-source models generally outperform open-source models. Notably, ArtiMuse achieves significantly higher performance in aesthetics scoring than these leading MLLMs despite having only 8B parameters, demonstrating its exceptional capability in image aesthetic assessment.

\begin{table}[htbp]
% \vspace{-10pt}
\centering
\caption{More comparison on aesthetics scoring. The best and second-best performances are highlighted in \textcolor{red}{red} and \textcolor{blue}{blue}, respectively. ArtiMuse demonstrates superior performance when compared to various state-of-the-art open-source \&  closed-source MLLMs.}
\label{tab:score_comp_supp}
\resizebox{\textwidth}{!}{%
% \fontsize{8}{9}\selectfont
% \renewcommand{\arraystretch}{0.8} % 调整行距，值越小行距越窄
\begin{tabular}{lcccccccccc}
\toprule
\multirow{2}{*}{\textbf{Model}} & \multicolumn{2}{c}{\textbf{AVA}~\cite{ava}} & \multicolumn{2}{c}{\textbf{PARA}~\cite{para}} & \multicolumn{2}{c}{\textbf{TAD66K}~\cite{tad66k}} & \multicolumn{2}{c}{\textbf{FLICKR-AES}~\cite{flickr}} & \multicolumn{2}{c}{\textbf{ArtiMuse-10K}} \\
\cmidrule(lr){2-3} \cmidrule(lr){4-5} \cmidrule(lr){6-7} \cmidrule(lr){8-9} \cmidrule(lr){10-11}
 & SRCC & PLCC & SRCC & PLCC & SRCC & PLCC & SRCC & PLCC & SRCC & PLCC \\
\midrule
\multicolumn{11}{c}{\textit{\textbf{Comparison with SOTA Open-Source \&  Closed-Source MLLMs}}} \\
Qwen-2.5-VL-72B-instruct~\cite{qwen2.5} & 0.408 & 0.387 & \textcolor{blue}{0.727} & \textcolor{blue}{0.763} & 0.232 & 0.235 & 0.626 & 0.589 & 0.233 & 0.197 \\
InternVL3-78B~\cite{internvl3} & 0.385 & 0.344 & 0.666 & 0.694 & 0.221 & 0.220 & 0.518 & 0.433 & 0.223 & 0.206 \\
GPT-4o~\cite{gpt4} & \textcolor{blue}{0.509} & \textcolor{blue}{0.485} & 0.697 & 0.744 & 0.278 & 0.282 & 0.605 & 0.597 & \textcolor{blue}{0.333} & \textcolor{blue}{0.276} \\
Gemini-2.0-flash~\cite{gemini} & 0.474 & 0.457 & 0.703 & 0.704 & \textcolor{blue}{0.319} & \textcolor{blue}{0.323} & \textcolor{blue}{0.658} & \textcolor{blue}{0.651} & 0.286 & 0.265 \\
\rowcolor{gray!30} \textbf{ArtiMuse (Ours)} & \textbf{\textcolor{red}{0.827}} & \textbf{\textcolor{red}{0.826}} & \textbf{\textcolor{red}{0.936}} & \textbf{\textcolor{red}{0.958}} & \textbf{\textcolor{red}{0.510}} & \textbf{\textcolor{red}{0.543}} & \textbf{\textcolor{red}{0.814}} & \textbf{\textcolor{red}{0.837}} & \textbf{\textcolor{red}{0.614}} & \textbf{\textcolor{red}{0.627}} \\

\bottomrule
\end{tabular}
}
% \vspace{-10pt}
\end{table}

\subsection{Further Comparison of Generalization Ability}
We further experimentally validate ArtiMuse's generalization ability through comprehensive cross-dataset evaluations. As shown in Tab.~\ref{tab:generation}, we train both the state-of-the-art open-source IAA model Q-Align~\cite{qalign} and ArtiMuse on AVA~\cite{ava}, PARA~\cite{para}, TAD66K~\cite{tad66k}, FLICKR-AES~\cite{flickr}, and ArtiMuse-10K, then evaluate them across all five datasets. The results demonstrate that ArtiMuse consistently outperforms Q-Align on unseen datasets in most cases, confirming its superior generalization capability.

\begin{table}[htbp]
% \vspace{-10pt}
\centering
\caption{Further comparison of generalization ability. The best performances are highlighted in \textcolor{red}{red}. * Results are trained only on single dataset to compare the generalization ability. ArtiMuse demonstrates strong generalization capabilities when compared to state-of-the-art IAA models.}
\label{tab:generation}
\resizebox{\textwidth}{!}{%
% \fontsize{8}{9}\selectfont
% \renewcommand{\arraystretch}{0.8} % 调整行距，值越小行距越窄
\begin{tabular}{lcccccccccc}
\toprule
\multirow{2}{*}{\textbf{Model}} & \multicolumn{2}{c}{\textbf{AVA}~\cite{ava}} & \multicolumn{2}{c}{\textbf{PARA}~\cite{para}} & \multicolumn{2}{c}{\textbf{TAD66K}~\cite{tad66k}} & \multicolumn{2}{c}{\textbf{FLICKR-AES}~\cite{flickr}} & \multicolumn{2}{c}{\textbf{ArtiMuse-10K}} \\
\cmidrule(lr){2-3} \cmidrule(lr){4-5} \cmidrule(lr){6-7} \cmidrule(lr){8-9} \cmidrule(lr){10-11}
 & SRCC & PLCC & SRCC & PLCC & SRCC & PLCC & SRCC & PLCC & SRCC & PLCC \\
\midrule
\multicolumn{11}{c}{\textit{\textbf{Further Comparison of Generalization Ability}}} \\

Q-Align (AVA) * & 0.822 & 0.817 & 0.694 & 0.711 & 0.417 & 0.445 & 0.643 & 0.664 & 0.337 & 0.320 \\
 \rowcolor{gray!30} \textbf{ArtiMuse (AVA) *} & \textbf{\textcolor{red}{0.827}} & \textbf{\textcolor{red}{0.826}} & \textbf{\textcolor{red}{0.697}} & \textbf{\textcolor{red}{0.725}} & \textbf{\textcolor{red}{0.419}} & \textbf{\textcolor{red}{0.451}} & \textbf{\textcolor{red}{0.647}} & \textbf{\textcolor{red}{0.676}} & \textbf{\textcolor{red}{0.395}} & \textbf{\textcolor{red}{0.376}} \\ \midrule

Q-Align (PARA) * & 0.492 & 0.456 & 0.913 & 0.888 & 0.300 & 0.281 & 0.913 & 0.888 & 0.158 & 0.115 \\
\rowcolor{gray!30} \textbf{ArtiMuse (PARA) *} & \textbf{\textcolor{red}{0.493}} & \textbf{\textcolor{red}{0.510}} & \textbf{\textcolor{red}{0.936}} & \textbf{\textcolor{red}{0.958}} & \textbf{\textcolor{red}{0.301}} & \textbf{\textcolor{red}{0.311}} & \textbf{\textcolor{red}{0.936}} & \textbf{\textcolor{red}{0.958}} & \textbf{\textcolor{red}{0.229}} & \textbf{\textcolor{red}{0.188}} \\ \midrule

Q-Align (TAD66K) * & \textcolor{red}{0.695} & \textcolor{red}{0.699} & 0.688 & 0.667 & 0.501 & 0.531 & 0.688 & 0.667 & 0.317 & 0.304 \\
\rowcolor{gray!30} \textbf{ArtiMuse (TAD66K) *} & \textbf{0.671} & \textbf{0.676} & \textbf{\textcolor{red}{0.719}} & \textbf{\textcolor{red}{0.677}} & \textbf{\textcolor{red}{0.510}} & \textbf{\textcolor{red}{0.543}} & \textbf{\textcolor{red}{0.719}} & \textbf{\textcolor{red}{0.677}} & \textbf{\textcolor{red}{0.397}} & \textbf{\textcolor{red}{0.369}} \\ \midrule

Q-Align (FLICKR-AES) * & \textcolor{red}{0.609} & \textcolor{red}{0.611} & 0.836 & 0.839 & 0.366 & 0.376 & 0.798 & 0.818 & 0.215 & 0.208 \\
\rowcolor{gray!30} \textbf{ArtiMuse (FLICKR-AES) *} & \textbf{0.581} & \textbf{0.594} & \textbf{\textcolor{red}{0.854}} & \textbf{\textcolor{red}{0.874}} & \textbf{\textcolor{red}{0.379}} & \textbf{\textcolor{red}{0.397}} & \textbf{\textcolor{red}{0.814}} & \textbf{\textcolor{red}{0.837}} & \textbf{\textcolor{red}{0.294}} & \textbf{\textcolor{red}{0.285}} \\ \midrule

Q-Align (ArtiMuse-10K) * & \textcolor{red}{0.398} & \textcolor{red}{0.386} & 0.346 & 0.395 & 0.194 & 0.197 & 0.137 & 0.123 & 0.551 & 0.573 \\
 \rowcolor{gray!30} \textbf{ArtiMuse (ArtiMuse-10K) *} & \textbf{0.397} & \textbf{0.385} & \textbf{\textcolor{red}{0.446}} & \textbf{\textcolor{red}{0.461}} & \textbf{\textcolor{red}{0.230}} & \textbf{\textcolor{red}{0.232}} & \textbf{\textcolor{red}{0.349}} & \textbf{\textcolor{red}{0.334}} & \textbf{\textcolor{red}{0.614}} & \textbf{\textcolor{red}{0.627}} \\ 

\bottomrule
\end{tabular}
}
% \vspace{-10pt}
\end{table}

% 我们进一步通过实验验证了ArtiMuse的Generalization Ability，如表格所示。我们将当前最好的开源IAA模型Q-Align和ArtiMuse分别在AVA~\cite{ava}, PARA~\cite{para}, TAD66K~\cite{tad66k},  FLICKR-AES~\cite{flickr} and ArtiMuse-10K上进行训练，然后在所有五个数据集上进行测试，结果表明绝大多数情况下ArtiMuse对于unseen的dataset都有着更好的表现，证明了其更优秀的泛化能力。
% \subsection{Further Comparison of Generalization Ability}
% We further experimentally validate ArtiMuse's generalization ability through comprehensive cross-dataset evaluations. As shown in Tab.~\ref{tab:generation}, we train both the state-of-the-art open-source IAA model Q-Align~\cite{qalign} and ArtiMuse on AVA~\cite{ava}, PARA~\cite{para}, TAD66K~\cite{tad66k}, FLICKR-AES~\cite{flickr}, and ArtiMuse-10K, then evaluate them across all five datasets. The results demonstrate that ArtiMuse consistently outperforms Q-Align on unseen datasets in most cases, confirming its superior generalization capability.

\subsection{Image Examples in ArtiMuse-10K} 

\begin{figure}[htbp]
  \centering
    \includegraphics[width=0.99\linewidth]{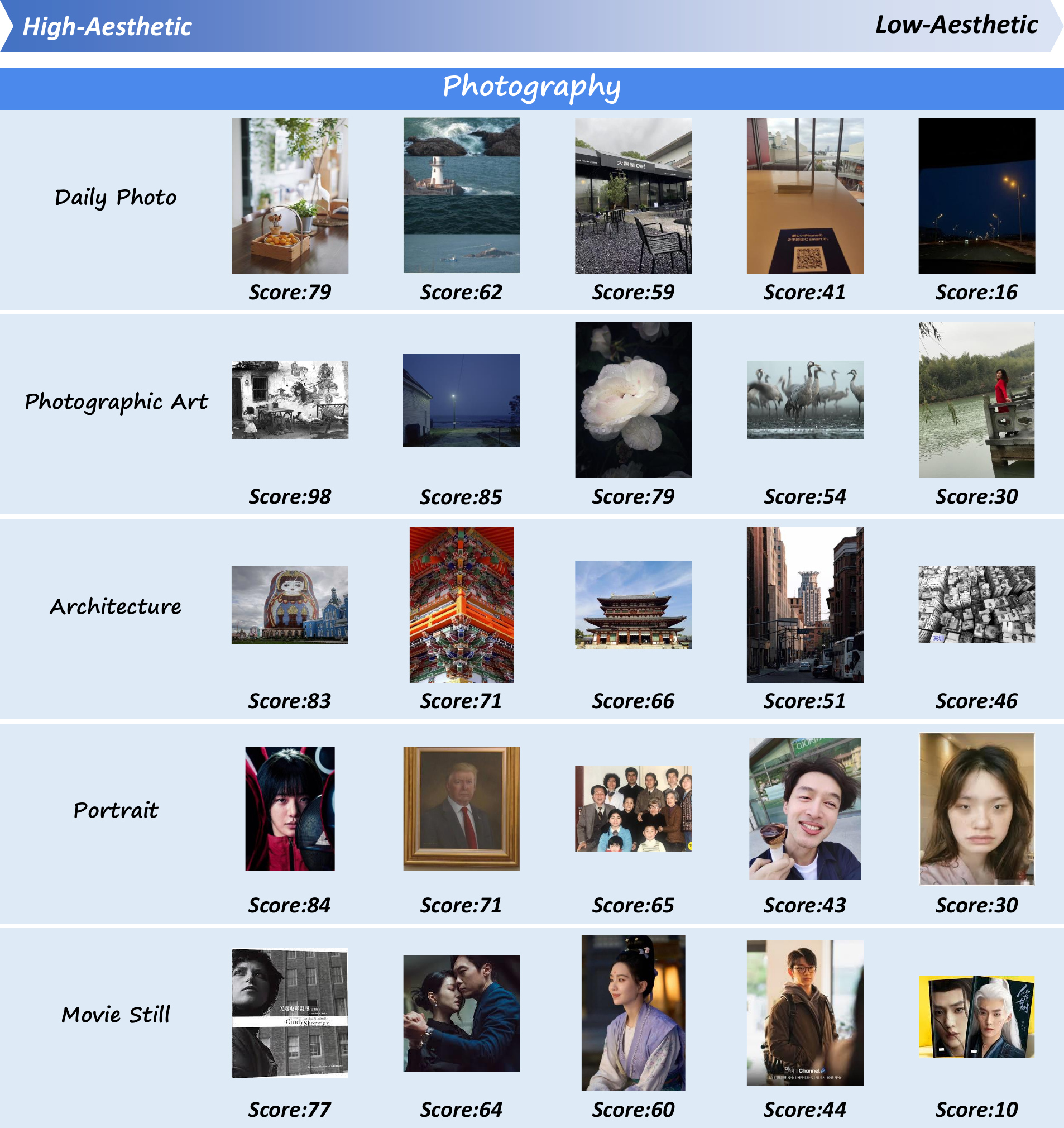}
   % \vspace{-10pt}
   \caption{Image examples from the \textit{Photography} category in ArtiMuse-10K dataset.}
   \label{fig:dataset_example_photo}
   \vspace{20pt}
\end{figure}

\begin{figure}[htbp]
  \centering
    \includegraphics[width=0.99\linewidth]{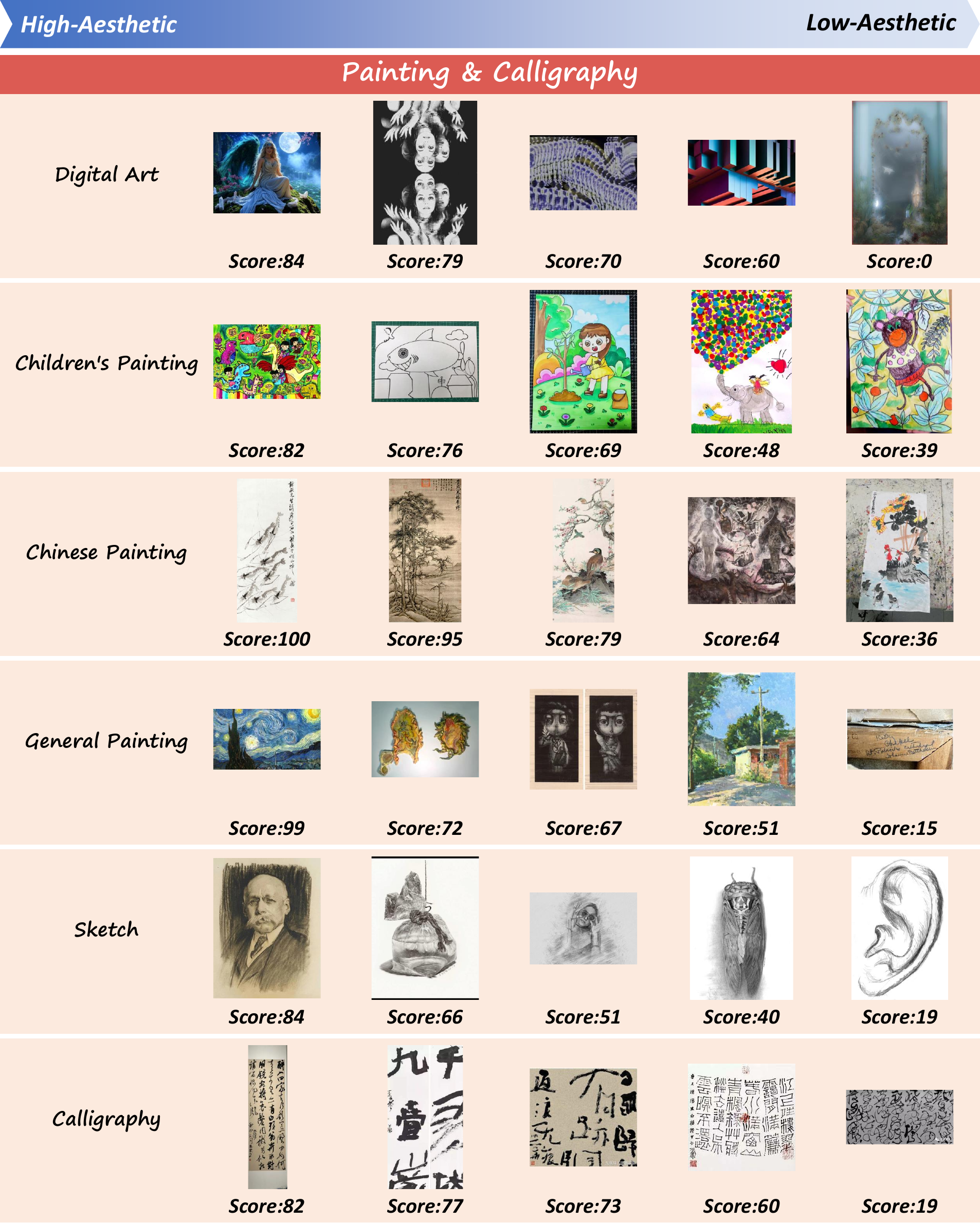}
   % \vspace{-10pt}
   \caption{Image examples from the \textit{Painting \& Calligraphy} category in ArtiMuse-10K dataset.}
   \label{fig:dataset_example_painting}
   % \vspace{-5pt}
\end{figure}

\begin{figure}[htbp]
  \centering
    \includegraphics[width=0.95\linewidth]{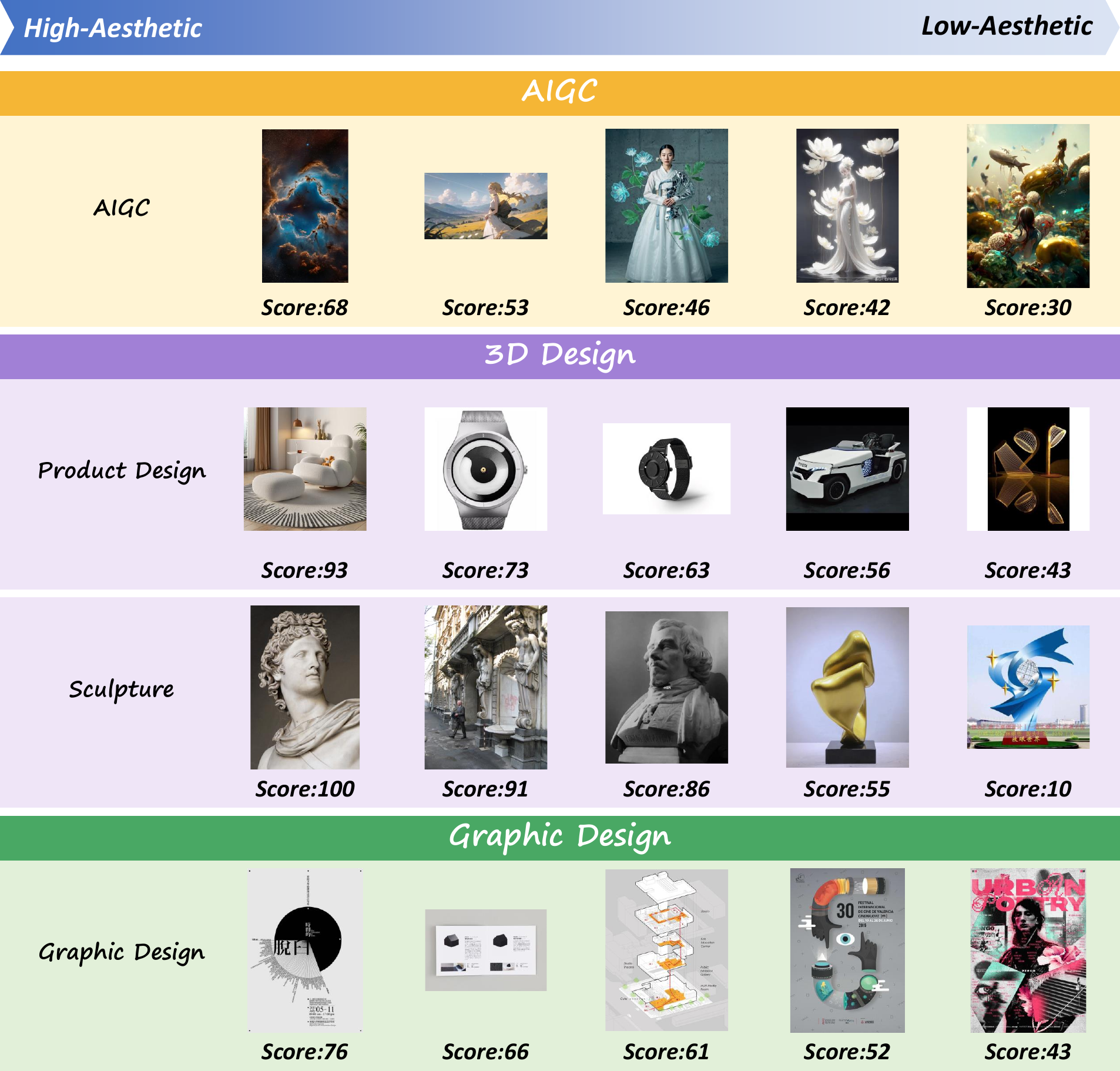}
   \vspace{-10pt}
   \caption{Image examples from the \textit{AIGC}, \textit{3D Design} and \textit{Graphic Design categories} in ArtiMuse-10K dataset.}
   \label{fig:dataset_example_others}
   \vspace{-5pt}
\end{figure}

% ArtiMuse-10K的详细图像示例，按照所有subcategories进行排列，展示如Fig.1
As illustrated in Fig.~\ref{fig:dataset_example_photo}, Fig.~\ref{fig:dataset_example_painting} and Fig.~\ref{fig:dataset_example_others}, the ArtiMuse-10K dataset includes a diverse collection of images, meticulously organized across all specified subcategories. The dataset encompasses a wide range of aesthetic qualities and sources, ensuring rich variability and broad representativeness.

\subsection{Complete Examples in ArtiMuse-10K}
\begin{figure}[htbp]
  \centering
    % \includegraphics[width=0.99\linewidth]{figures/supp/ex_photo_high.pdf}
    % \textbf{\bigskip} % Adds 1cm of vertical space
    \includegraphics[width=0.99\linewidth]{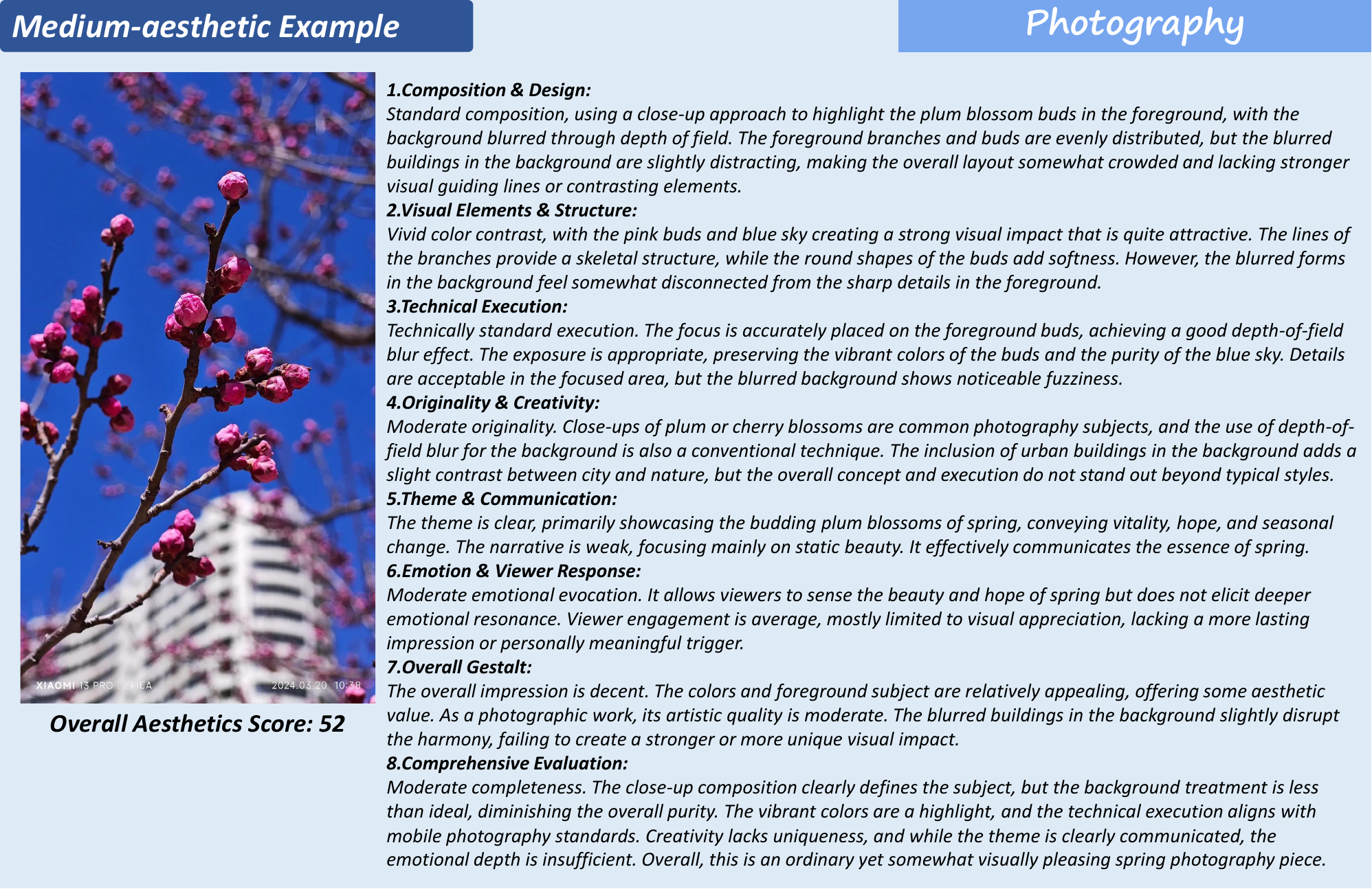}
    \includegraphics[width=0.99\linewidth]{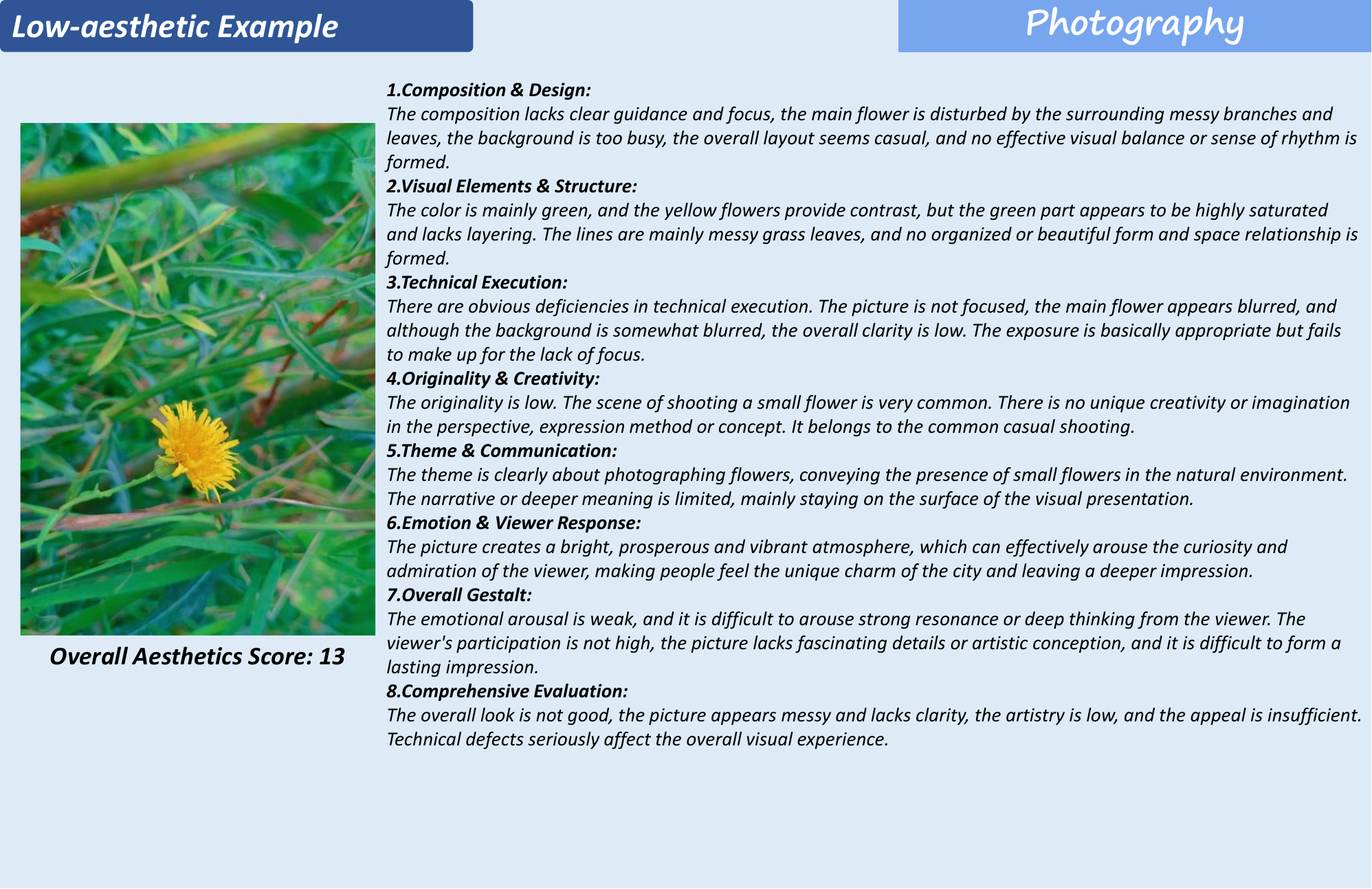}
   \caption{Medium-aesthetic example and low-aesthetic example from \textit{Photography} category.}
   \label{fig:full_ex1}
   \vspace{5cm}
\end{figure}

\begin{figure}[htbp]
  \centering
    % \includegraphics[width=0.99\linewidth]{figures/supp/ex_photo_low.pdf}
    % \textbf{\bigskip} % Adds 1cm of vertical space
    \includegraphics[width=0.99\linewidth]{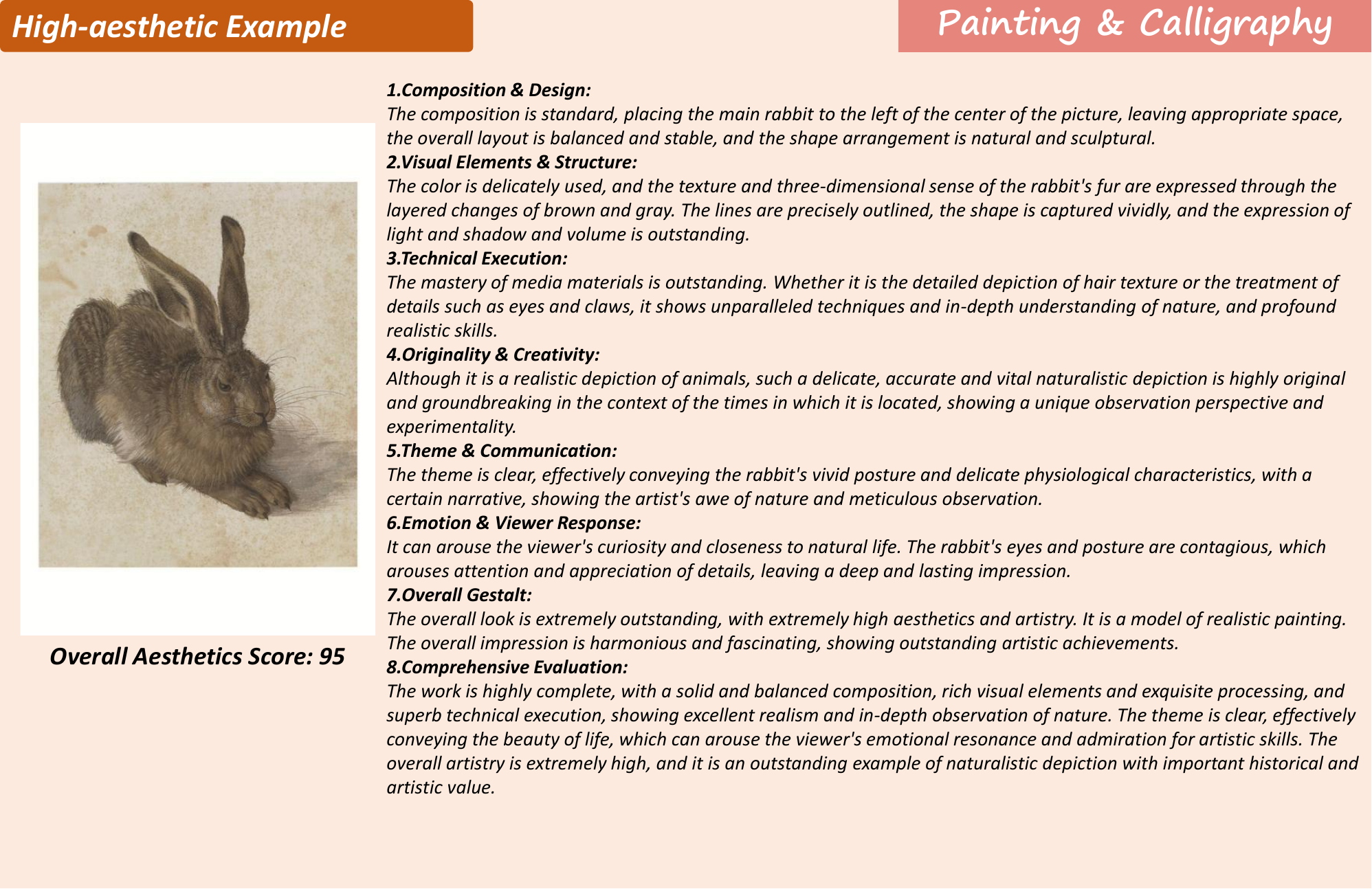}
    \includegraphics[width=0.99\linewidth]{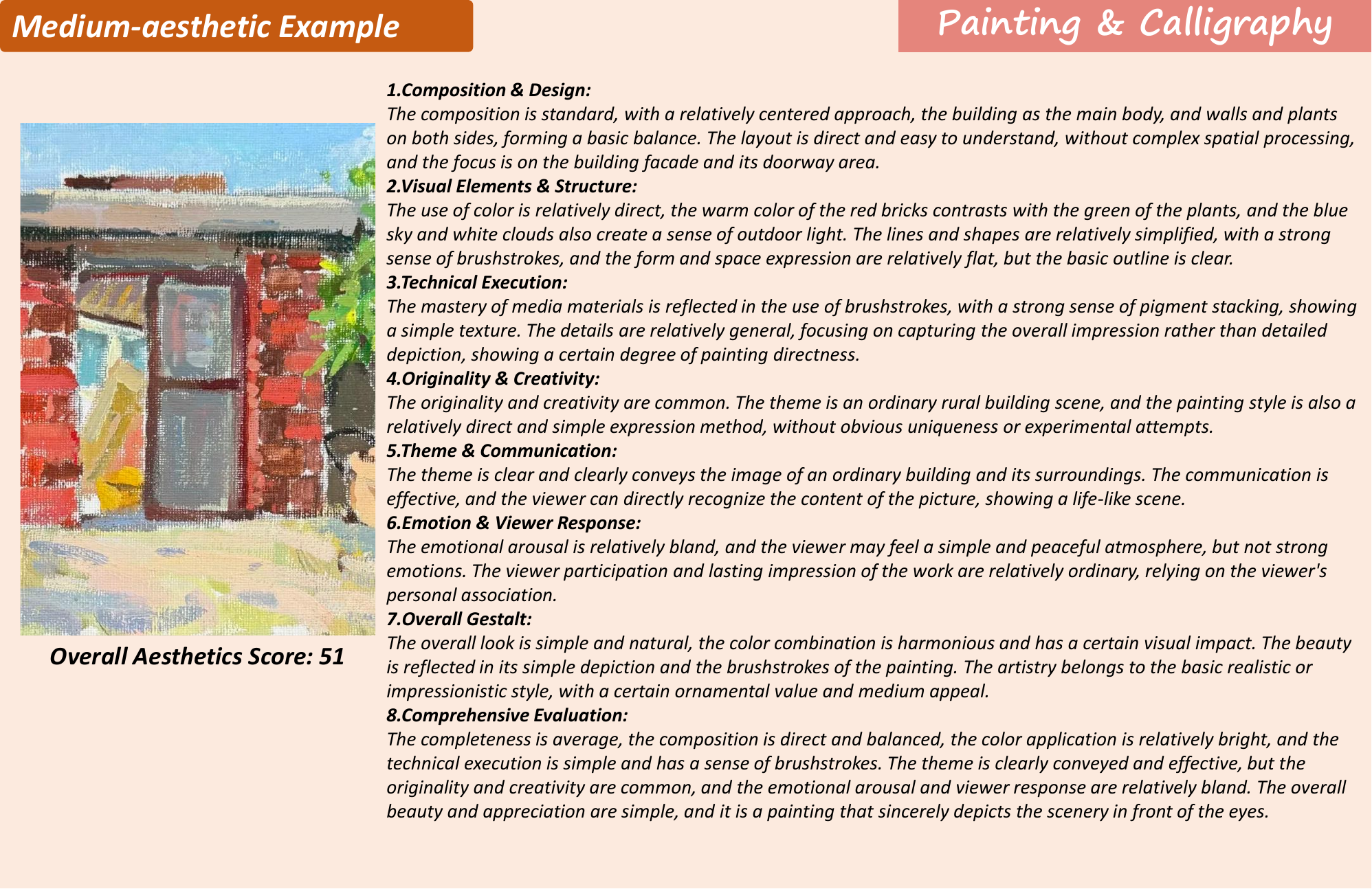}
   \caption{High-aesthetic and medium-aesthetic example from \textit{Painting \& Calligraphy} category.}
   \label{fig:full_ex2}
   \vspace{5cm}
\end{figure}

\begin{figure}[htbp]
  \centering
    % \includegraphics[width=0.99\linewidth]{figures/supp/ex_painting_medium.pdf}
    % \textbf{\bigskip} % Adds 1cm of vertical space
    \includegraphics[width=0.99\linewidth]{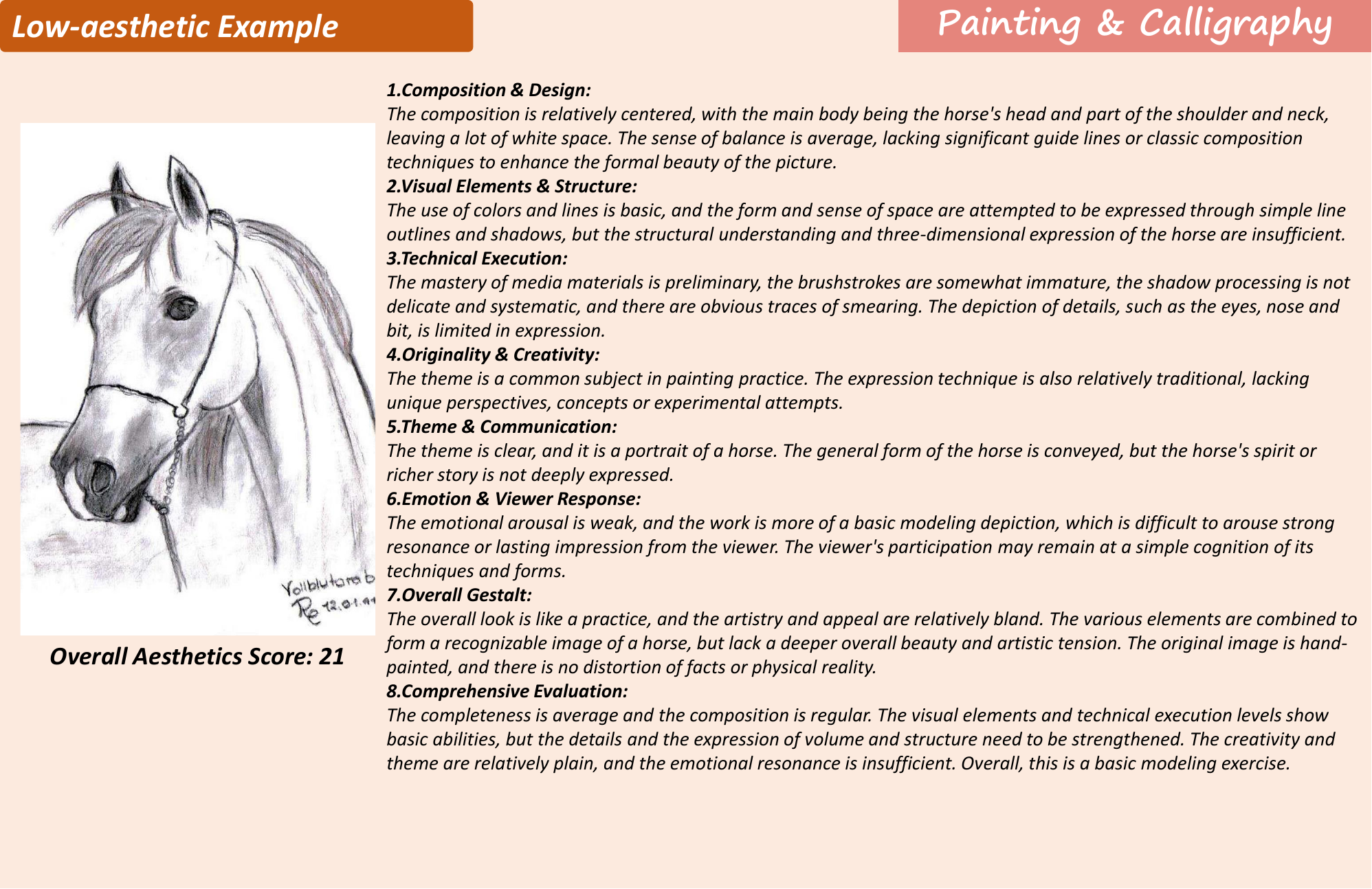}
    \includegraphics[width=0.99\linewidth]{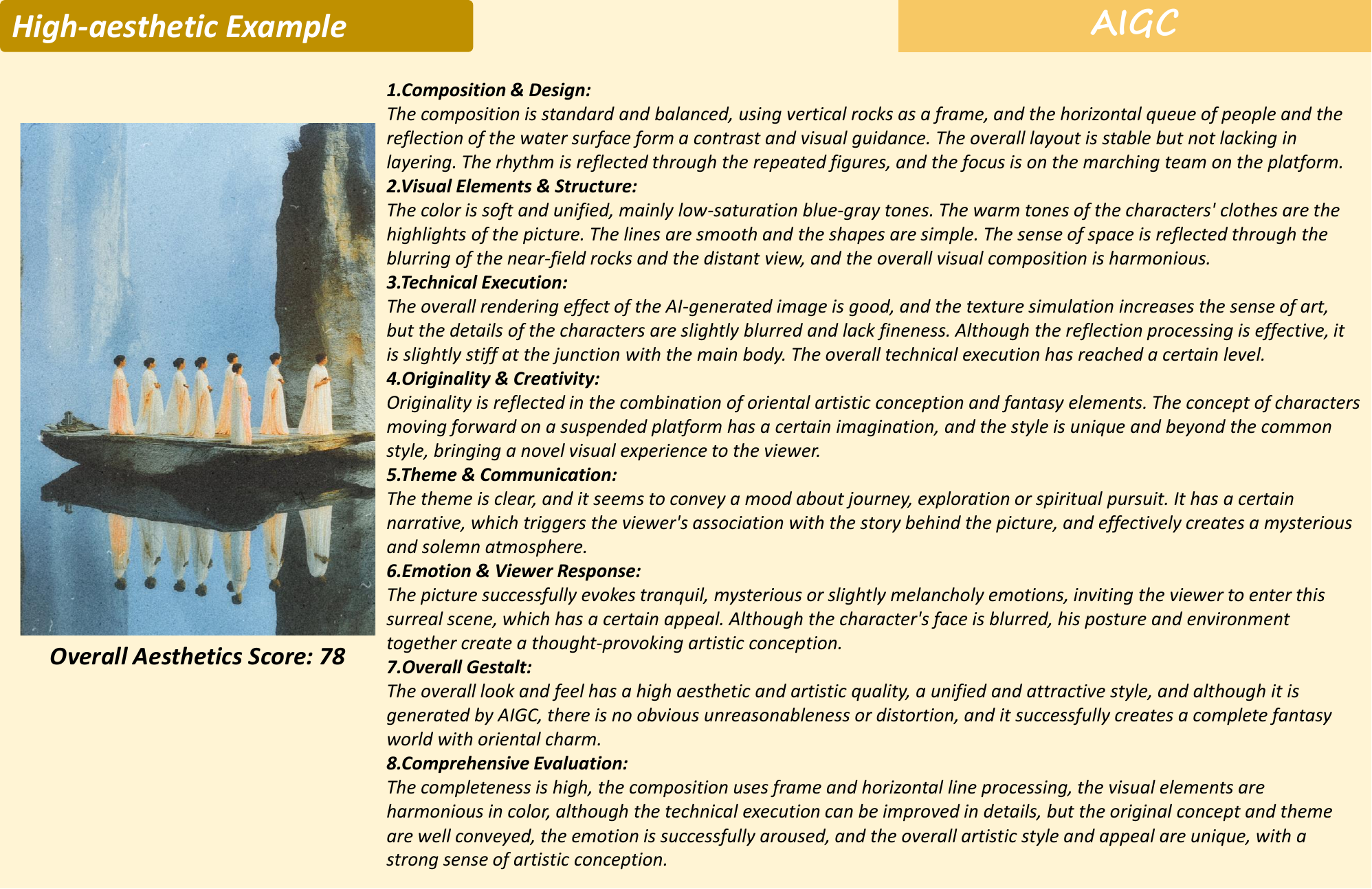}
   \caption{Low-aesthetic example from \textit{Painting \& Calligraphy} category and high-aesthetic example from \textit{AIGC} category.}
   \label{fig:full_ex3}
   \vspace{5cm}
\end{figure}

\begin{figure}[htbp]
  \centering
    % \includegraphics[width=0.99\linewidth]{figures/supp/ex_aigc_high.pdf}
    % \textbf{\bigskip} % Adds 1cm of vertical space
    \includegraphics[width=0.99\linewidth]{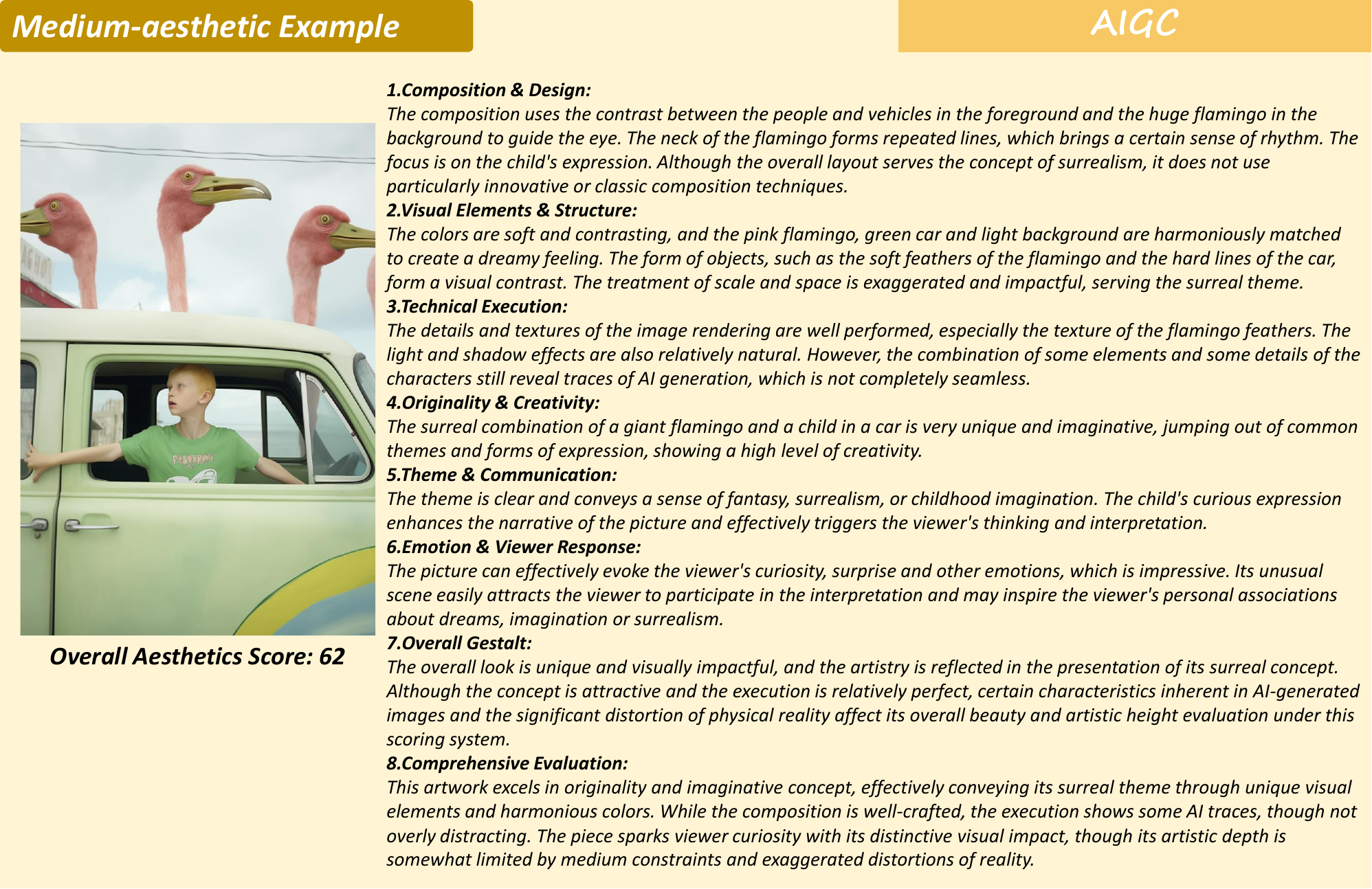}
    \includegraphics[width=0.99\linewidth]{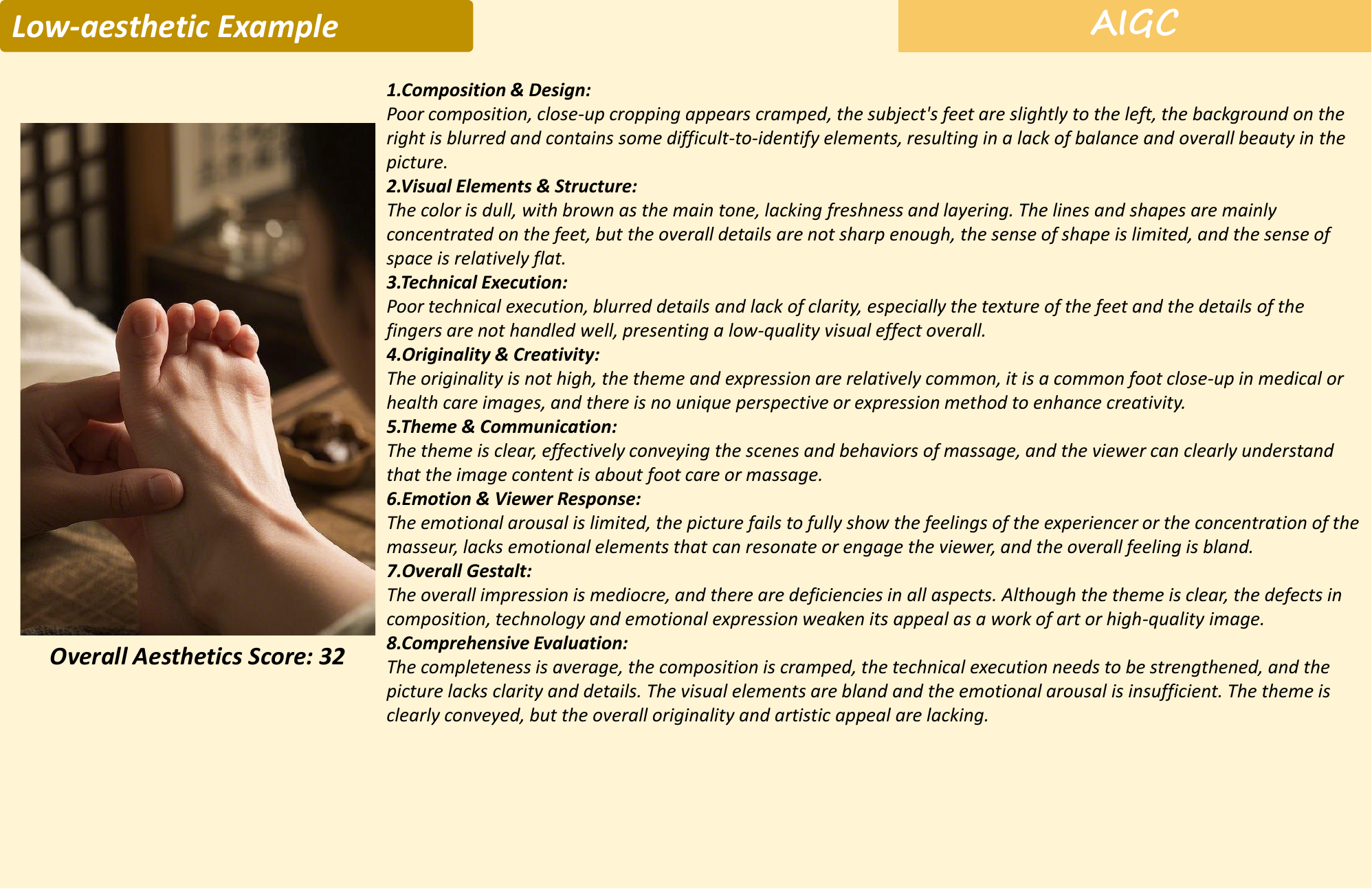}
   \caption{Medium-aesthetic example and low-aesthetic example from \textit{AIGC} category.}
   \label{fig:full_ex4}
   \vspace{5cm}
\end{figure}

\begin{figure}[htbp]
  \centering
    % \includegraphics[width=0.99\linewidth]{figures/supp/ex_aigc_low.pdf}
    % \textbf{\bigskip} % Adds 1cm of vertical space
    \includegraphics[width=0.99\linewidth]{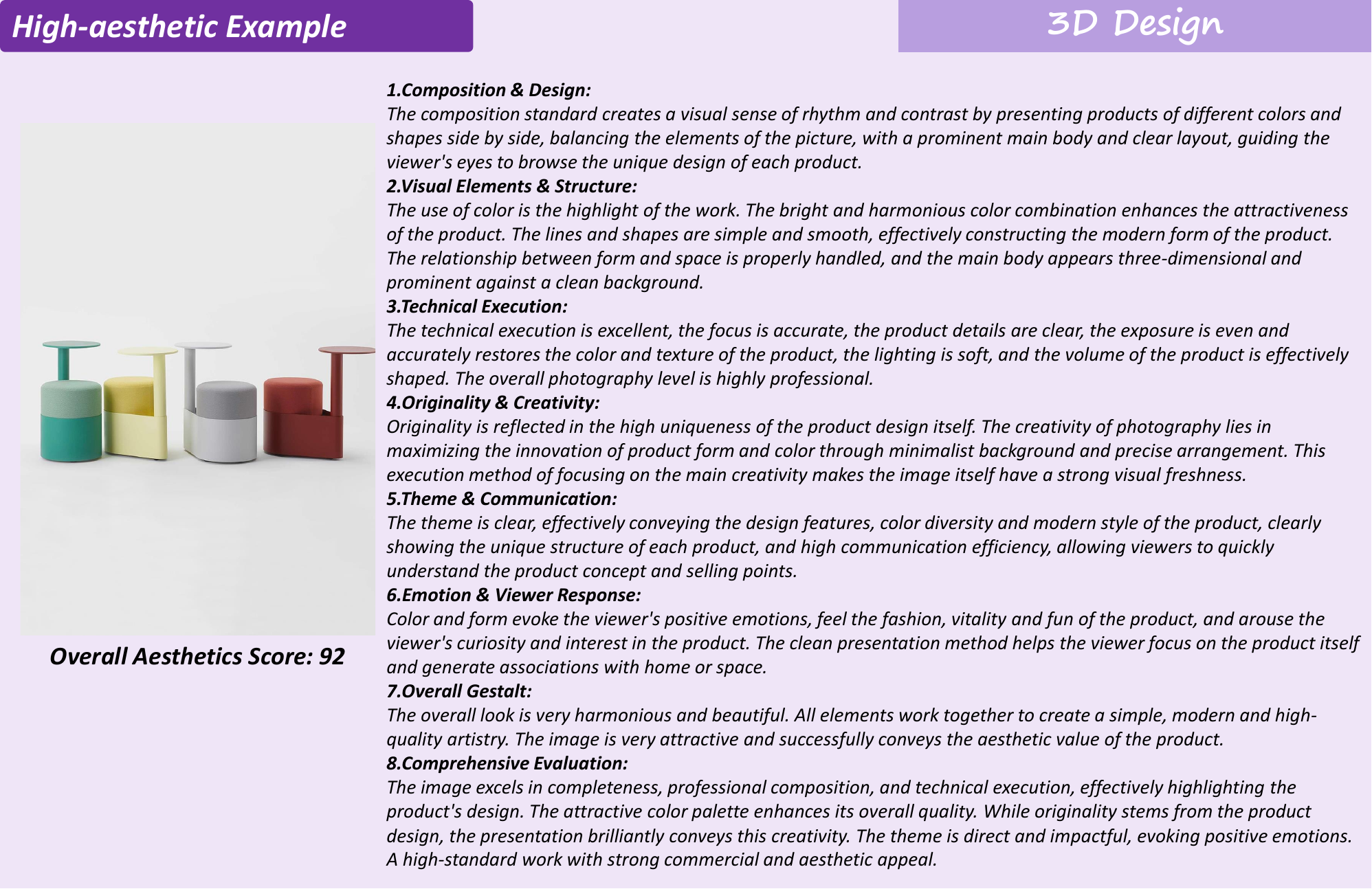}
    \includegraphics[width=0.99\linewidth]{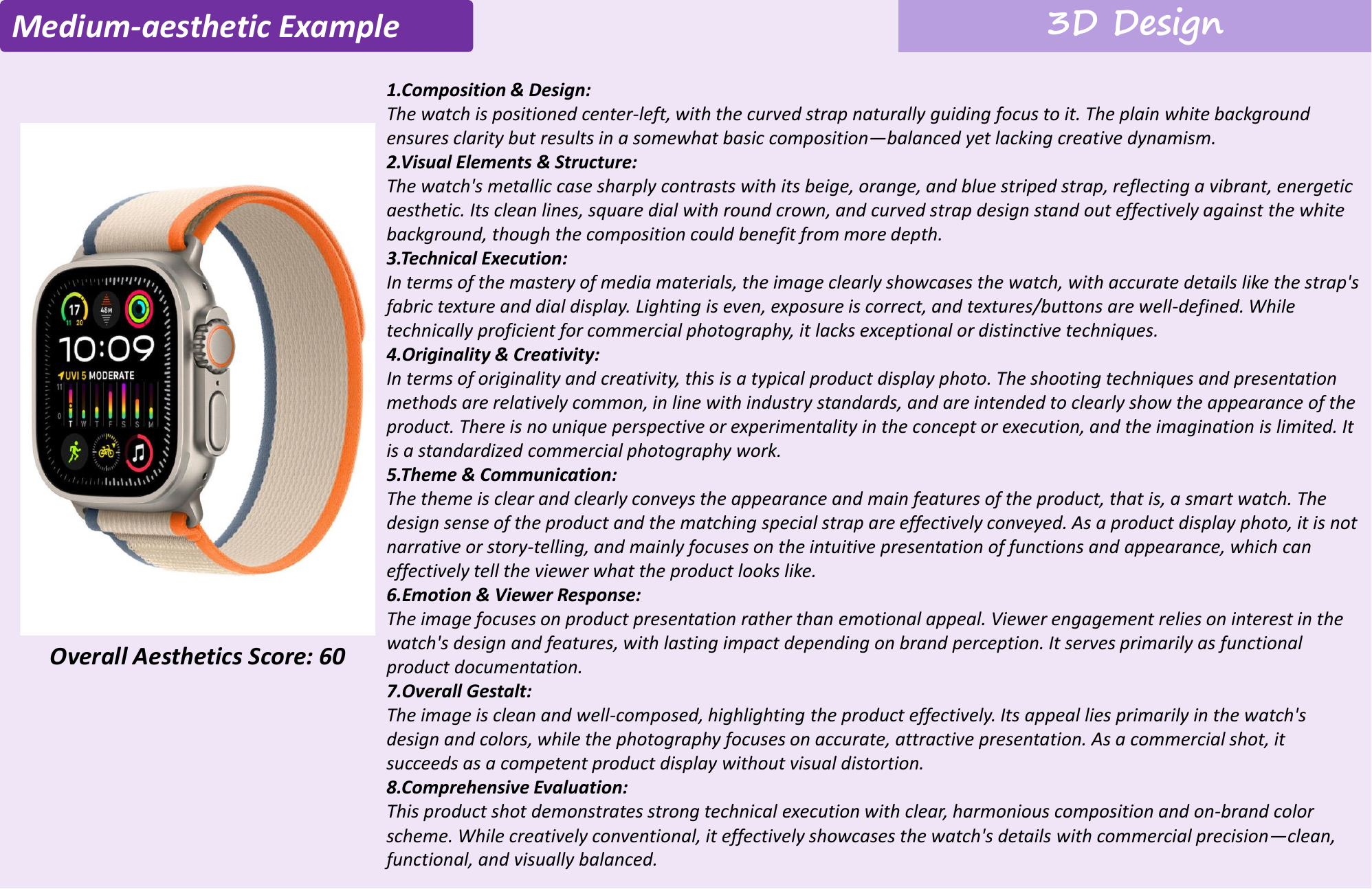}
   \caption{High-aesthetic and medium-aesthetic example from \textit{3D Design} category.}
   \label{fig:full_ex5}
   \vspace{5cm}
\end{figure}

\begin{figure}[htbp]
  \centering
    % \includegraphics[width=0.99\linewidth]{figures/supp/ex_3d_medium.pdf}
    % \textbf{\bigskip} % Adds 1cm of vertical space
    \includegraphics[width=0.99\linewidth]{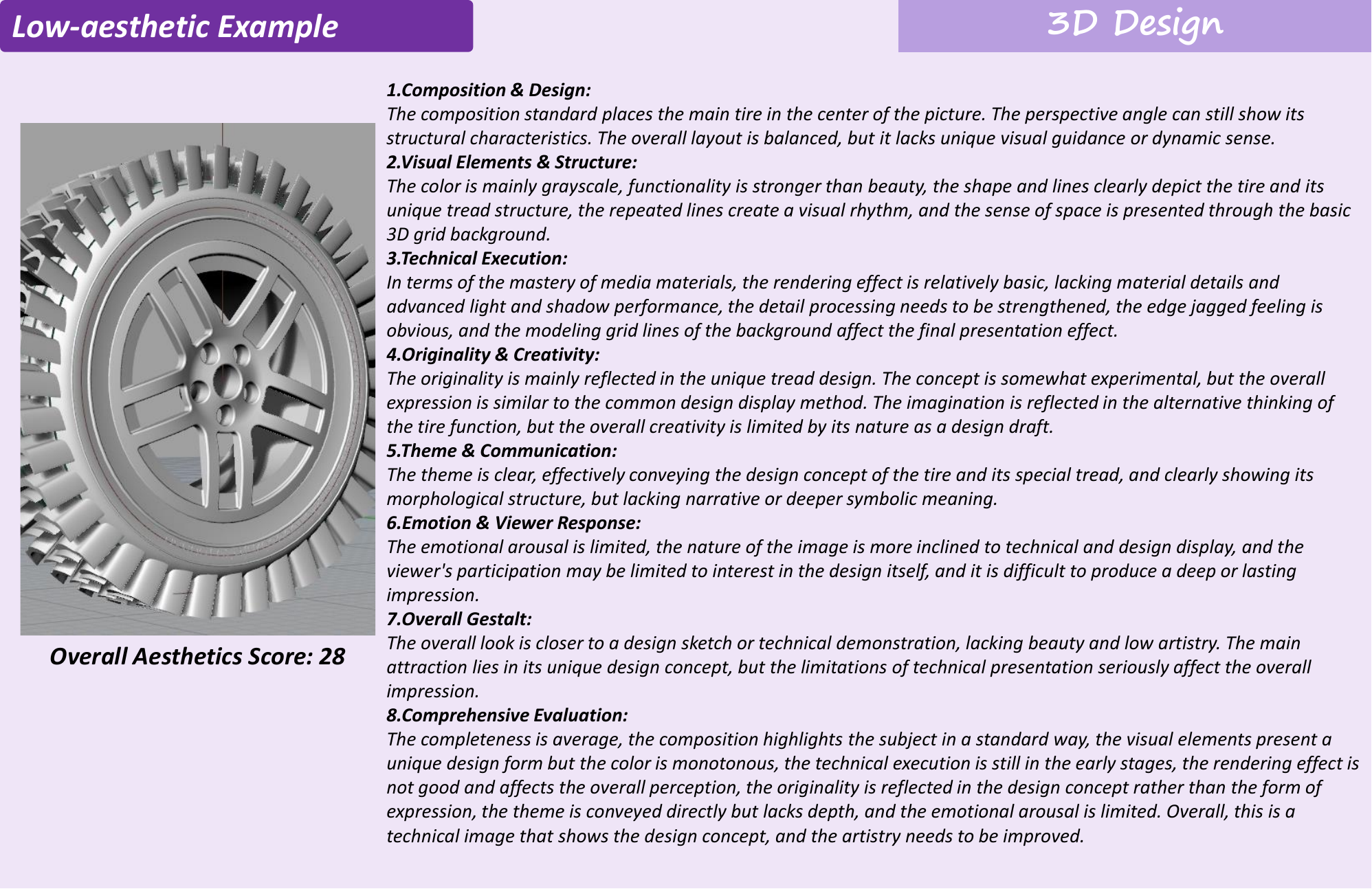}
    \includegraphics[width=0.99\linewidth]{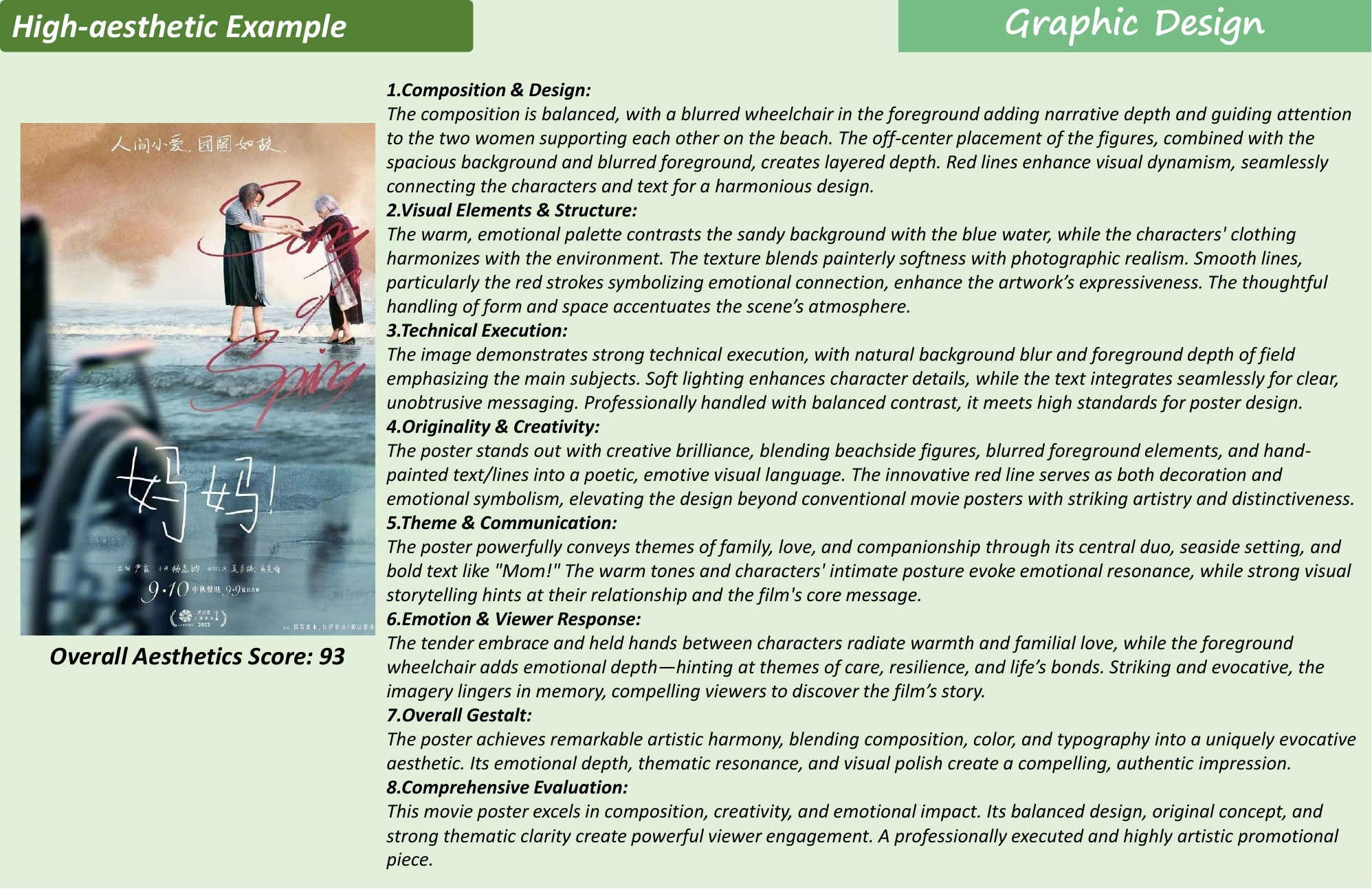}
   \caption{Low-aesthetic example from \textit{3D Design} category and high-aesthetic example from \textit{Graphic Design} category.}
   \label{fig:full_ex6}
   \vspace{5cm}
\end{figure}

\begin{figure}[htbp]
  \centering
    % \includegraphics[width=0.99\linewidth]{figures/supp/ex_graohic_high.pdf}
    % \textbf{\bigskip} % Adds 1cm of vertical space
    \includegraphics[width=0.99\linewidth]{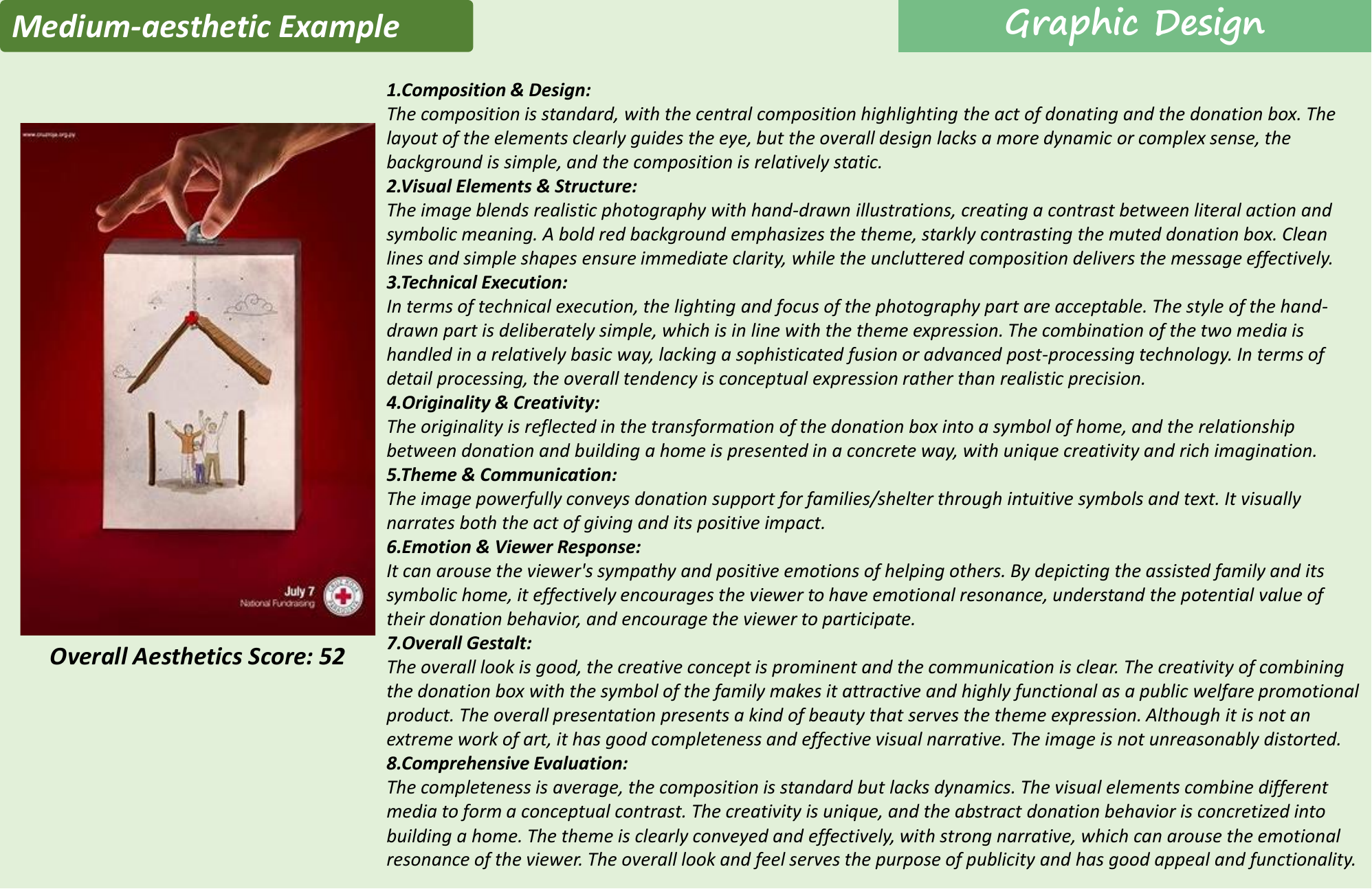}
    \includegraphics[width=0.99\linewidth]{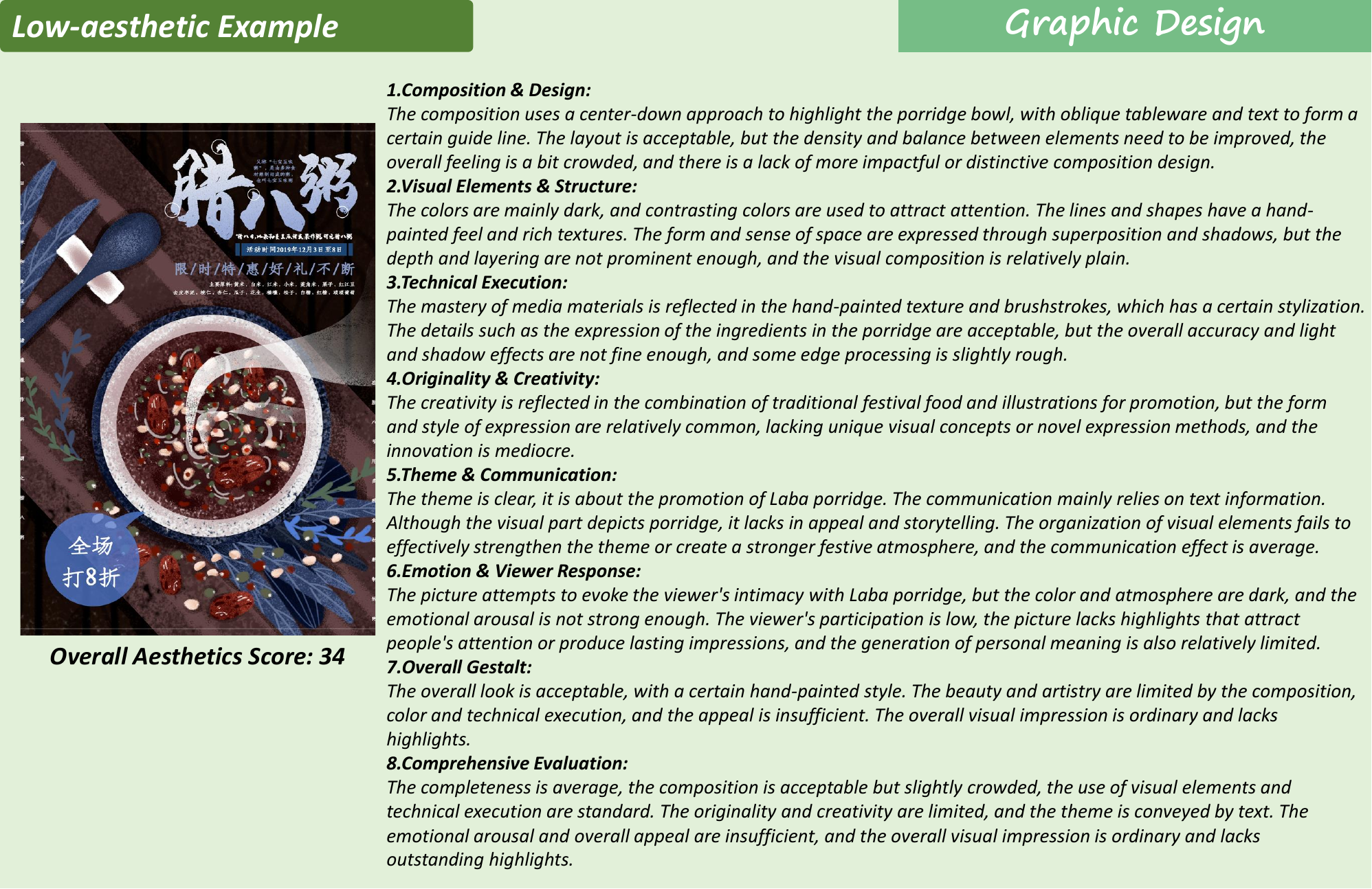}
   \caption{Medium-aesthetic example and low-aesthetic example from \textit{Graphic Design} category.}
   \label{fig:full_ex7}
   \vspace{5cm}
\end{figure}

In ArtiMuse-10K, professional experts meticulously evaluate each image across eight aesthetic attributes, providing detailed textual assessments along with an overall aesthetics score. Here, we present the complete data examples form each main categories in the dataset, including Photography, Painting \& Calligraphy, AIGC, 3D Design and Graphic Design, as shown in Fig.~\ref{fig:full_ex1},  Fig.~\ref{fig:full_ex2},  Fig.~\ref{fig:full_ex3},  Fig.~\ref{fig:full_ex4},  Fig.~\ref{fig:full_ex5},  Fig.~\ref{fig:full_ex6}, and Fig.~\ref{fig:full_ex7}.

% 我们选用real-world images来测试ArtiMuse对于训练数据集分布之外的图像的处理能力，结果展示在Fig.1中，可以看到处理真实世界图像时我们的模型仍然能够进行准确、专家级别的分析，结合图中的细节从美学专业角度指出相应的优点和缺点

\subsection{Further Comparison of Textual Analysis} 
\begin{figure}[htbp]
  \centering
    \includegraphics[width=0.99\linewidth]{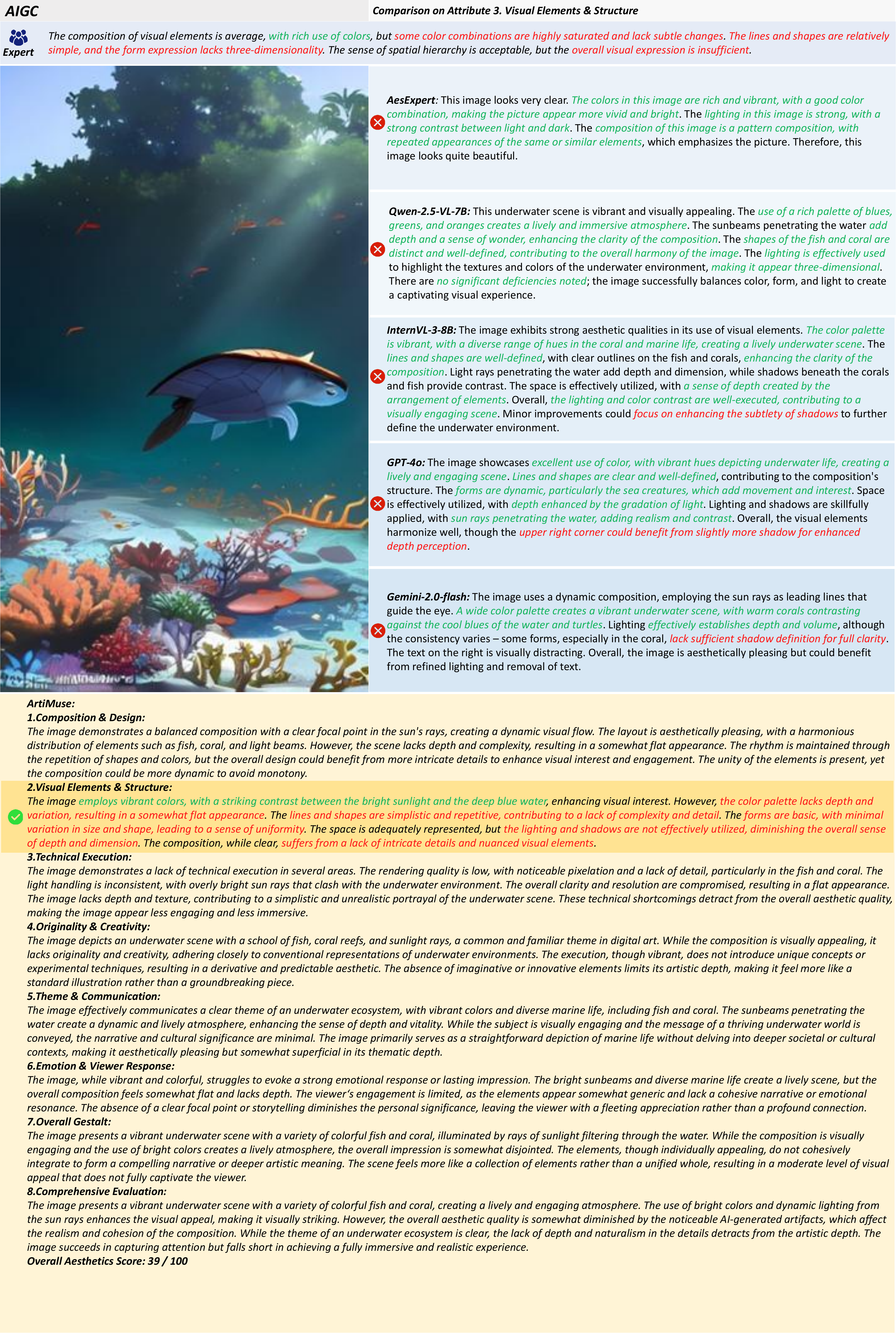}
    % \includegraphics[width=0.99\linewidth]{figures/supp/ArtiMuse_ex1.pdf}
   % \vspace{-10pt}
   \caption{Structural aesthetic analysis results and comparisons on AIGC images from ArtiMuse-10K test set. \textcolor{red}{Red} and \textcolor{green}{green} denote positive and negative analyses, respectively. ArtiMuse demonstrates superior performance in identifying flaws in low-aesthetic AIGC images, achieving higher accuracy than competing models and closely aligning with expert human evaluations.}
   \label{fig:text_aigc_comp}
   % \vspace{-5pt}
\end{figure}

\begin{figure}[htbp]
  \centering
    \includegraphics[width=0.99\linewidth]{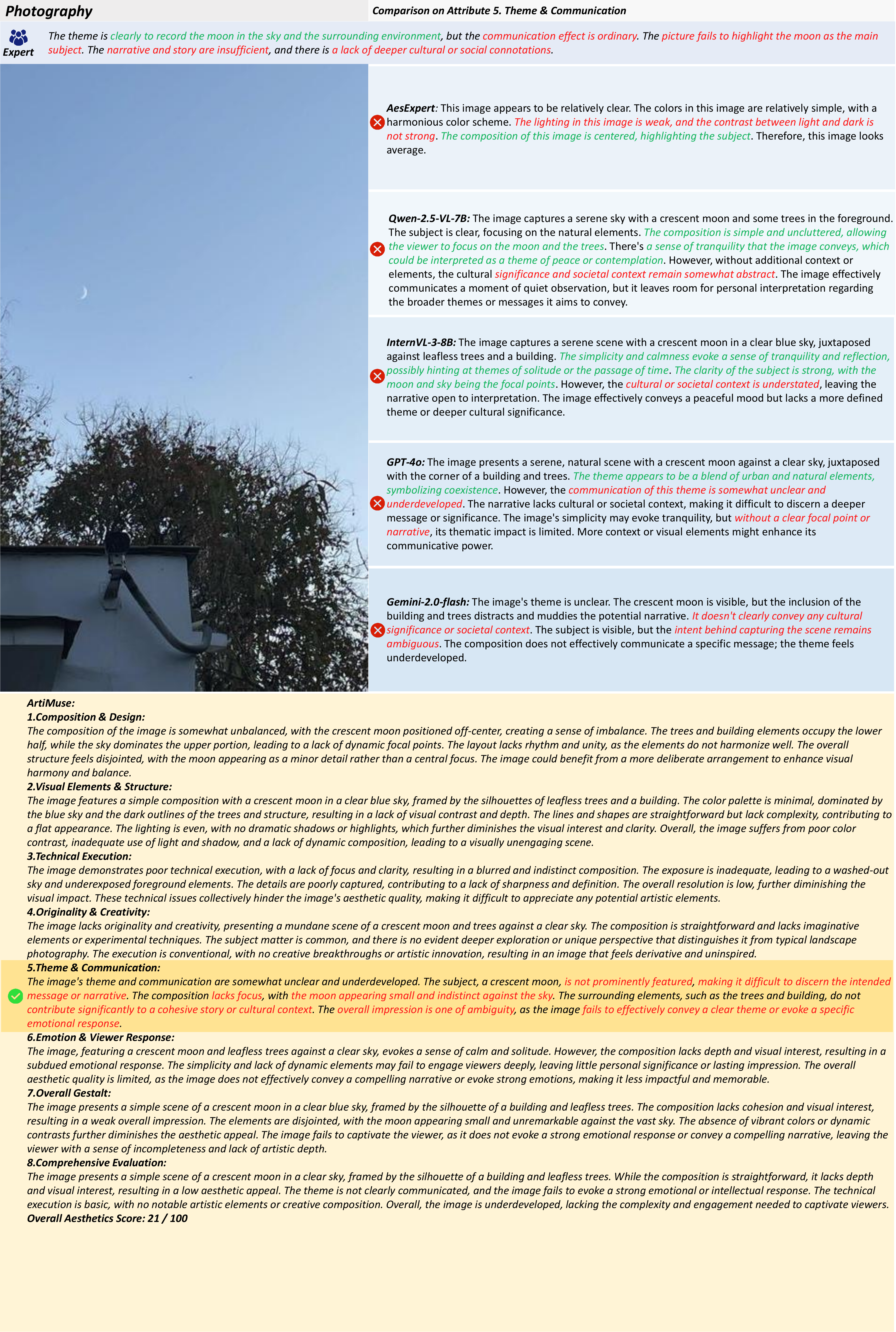}
    % \includegraphics[width=0.99\linewidth]{figures/supp/ArtiMuse_ex1.pdf}
   % \vspace{-10pt}
   \caption{Structural aesthetic analysis results and comparisons on photographic  images from ArtiMuse-10K test set. \textcolor{red}{Red} and \textcolor{green}{green} denote positive and negative analyses, respectively. ArtiMuse demonstrates superior performance in identifying flaws in low-aesthetic photographic images, achieving higher accuracy than competing models and closely aligning with expert human evaluations.}
   \label{fig:text_photo_comp}
   % \vspace{-5pt}
\end{figure}

\begin{figure}[htbp]
  \centering
    \includegraphics[width=0.99\linewidth]{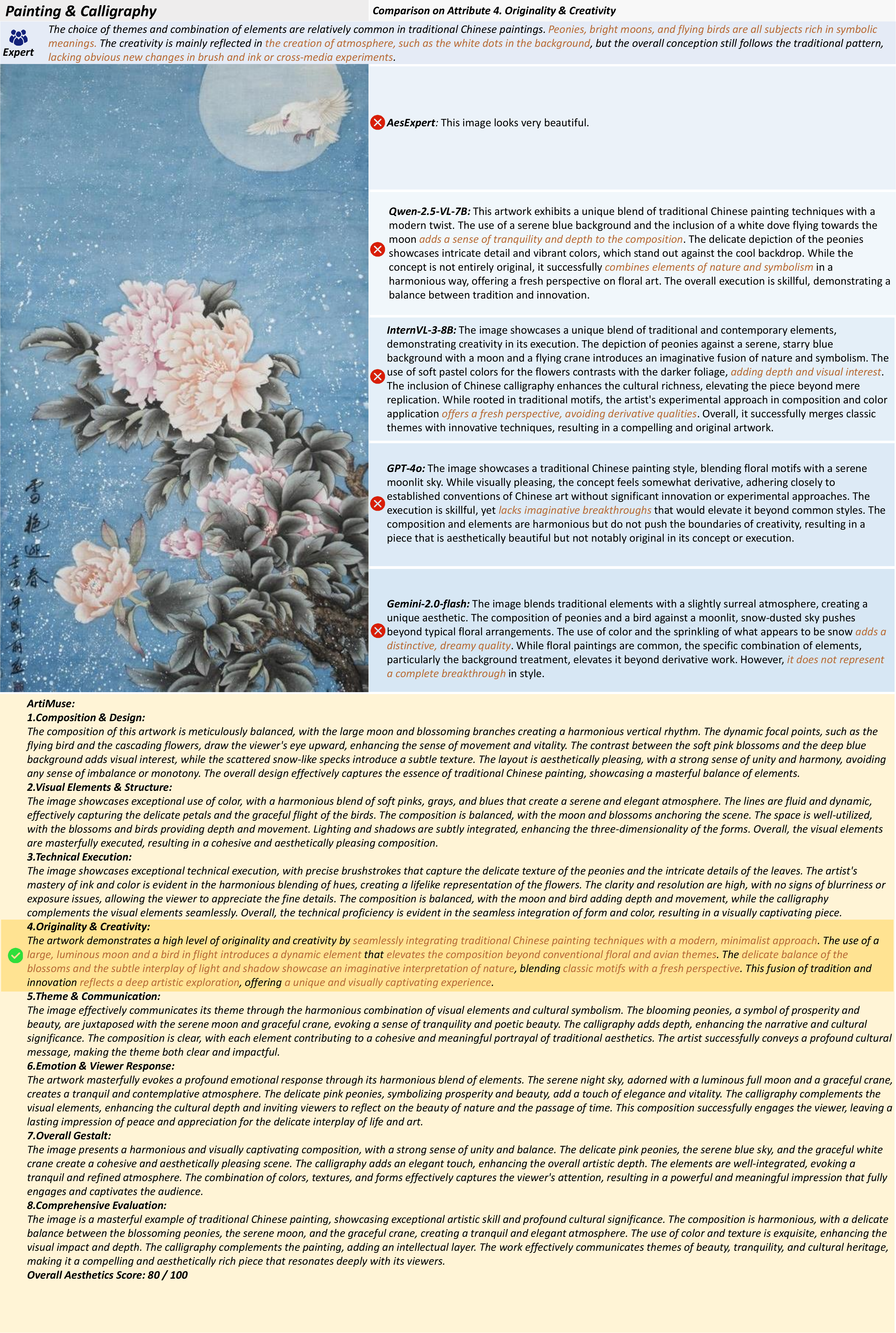}
    % \includegraphics[width=0.99\linewidth]{figures/supp/ArtiMuse_ex1.pdf}
   % \vspace{-10pt}
   \caption{Structural aesthetic analysis results and comparisons on paintings from ArtiMuse-10K test set. \textcolor{brown}{Brown} denote expert-level analyses. ArtiMuse demonstrates superior technical analysis capabilities for high-aesthetic images, exhibiting higher accuracy than competing models and achieving near-expert-level performance in aesthetic evaluation.}
   \label{fig:text_painting_comp}
   % \vspace{-5pt}
\end{figure}

We provide comprehensive examples of ArtiMuse's structural aesthetic analysis on images, accompanied by expert commentary and comparative evaluations with other models, as illustrated in Fig.~\ref{fig:text_aigc_comp}, Fig.~\ref{fig:text_photo_comp}, and Fig.~\ref{fig:text_painting_comp}. All images used in this analysis are sourced from the ArtiMuse-10K test set.
\label{sec:further_text}

\subsection{Results on Real-world Images} 
\begin{figure}[htbp]
  \centering
  \vspace{-10pt}
    \includegraphics[width=0.93\linewidth]{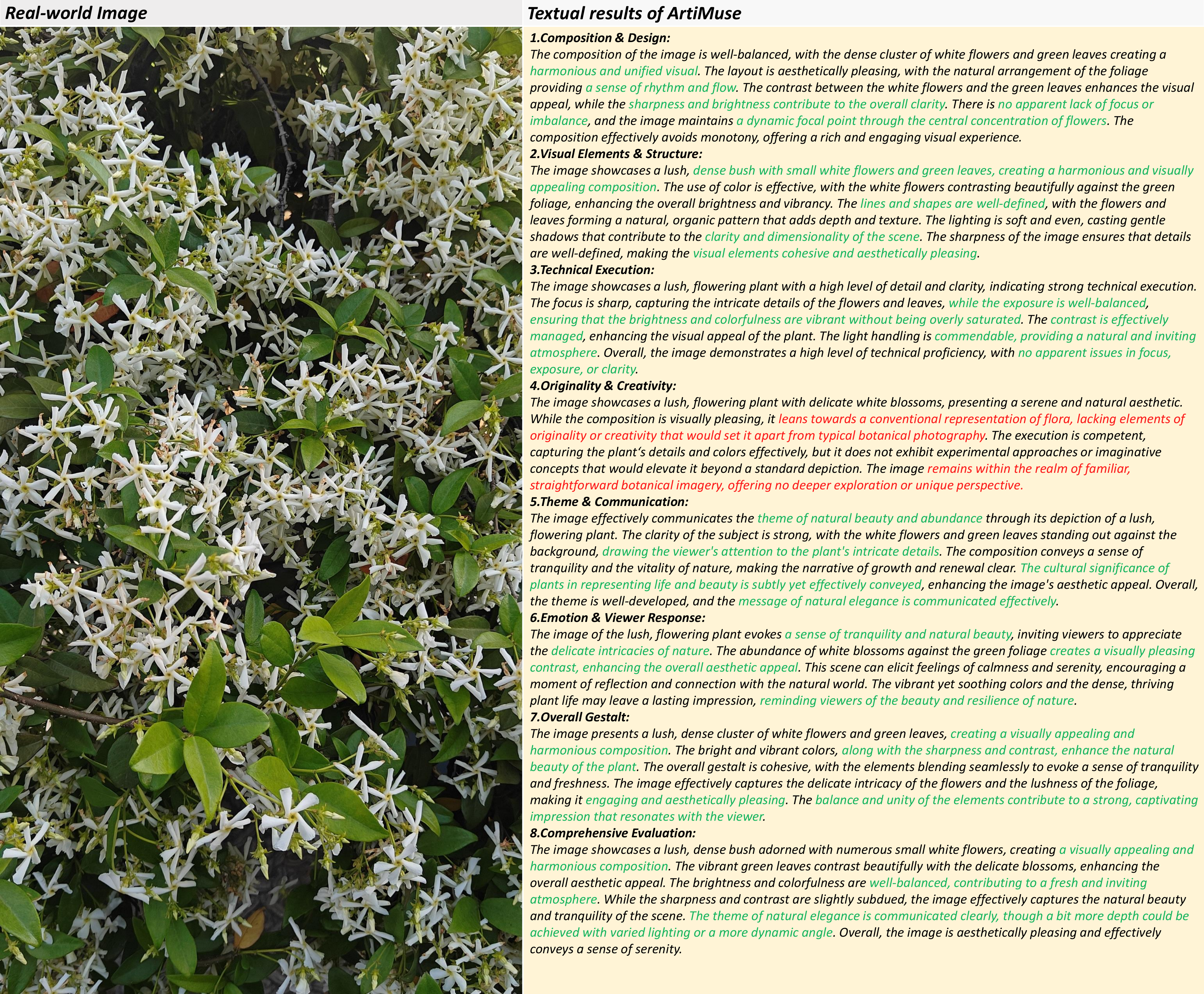}
    % \textbf{\bigskip} % Adds 1cm of vertical space
    % \includegraphics[width=0.99\linewidth]{figures/supp/ex_photo_medium.pdf}
    \vspace{-10pt}
   \caption{Textual results of ArtiMuse on real-world images. \textcolor{red}{Red} and \textcolor{green}{green} denote positive and negative analyses, respectively. ArtiMuse delivers expert-level image analysis, offering accurate evaluations of both strengths and weaknesses.}
   \label{fig:real1}
   \vspace{-10pt}
\end{figure}

\begin{figure}[htbp]
  \centering
  \vspace{-5pt}
    \includegraphics[width=0.93\linewidth]{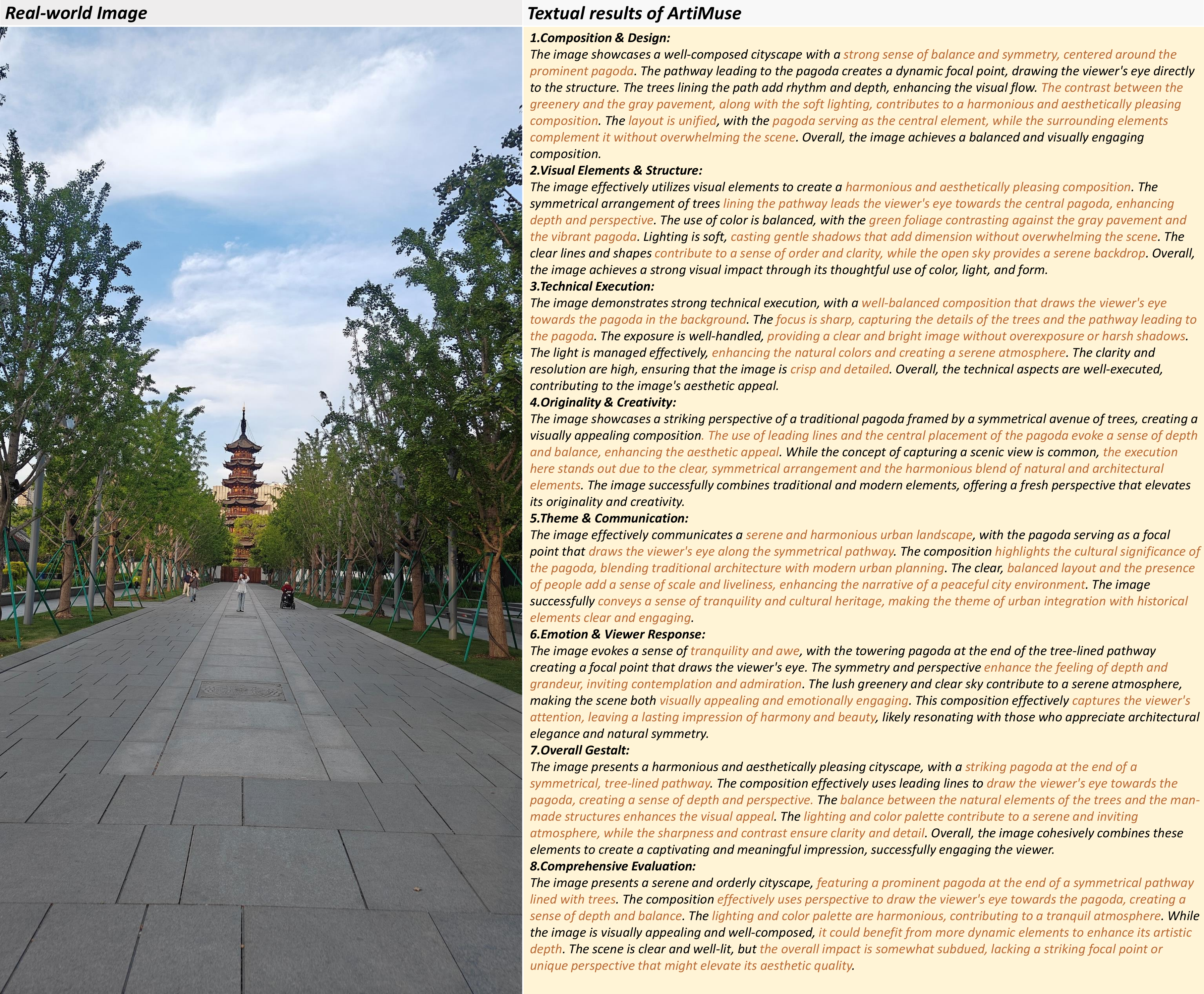}
    % \textbf{\bigskip} % Adds 1cm of vertical space
    % \includegraphics[width=0.99\linewidth]{figures/supp/ex_photo_medium.pdf}
    \vspace{-10pt}
   \caption{Textual results of ArtiMuse on real-world images. \textcolor{brown}{Brown} denote expert-level analyses. ArtiMuse is capable of generating expert-level, granular assessments of visual content.}
   \label{fig:real2}
   \vspace{-10pt}
\end{figure}

\begin{figure}[htbp]
  \centering
  \vspace{-10pt}
    \includegraphics[width=0.93\linewidth]{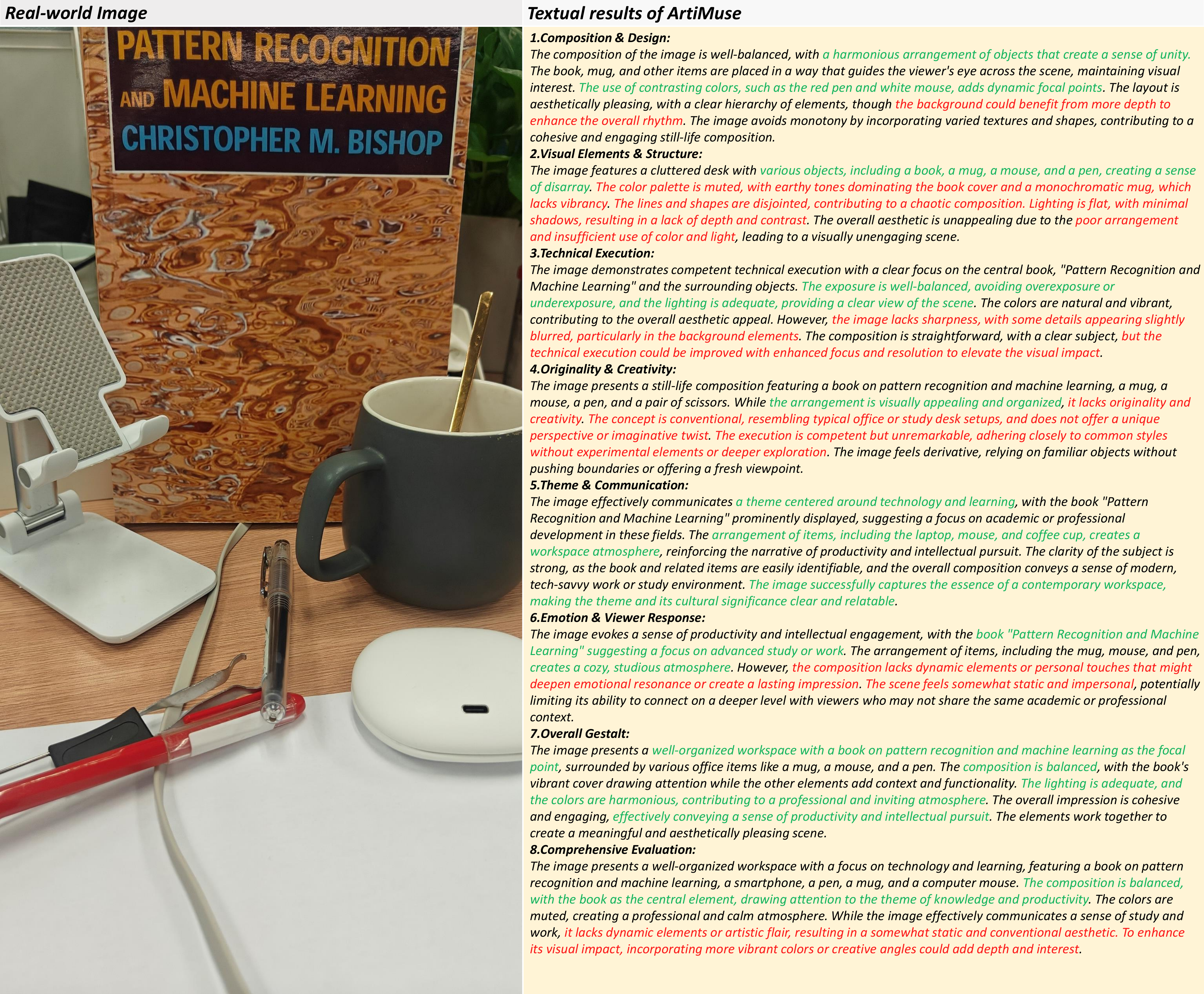}
    % \textbf{\bigskip} % Adds 1cm of vertical space
    % \includegraphics[width=0.99\linewidth]{figures/supp/ex_photo_medium.pdf}
    \vspace{-10pt}
   \caption{Textual results of ArtiMuse on real-world images. \textcolor{red}{Red} and \textcolor{green}{green} denote positive and negative analyses, respectively. ArtiMuse delivers expert-level image analysis, offering accurate evaluations of both strengths and weaknesses.}
   \label{fig:real3}
   \vspace{-10pt}
\end{figure}

To evaluate ArtiMuse's capability in processing out-of-distribution images, we employed real-world images for testing. As demonstrated in Fig.~\ref{fig:real1}, Fig.~\ref{fig:real2} and Fig.~\ref{fig:real3}, our model maintains accurate and expert-level analysis even when handling real-world scenarios. The results showcase ArtiMuse's ability to provide professional aesthetic assessments, systematically identifying both strengths and weaknesses based on detailed visual characteristics.

\begin{figure}[htbp]
  \centering
  \vspace{-10pt}
    \includegraphics[width=0.95\linewidth]{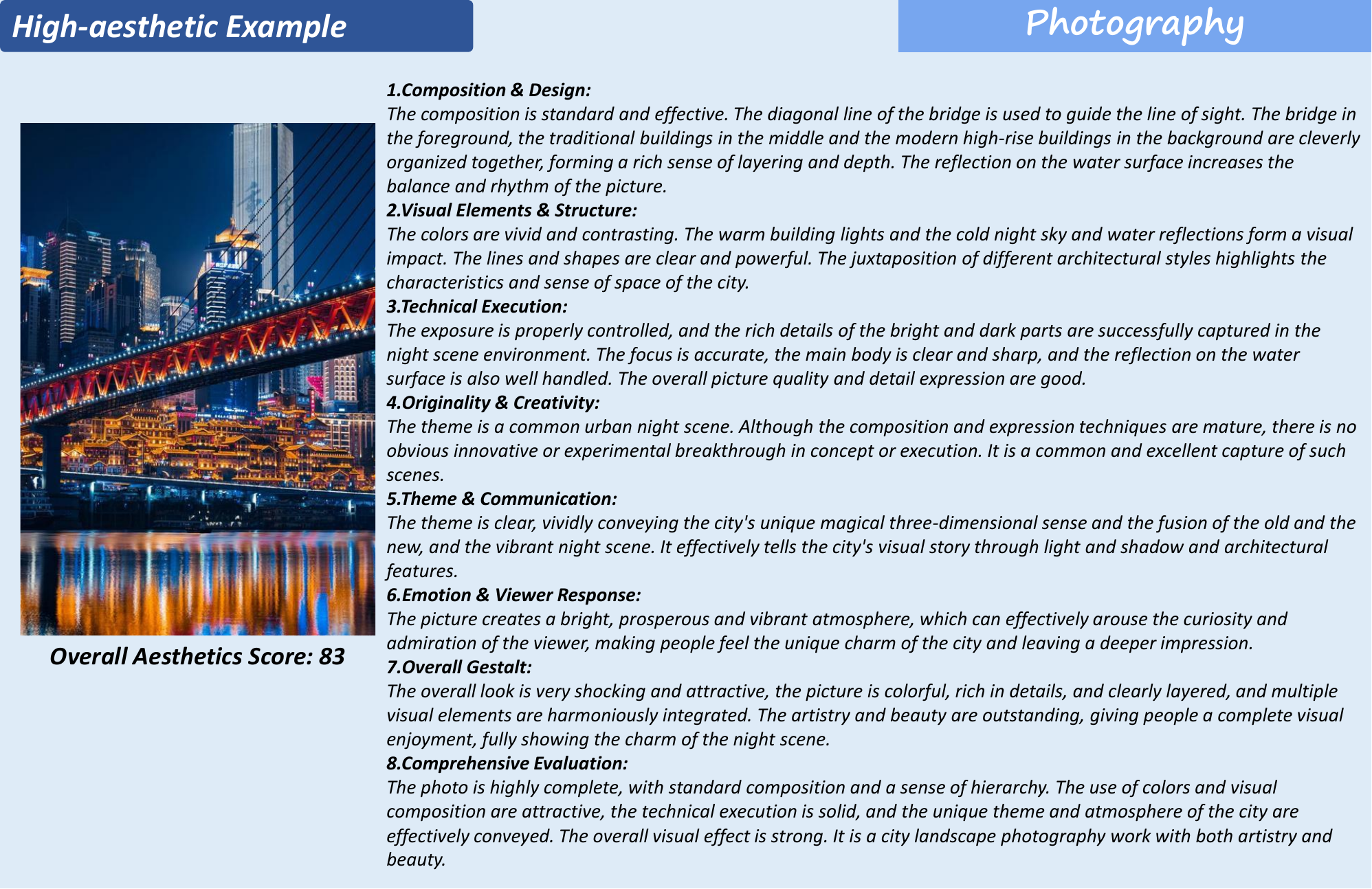}
    % \textbf{\bigskip} % Adds 1cm of vertical space
    % \includegraphics[width=0.99\linewidth]{figures/supp/ex_photo_medium.pdf}
    \vspace{-10pt}
   \caption{High-aesthetic example from \textit{Photography} category.}
   \label{fig:test}
   \vspace{-10pt}
\end{figure}

\end{document}